# Conservative Inference Rule
# for Uncertain Reasoning under Incompleteness

**Marco Zaffalon**                                                    ZAFFALON@IDSIA.CH
*Galleria 2*
*IDSIA*
*CH-6928 Manno (Lugano), Switzerland*

**Enrique Miranda**                                          MIRANDAENRIQUE@UNIOVI.ES
*Department of Statistics and Operations Research*
*University of Oviedo*
*C-Calvo Sotelo, s/n*
*33007 Oviedo, Spain*

## Abstract

In this paper we formulate the problem of inference under incomplete information in very general terms. This includes modelling the process responsible for the incompleteness, which we call the *incompleteness process.* We allow the process' behaviour to be partly unknown. Then we use Walley's theory of *coherent lower previsions*, a generalisation of the Bayesian theory to imprecision, to derive the rule to update beliefs under incompleteness that logically follows from our assumptions, and that we call *conservative inference rule.* This rule has some remarkable properties: it is an abstract rule to update beliefs that can be applied in any situation or domain; it gives us the opportunity to be neither too optimistic nor too pessimistic about the incompleteness process, which is a necessary condition to draw reliable while strong enough conclusions; and it is a *coherent* rule, in the sense that it cannot lead to inconsistencies. We give examples to show how the new rule can be applied in expert systems, in parametric statistical inference, and in pattern classification, and discuss more generally the view of incompleteness processes defended here as well as some of its consequences.

## 1. Introduction

We consider a very general inference problem: we want to draw *conclusions* $Z$ from the observation of *facts* $Y$. Here $Z$ and $Y$ are variables that are related, in the sense that observing the value $y$ of $Y$ in $\mathcal{Y}$ may change our beliefs about which value $z$ the target variable $Z$ assumes in $\mathcal{Z}$.[1]

Although apparently simple, the above setting already captures the main features of many important problems, such as making inference in expert systems, learning the values of some statistical parameters from data, learning from data how to classify new objects into one out of a set of preestablished categories (i.e., doing so-called *pattern classification*), and others.

---

1. Throughout the paper, we shall maintain the convention of using capital letters for variables, the corresponding calligraphic letters for their spaces of possibilities, and lower-case letters for the elements of such spaces.

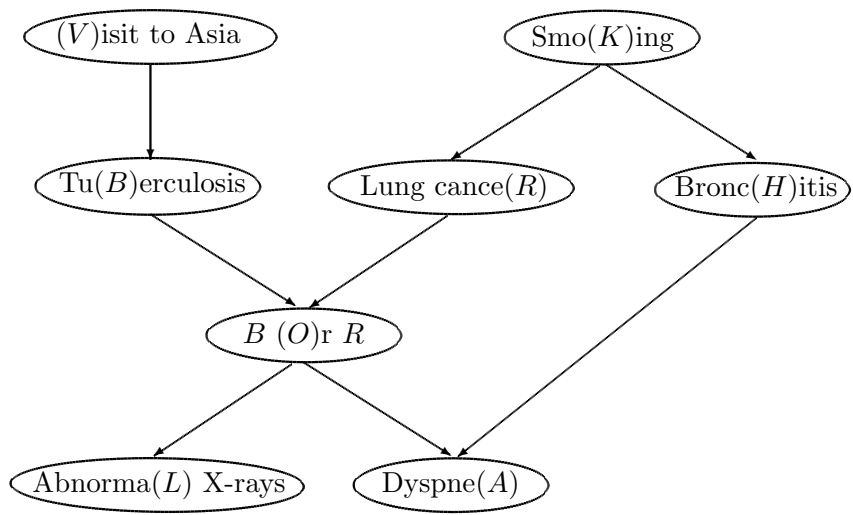

Figure 1: The Bayesian network called Asia. The letters in parentheses denote the variables corresponding to the nodes. Each variable, say $X$, is binary with states $x'$ for 'yes' and $x''$ for 'no'.

Let us make this more concrete with the help of the graphical language of *Bayesian networks* (Pearl, 1988):[2] consider the well-known 'Asia' net displayed in Figure 1, which is intended to model an artificial medical problem. The nodes of a Bayesian network are variables and the arcs model probabilistic dependencies between them; each node holds the probability distribution of the node itself given any joint state of the parent nodes. The probabilities making up these distributions are also called the *parameters* of the network.

Now, if we assume that the network (both the graph and the parameters) is provided by a domain expert, then we are in the field of expert systems. Say that the network is used for diagnosing lung cancer; in this case $R$ is the target node while the others are used to predict the value of $R$. Therefore in this case $Z$ corresponds to $R$ and $Y$ is the vector $(V, K, B, H, O, L, A)$. In another situation, we may want to infer some of the parameters from data. For example, denote by $\Theta$ the chance that there is tuberculosis conditional on a recent visit to Asia, and say that there is a sample $D$ of joint values of $B$ and $V$ from which we wish to infer the value of $\Theta$. In this problem of so-called *parametric inference*, $Z$ corresponds to $\Theta$ and $Y$ to the sample $D$. Finally, say that our goal is to use the Asia net to learn from data how to diagnose lung cancer, i.e., to predict the state of $R$ for the next patient we see, whom we characterise by the vector $(V, K, B, H, O, L, A)$. In this case we need to collect, in a data set $D$, the values of *all* the variables in the network for the past patients. This data set is exploited to infer the parameters of the Asia net, which is then used for classification by predicting the value of $R$ as in the case of expert systems. Therefore, when the focus is on pattern classification, $Z$ corresponds to the so-called *class variable*, namely $R$, while $Y$ is the tuple $(D, V, K, B, H, O, L, A)$.

---

2. The results in the present paper are not restricted to the case of Bayesian networks, but since they can be applied to Bayesian networks, we often use them to convey the intuition more easily.

The common feature of all the previous examples is that observing $Y$ is useful for inferring $Z$, and indeed this is the reason why $Y$ has been introduced in the model. But there is a subtle point about observing $Y$ that is important to realise: very often, the observation of a fact does not coincide with the fact itself. For example, consider again the Asia network, focusing on the problem of parametric inference: in this case, it may well happen that we make mistakes while collecting the values of $B$ and $V$ in a sample; we might for instance mix up some 'yes' values with 'no' values. It is useful to regard this situation as related to two levels of information: the *latent* level of the *actual* sample that records the right values of the variables, and the *manifest* level of the *observed* sample, which is related to, but does not necessarily coincide with the actual sample, and which is going to be used for the inference. We can exploit the paradigm based on the latent and the manifest level more generally as a powerful conceptual tool. For instance, when we use the Asia network as an expert system, it may be the case that for a certain patient characterised by the vector of values $(v', k'', b'', h'', o', l', a')$, it is only possible to access the values of the variables $H$ and $A$. Therefore our observation of the characteristics of the patient will be $(?, ?, ?, h'', ?, ?, a')$, where we denote the symbol of missing value by a question mark. Again, we can think of the former vector as latent and the latter as manifest. More generally speaking, the idea underlying the present discussion is that the devices that we use to observe $Y$, whatever they are, may not let us see $Y$ exactly as it is.

In order to account for this problem, we explicitly model the observation of $Y$ by a new variable $W$, taking values in a finite set $\mathcal{W}$ of possible observations. We call $W$ the *observation of $Y$*. In our previous terminology, $W$ is a manifest variable while $Y$ is a latent one. In other words, we regard $W$ as the output of the process of observing facts, which is called the *observational process* in this paper (other authors call it the *measurement* process). We can think of many different types of observational processes. In this paper we restrict the attention to the special case of observational processes called *incompleteness processes* (IPs). IPs are processes that lead to set-valued observations by coarsening facts. A special case of IPs are *missingness processes*, i.e., those that turn some facts into their entire possibility space.

For example, the process that prevented some of the variables from being observed in the expert system case is a coarsening process: it has turned the fact $(v', k'', b'', h'', o', l', a')$ into the observation $(?, ?, ?, h'', ?, ?, a')$, which can be regarded as the set of the $2^5$ complete vectors obtained by replacing the question marks with values of the unobserved variables in all the possible ways. On the other hand, if we take the fact under consideration to be a single variable, say $V$, then the process that makes $V$ not observed is just a missingness process as writing a question mark is equivalent to writing the entire possibility space for such a variable.[3]

---

3. With IPs it makes sense to introduce the additional variable $W$ even only because the incomplete observations are not in the possibility space for $Y$; in the case of the Asia net, this happens because Missing or coarsened data are indeed a commonplace with expert systems, because the evidence on which an inference task is based is usually incomplete. But they arise frequently also in many other fields; in data mining, just to mention one, missing data are a pervasive problem in applications as well as an important theoretical area of research. In other words, the challenges posed by IPs are widespread. This also means that the problem of incomplete information appears to be a fundamental component of the general task of uncertain reasoning; it is conceptually deep and leads to complicated problems in practice.

Moreover, while we have powerful theoretical tools for the general task of uncertain reasoning, such as Bayes rule and its generalisations, there are few tools for *uncertain reasoning under incompleteness* in the general case. Currently, the most popular approach is based on the assumption that the IP produces incompleteness by coarsening facts at random, which means non-selectively. The assumption that models non-selectiveness is called *coarsening at random* (or CAR, see Gill et al., 1997), or *missing at random* (or MAR, see Little & Rubin, 1987) in the special case of missingness processes (see also Jaeger, 2008). CAR implies that the incompleteness is non-informative and can be ignored; it thus creates the formal basis for applying the methods developed for complete information to the incomplete case. For example, if we assume that the vector $(?,?,?,h'',?,?,a')$ has been created by a CAR IP, then we are allowed to infer the value of $R$ on the sole basis of the sub-vector $(h'',a')$; more precisely, if we aim at computing the posterior probability of $R=r'$, CAR allows us to write $P(R=r'|W=(?,?,?,h'',?,?,a'))=P(R=r'|H=h'',A=a')$. But this is not the case in general.

In fact, incompleteness may well be informative. For example, in the Asia network, it may be the case that the information on whether or not a person has been to Asia is not provided with the same frequency in the two groups. This is an example of incompleteness generated within a communication protocol, where giving or not some information is a key part of the communication. This somewhat selective way of reporting information, albeit very frequent, is not compatible with the CAR assumption. This was pointed out long ago by Shafer (1985), who also has outlined the implications as well as the complications for uncertain reasoning: among these, the fact that modelling IPs can be a very difficult task. More recently, it has been argued against the frequent use of CAR (Grünwald & Halpern, 2003); by now there is a large agreement in the scientific community that CAR is strong, and hence inappropriate in many situations (see Manski, 2003).

De Cooman and Zaffalon (2004) tried to remedy this by an approach to IPs alternative to CAR that is based on *coherent lower previsions* (Walley, 1991), i.e., closed convex sets of probabilities also called *credal sets* by Levi (1980). This has led to a rule for updating beliefs under incomplete information in expert systems called the *conservative updating rule* (CUR).

If we regard CAR as the most optimistic approach to incomplete information, CUR should be regarded as the most pessimistic: it does not assume nearly anything about the IP, and in practice it leads to inference based on working with the set of all facts consistent with (i.e., all the completions of) the incomplete information at hand. In the previous example in which we wish to compute the posterior probability of $R=r'$, CUR leads us to consider all the $2^5$ completions of the vector $(?,?,?,h'',?,?,a')$, i.e.,

---

the symbol '?' is not a possible value for any variable; observations that contain question marks must necessarily be in the possibility space of $W$.

$(v, k, b, h'', o, l, a')$, $v \in \mathcal{V}$, $k \in \mathcal{K}$, $b \in \mathcal{B}$, $o \in \mathcal{O}$, $l \in \mathcal{L}$, and to compute for each of them $P(r'|v, k, b, h'', o, l, a')$. The posterior inference is then summarised by the lower and upper probabilities $\underline{P}(R = r'|W = (?, ?, ?, h'', ?, ?, a')) := \min_{v,k,b,o,l} P(r'|v, k, b, h'', o, l, a')$ and $\overline{P}(R = r'|W = (?, ?, ?, h'', ?, ?, a')) := \max_{v,k,b,o,l} P(r'|v, k, b, h'', o, l, a')$. In other words, the inference is *imprecise*, i.e., the width of the interval determined by the lower and upper probabilities does not need to be zero, as a logical consequence of the ignorance CUR assumes about the IP. As an interesting side remark, CUR is as flexible as CAR in dealing with incomplete data: both of them eventually lead to conclusions that depend only on the $Y$ variable. The $W$ variable at a certain point of the derivation cancels out.

CUR has some drawbacks. It forces us to always assume ignorance about the IP, even when we know that the process is CAR, for instance. This could be the case of the previous example in which we might know that the variables $B, O, L$ are subject to a CAR IP; by using CUR we should be obliged to ignore this information and take all of their completions, as above; and this would lead to inference that is much too weak. Furthermore, CUR has been developed for expert systems only. It cannot be applied to parametric inference when the parameter space is infinite, and some other limitation in the assumptions leading to CUR may sometimes prevent it from being applied more generally.

This paper is an attempt to get the best out of CAR and CUR. We assume that the IP is actually made of two parts, one that acts as a CAR process and another that is unknown to us. Then we use the theory of coherent lower previsions to derive the corresponding rule, which we call the *conservative inference rule* (CIR).

CIR has the following properties:

- Much like traditional, CAR-based, updating, and unlike CUR, it is a rule for the abstract task of updating beliefs on the basis of observations. As such, it can be applied in every situation. With CIR, different applications follow by simply giving different meanings to facts $Y$, observations $W$, and to the quantity $Z$ in which we are interested.

- CIR allows all the variables involved in an analysis, except for $W$ and $Y$, to take values from infinite spaces (CUR allows only finite spaces). This allows us to easily focus on statistical problems where the goal of inference is often the value of a continuous parameter, or in problems where we also use auxiliary continuous variables that are later 'marginalised' out to build our model.

- It can deal with incomplete observations, not only missing ones as CUR.

- Finally, and importantly, CIR is shown to lead to self-consistent (or *coherent*) inference in the strong sense of Walley (1991, Section 7.1.4(b)).

CIR leads to treat the information made incomplete by the CAR IP similarly to the traditional updating, and that subject to the unknown one similarly to CUR. As an example, consider again the vector $(?, ?, ?, h'', ?, ?, a')$ by assuming that the variables $B, O, L$ are subject to a CAR IP and $V, K$ to an IP whose behaviour is unknown to us. CIR leads to the following posterior lower and upper probabilities for $R = r'$: $\underline{P}(R = r'|W = (?, ?, ?, h'', ?, ?, a')) := \min_{v,k} P(r'|v, k, h'', a')$ and $\overline{P}(R = r'|W = (?, ?, ?, h'', ?, ?, a')) := \max_{v,k} P(r'|v, k, h'', a')$. If, in addition, we know that only two completions of $V, K$ make sense, say $(v', k'')$ and

$(v'', k')$ (this means that the unknown IP is a coarsening rather than a missingness process), the ability of CIR to deal with coarsening processes allows it to take advantage of such an information, leading to the more informative lower and upper probabilities obtained optimising only over those two completions. Furthermore, CIR leads to inference that is based, as its predecessors, only on facts, not on the $W$ variable.

The organisation of the material in this paper follows an attept to make the paper accessible also to the readers that prefer not to go into the formal proofs behind CIR. In particular, the paper is logically divided in two parts. The first part, up to and including Section 5, is intended to describe what can be obtained by CIR and its scope of application. Therefore this part briefly gives some introductory material that is needed to define CIR, such as some notions of coherent lower previsions in Section 2 and of incompleteness processes in Section 3. The definition of CIR is given in Section 4. We discuss its significance and compare it with CUR in Section 4.1. Then we show how CIR can be applied to probabilistic expert systems (Section 5.1), parametric inference (Section 5.2), and pattern classification (Section 5.3), each time giving also an example. In this first part we also discuss a number of properties and consequences of the wider view of IPs that we are defending here (e.g., Sections 5.2.2–5.2.3). Some of these (Section 5.3.2) are particularly important for data mining; this depends on the fact that it is usually not possible to learn about the IP from data, and yet the inferences we do may critically depend on it.

The second part of this paper is more technical as it works out the foundations of CIR. To this aim, we first give advanced notions about coherent lower previsions in Section 6, such as updating, independence, and coherence in the sense of Walley (1991). Then we state a number of probabilistic assumptions in Section 7.1, including CAR, discuss them in Section 7.2 (CAR is also discussed further in Appendix A) and derive CIR in Section 7.3. The parts that are even more technical, and which are needed to show that applying CIR cannot lead to inconsistencies, are relegated to Appendix B.

## 2. Coherent Lower Previsions

We give a short introduction to the concepts and results from the behavioural theory of imprecise probabilities that we shall need to introduce the conservative inference rule. We refer the reader to the work of Walley (1991) for an in-depth study of coherent lower previsions, and to the paper by Miranda (2008) for a survey of the theory.

Consider a possibility space $\Omega$. It may represent for instance the set of possible outcomes $\omega$ of an experiment. In the theory of coherent lower previsions, the beliefs about the likelihood of these outcomes are represented by means of *lower previsions* of *gambles* on $\Omega$:

**Definition 1** *Given a possibility space $\Omega$, a gamble is a bounded real-valued function on $\Omega$. The set of all gambles on $\Omega$ is denoted by $\mathcal{L}(\Omega)$. A lower prevision $\underline{P}$ is a real functional defined on some set of gambles $\mathcal{K} \subseteq \mathcal{L}(\Omega)$.*

A gamble $f$ represents a random reward, which depends on the a priori unknown value $\omega$ of $\Omega$. A lower prevision $\underline{P}$ on a set of gambles $\mathcal{K}$ represents a subject's supremum acceptable buying prices for these gambles, in the sense that for all $\epsilon > 0$ and all $f$ in $\mathcal{K}$, the subject is disposed to accept the uncertain reward $f - \underline{P}(f) + \epsilon$, where $\underline{P}(f)$ and $\epsilon$ can also be seen as constant gambles that are identically equal to the real values $\underline{P}(f)$ and $\epsilon$, respectively.[4] Intuitively, lower previsions represent lower expectations of gambles: our subject should be disposed to buy a gamble for anything smaller than the expected reward; however, his lack of knowledge about the probability of the different rewards may enable him only to give a lower bound for this expectation, accepting then any buying price smaller than this bound. The bound is his lower prevision for the gamble.

We shall use $\mathbb{I}_A$ to denote a special type of gamble: the *indicator function* of the set $A$, i.e., the function whose value is 1 for the elements of $A$ and 0 elsewhere. We shall sometimes use the notation $\underline{P}(A)$ for the lower prevision $\underline{P}(\mathbb{I}_A)$ when no confusion is possible.

Consider variables $X_1, \ldots, X_n$, taking values in the sets $\mathcal{X}_1, \ldots, \mathcal{X}_n$, respectively. For any subset $J \subseteq \{1, \ldots, n\}$ we shall denote by $X_J$ the (new) variable

$$X_J := (X_j)_{j \in J},$$

which takes values in the product space

$$\mathcal{X}_J := \times_{j \in J} \mathcal{X}_j.$$

We shall also use the notation $\mathcal{X}^n$ for $\mathcal{X}_{\{1, \ldots, n\}}$. We shall identify the possibility space $\Omega$ with $\mathcal{X}^n$.

**Definition 2** *Let $J$ be a subset of $\{1, \ldots, n\}$, and let $\pi_J : \mathcal{X}^n \to \mathcal{X}_J$ be the so-called projection operator, i.e., the operator that drops the elements of a vector in $\mathcal{X}^n$ that do not correspond to indexes in $J$. A gamble $f$ on $\mathcal{X}^n$ is called $\mathcal{X}_J$-measurable when for all $x, y \in \mathcal{X}^n$, $\pi_J(x) = \pi_J(y)$ implies that $f(x) = f(y)$. We shall denote by $\mathcal{K}_J$ the set of $\mathcal{X}_J$-measurable gambles.*

This notion means that the value $f$ takes depends only on the components of $x \in \mathcal{X}^n$ that belong to the set $J$.[5]

There is a one-to-one correspondence between the $\mathcal{X}_J$-measurable gambles on $\mathcal{X}^n$ and the gambles on $\mathcal{X}_J$: given an $\mathcal{X}_J$-measurable gamble $f$ on $\mathcal{X}^n$, we can define $f'$ on $\mathcal{X}_J$ by $f'(x) := f(x')$, where $x'$ is any element in $\pi_J^{-1}(x)$; conversely, given a gamble $g$ on $\mathcal{X}_J$, the gamble $g'$ on $\mathcal{X}^n$ given by $g'(x) := g(\pi_J(x))$ is $\mathcal{X}_J$-measurable.

Let $O$ be a subset of $\{1, \ldots, n\}$, and let $\underline{P}(X_O)$ be a *lower prevision* on the set $\mathcal{K}_O$ of $\mathcal{X}_O$-measurable gambles. We say that $\underline{P}$ is *coherent* if and only if the following three conditions hold for all $f, g \in \mathcal{K}_O$, and $\lambda > 0$:

(C1) $\underline{P}(f) \geq \inf f$.

(C2) $\underline{P}(\lambda f) = \lambda \underline{P}(f)$.

---

4. We say then that the gamble $f - \underline{P}(f) + \epsilon$ is *desirable* for our subject, and that $f - \underline{P}(f)$ is *almost-desirable*. It follows from this interpretation that the set of desirable (resp., almost-desirable) gambles should be closed under addition and multiplication by non-negative reals, and that a gamble $f$ that dominates a desirable gamble $g$ should also be desirable (resp., almost-desirable).

5. This notion is related to the common notion of measurability as it implies that the gamble $f : \mathcal{X}^n \to \mathbb{R}$ is indeed a measurable mapping if we consider the $\sigma$-field $\{\pi_J^{-1}(A) : A \subseteq \mathcal{X}_J\}$ on the initial space and any $\sigma$-field in the final space.

(C3) $\underline{P}(f+g) \geq \underline{P}(f) + \underline{P}(g)$.

Coherence means that a subject cannot raise the lower prevision $\underline{P}(f)$ of a gamble by considering the acceptable buying transactions that are implied by other gambles in the domain.

**Remark 1 (Coherent upper previsions)** *Although for most of the paper we shall work with coherent lower previsions, or supremum acceptable buying prices, we shall also use at times the so-called coherent upper previsions. The upper prevision of a gamble $f$, $\overline{P}(f)$, represents the infimum acceptable selling price for $f$ for our subject, in the sense that for any $\epsilon > 0$ the transaction $\overline{P}(f) - f + \epsilon$ is desirable for him.*

*Taking into account this interpretation, it follows that the upper and lower previsions must be conjugate functions, in the sense that $\overline{P}(f) = -\underline{P}(-f)$ for all gambles $f$. This is why we shall work almost exclusively with lower previsions, and use upper previsions only to refer to their conjugate functions when this helps to simplify the notation. Finally, we say that an upper prevision is coherent if so is its conjugate lower prevision.* $\diamond$

It is important at this point to introduce a particular case of coherent lower previsions that will be of special interest for us: *linear previsions*.

**Definition 3** *A lower prevision $\underline{P}(X_O)$ on the set $\mathcal{K}_O$ is* linear *if and only if it is coherent and $\underline{P}(f+g) = \underline{P}(f) + \underline{P}(g)$ for all $f, g \in \mathcal{K}_O$.*

Linear previsions correspond to the case where a subject's supremum acceptable buying price (lower prevision) coincides with his infimum acceptable selling price (or upper prevision) for every gamble on the domain. When a coherent lower prevision $\underline{P}(X_O)$ is linear, we denote it by $P(X_O)$. A linear prevision corresponds to the expectation operator (with respect to the Dunford integral, see the book by Bhaskara Rao & Bhaskara Rao, 1983) with respect to a finitely additive probability.

One interesting feature of linear previsions allows us to easily characterise coherence.

**Definition 4** *$P(X_O)$ is said to* dominate *$\underline{P}(X_O)$ if $P(f) \geq \underline{P}(f)$ for every $\mathcal{X}_O$-measurable gamble $f$.*

A lower prevision $\underline{P}(X_O)$ is coherent if and only if it is the lower envelope of a closed[6] and convex[7] set of dominating linear previsions, which we denote by $\mathcal{M}(\underline{P}(X_O))$. It follows also that $\underline{P}(X_O)$ is the lower envelope of the set of *extreme points* of $\mathcal{M}(\underline{P}(X_O))$. We denote the set of extreme points of $\mathcal{M}(\underline{P}(X_O))$ by $ext(\mathcal{M}(\underline{P}(X_O)))$.

**Example 1** *Assume that our subject has the information that the outcome of the variables in $X_O$ belongs to some finite subset $A$ of $\mathcal{X}_O$, and nothing more. Then he should model these beliefs by the so-called* vacuous *lower prevision $\underline{P}_A(X_O)$ given by $\underline{P}_A(f) := \min_{\omega \in A} f(\omega)$ for every $f \in \mathcal{K}_O$. The set $\mathcal{M}(\underline{P}_A(X_O))$ of dominating linear previsions corresponds to the*

---

6. In the *weak\* topology*, which is the smallest topology for which all the evaluation functionals given by $f(P) := P(f)$, where $f \in \mathcal{L}(\Omega)$, are continuous.

7. That is, for all linear previsions $P_1, P_2$ in the set and all $\alpha \in (0, 1)$, the linear prevision $\alpha P_1 + (1-\alpha)P_2$ also belongs to this set.

*finitely additive probabilities $P(X_O)$ satisfying the constraint $P(A) = 1$. Among these, the extreme points are the degenerate probability measures with respect to some $\omega \in A$, and it follows that any linear prevision in $\mathcal{M}(\underline{P}_A(X_O))$ is a convex combination of these.* $\diamondsuit$

Consider now two disjoint subsets $O, I$ of $\{1, \ldots, n\}$, with $O \neq \emptyset$. $\underline{P}(X_O|X_I)$ represents a subject's behavioural dispositions about the gambles that depend on the outcome of the variables $\{X_k, k \in O\}$, after coming to know the outcome of the variables $\{X_k, k \in I\}$. As such, it is defined on the set of gambles that depend on the values of the variables in $O \cup I$ only,[8] i.e., on the set $\mathcal{K}_{O \cup I}$ of the $\mathcal{X}_{O \cup I}$-measurable gambles on $\mathcal{X}^n$. Given such a gamble $f$ and $x \in \mathcal{X}_I$, $\underline{P}(f|X_I = x)$ represents a subject's supremum acceptable buying price for the gamble $f$, if he came to know that the variable $X_I$ took the value $x$ (and nothing else). We can thus consider the gamble $\underline{P}(f|X_I)$ on $\mathcal{X}_I$, that on $x \in \mathcal{X}_I$ takes the value $\underline{P}(f|X_I = x)$.

**Definition 5** *The functional $\underline{P}(\cdot|X_I)$ that maps a gamble $f$ in its domain $\mathcal{K}_{O \cup I}$ to the gamble $\underline{P}(f|X_I)$ is called a* conditional lower prevision.

This definition is well posed as the sets $\{\pi_I^{-1}(x) : x \in \mathcal{X}_I\}$ form a partition of $\mathcal{X}^n$. When there is no possible confusion about the variables involved in the lower prevision, we shall use the notation $\underline{P}(f|x)$ for $\underline{P}(f|X_I = x)$. In particular, $\underline{P}(y|x)$ will mean $\underline{P}(\mathbb{I}_y|x)$ for all pairs of values $x, y$.

In Walley's theory a conditional lower prevision $\underline{P}(X_O|X_I)$ defined on $\mathcal{K}_{O \cup I}$ is required to be self-consistent, or *separately coherent*. Separate coherence means on the one hand that if a subject knows that the variable $X_I$ has taken the value $x$, he cannot raise the (conditional) lower prevision $\underline{P}(f|x)$ of a gamble by considering the acceptable buying transactions that are implied by other gambles in the domain, and on the other hand that he should bet at any odds on the event that $X_I = x$ after having observed it.

In this case, where the domain is a linear set of gambles, the definition is the following:

**Definition 6** *The conditional lower prevision $\underline{P}(X_O|X_I)$ is* separately coherent *if and only if for all $x \in \mathcal{X}_I, f, g \in \mathcal{K}_{O \cup I}$, and $\lambda > 0$:*

*(SC1)* $\underline{P}(f|x) \geq \inf_{\omega \in \pi_I^{-1}(x)} f(\omega)$.

*(SC2)* $\underline{P}(\lambda f|x) = \lambda \underline{P}(f|x)$.

*(SC3)* $\underline{P}(f + g|x) \geq \underline{P}(f|x) + \underline{P}(g|x)$.

Using these conditions, we see more clearly that a separately coherent conditional lower prevision can also be regarded as a lower bound for a conditional expectation. It also follows that if $I = \emptyset$, separate coherence coincides with the notion of coherence introduced above.

Given a separately coherent conditional lower prevision $\underline{P}(X_O|X_I)$ with domain $\mathcal{K}_{O \cup I}$, we can see $\underline{P}(X_O|x)$ as defined on the set of $\mathcal{X}_O$-measurable gambles, because for any $f \in \mathcal{K}_{O \cup I}$, $\underline{P}(f|x) = \underline{P}(g|x)$, where $g$ is the $\mathcal{X}_O$-measurable gamble given by $g(w) = f(\pi_{I^c}(w), x)$ for all $w \in \mathcal{X}^n$.

---

8. We refer to the work by Miranda and De Cooman (2005) and Walley (1991) for more general definitions of the following notions in this section in terms of partitions, and for domains that are not necessarily (these) linear sets of gambles.

As with the unconditional case, we can also consider conditional *upper* previsions, which represent our subject's infimum acceptable selling prices for the gambles $f$ in $\mathcal{K}_{O \cup I}$, after coming to know the value of the variables in $X_I$ (and nothing else). We have $\overline{P}(f|x) = -\underline{P}(-f|x)$ for all gambles $f$. Similarly, a conditional lower prevision $\underline{P}(X_O|X_I)$ on the set $\mathcal{K}_{O \cup I}$ is *linear* if and only if it is separately coherent and $\underline{P}(f + g|x) = \underline{P}(f|x) + \underline{P}(g|x)$ for all $x \in \mathcal{X}_I$ and $f, g \in \mathcal{K}_{O \cup I}$. Conditional linear previsions correspond to the case where a subject's supremum acceptable buying price (lower prevision) coincides with his infimum acceptable selling price (or upper prevision) for every gamble in the domain. When a separately coherent conditional lower prevision $\underline{P}(X_O|X_I)$ is linear, we denote it by $P(X_O|X_I)$.

A conditional lower prevision $P(X_O|X_I)$ *dominates* $\underline{P}(X_O|X_I)$ if $P(f|x) \geq \underline{P}(f|x)$ for every $\mathcal{X}_{O \cup I}$-measurable gamble $f$ and every $x \in \mathcal{X}_I$. A conditional lower prevision $\underline{P}(X_O|X_I)$ is separately coherent if and only if it is the lower envelope of a closed and convex set of dominating conditional linear previsions, which we denote by $\mathcal{M}(\underline{P}(X_O|X_I))$. This is a bit of an abuse of notation, since actually for every $x \in \mathcal{X}_I$ the set $\mathcal{M}(\underline{P}(X_O|x))$ is a set of linear previsions. It follows also that $\underline{P}(X_O|X_I)$ is the lower envelope of the set of *extreme points* of $\mathcal{M}(\underline{P}(X_O|X_I))$, where we say that $P(X_O|X_I)$ is an extreme point of $\mathcal{M}(\underline{P}(X_O|X_I))$ when for every $x \in \mathcal{X}_I$, $P(X_O|x)$ is an extreme point of the closed convex set $\mathcal{M}(\underline{P}(X_O|x))$. We denote the set of extreme points of $\mathcal{M}(\underline{P}(X_O|x))$ by $ext(\mathcal{M}(\underline{P}(X_O|x)))$.

**Example 2** *Consider the following experiment: a subject throws a coin; if it lands on heads, he selects a ball from an urn with red and white balls of unknown composition; if it lands on tails, he selects a ball from an urn with red and blue balls also of unknown composition. Let $X_1$ be the result of the first experiment, with values in $\{heads, tails\}$, and $X_2$ be the color of the ball drawn in the second experiment, with values in $\{red, white, blue\}$.*

*We may model this using the conditional prevision $\underline{P}(X_2|X_1)$ where $\underline{P}(X_2|X_1 = heads)$ is vacuous on $\{red, white\}$, and $\underline{P}(X_2|X_1 = tails)$ is vacuous on $\{red, blue\}$. The extreme points of $\mathcal{M}(\underline{P}(X_2|X_1))$ are the conditional previsions $P(X_2|X_1)$, where $P(X_2|heads)$ is degenerate on either red or white, and $P(X_2|tails)$ is degenerate on either red or blue.* $\diamondsuit$

## 3. The Basic Setting

In this section we introduce the basic assumptions about the spaces considered in our model and about the incompleteness process.

### 3.1 The Domain

As we mentioned in the Introduction, in this paper we consider the problem of drawing conclusions about the value that a target variable $Z$ takes in $\mathcal{Z}$ from information about of the value that another variable $Y$ takes in a set $\mathcal{Y}$. In this paper we shall assume that the set $\mathcal{Y}$ is finite, but the set $\mathcal{Z}$ of possible values for the target variable $Z$ can be infinite.

### 3.2 The Incompleteness Process

It is not uncommon that the devices that we use to get information about $Y$, whatever they are, may not let us see $Y$ exactly as it is. Because of this, we explicitly model the observation of $Y$ by a new variable $W$, taking values in the finite set $\mathcal{W}$ of possible observations. We call $W$ the *observation of $Y$*. $W$ represents the outcome of the observational process. In this paper we focus on the observational processes called incompleteness processes, which can not only turn a fact into its entire possibility space, but also into other nonempty subsets of the possibility space.

**Remark 2 (Concerning the $W$ variable)** *We remark that it is necessary to introduce the variable $W$ even if it is quite a common habit in applications to deal only with the variables $Z$ and $Y$. The possibility to drop $W$ depends on the assumptions done. For instance, if we assume CAR/MAR then the $W$ variable at a certain point of the derivation cancels out, leading to formulae that only involve $Z$ and $Y$. But one should not confuse the operational procedures with the theoretical representation needed for the derivation of a rule. The theoretical derivation has to take into account that the latent level of information, represented by $Y$, does not need to coincide with the manifest level, namely $W$, and hence that we need to model the process that leads from $Y$ to $W$. This is also necessary because there are assumptions other than those made in this paper that do not lead eventually to drop $W$. Moreover, the distinction between the latent and the manifest level is not a new proposal at all: on the contrary, it is present, somewhat implicitly, even in the original works about MAR (Little & Rubin, 1987) and fully explicitly in more recent works (Grünwald & Halpern, 2003).*◇

We focus now on representing the IP. We start by characterising the IP for $Y$ by a so-called *multi-valued map* $\Gamma_Y$ (this idea of modelling the IP through a multi-valued map goes back essentially to Strassen, 1964). $\Gamma_Y$ is what connects facts with observations: for all $y \in \mathcal{Y}$, $\Gamma_Y$ gives us the set $\Gamma(y) \subseteq \mathcal{W}$ of observations into which the IP may turn fact $y$. We require such a set to be nonempty:

$$y \in \mathcal{Y} \Rightarrow \Gamma(y) \neq \emptyset. \tag{IP1}$$

Take for example the Asia network. If the fact under consideration is the instance $y := (v', k'', b'', h'', o', l', a')$ of the vector $(V, K, B, H, O, L, A)$, taken to be equal to $Y$, then $\Gamma(y)$ represents the set of all the incomplete instances that may be generated by the IP starting from $y$. For instance, the IP may be such that $\Gamma(y)$ is equal to the set of $2^7$ incomplete instances obtained from $(v', k'', b'', h'', o', l', a')$ by giving the possibility to replace the values in the vector with question marks in all the possible ways.

$\Gamma_Y$ makes it possible to associate, to each observation $w \in \mathcal{W}$, the set of facts that may originate it: i.e.,

$$\{w\}^* := \{y \in \mathcal{Y} : w \in \Gamma(y)\}.$$

In the Asia network, our observation might be $w = (?, ?, ?, h'', ?, ?, a')$. If $\Gamma(y)$ is defined as above, then $\{w\}^*$ is the set of the $2^5$ completions of $(?, ?, ?, h'', ?, ?, a')$. $\Gamma(y)$ might also be defined differently, for instance by allowing only for *some* replacements of values in the vector $y$ with question marks (the possibility to replace values with question marks might also depend on the values that some variables in the vector take jointly). In this case $\{w\}^*$ would be a subset of the $2^5$ completions. What is important is that the set $\{w\}^*$ does not allow us to identify the value of $Y$ uniquely, unless it is a singleton. This tells us that IPs are observational processes that produce $W$ by *coarsening* $Y$, i.e., by yielding an observation $w$ corresponding to a set $\{w\}^*$ of possible values for $Y$ that we expect to encompass $y$ (this expectation will be formalised in Section 7.1.2). It follows that IPs are a generalisation of the, perhaps more popular, *missingness processes*, which consider only two possibilities: either $\{w\}^*$ is a singleton or it is $\mathcal{Y}$, in which case $Y$ is said to be *missing*. In the case of the Asia net, we could characterise a missingness process by writing $\Gamma(y) = \{(v, k, b, h, o, l, a), (?, ?, ?, ?, ?, ?, ?)\}$ for all $y = (v, k, b, h, o, l, a) \in \mathcal{Y}$.[9] If $\{w\}^* = \emptyset$, $w$ cannot be produced by any $y \in \mathcal{Y}$, and can therefore be eliminated from $\mathcal{W}$ without any further consequences; we shall henceforth assume that

$$w \in \mathcal{W} \Rightarrow \{w\}^* \neq \emptyset. \tag{IP2}$$

Finally, we define for each $w \in \mathcal{W}$ the set of facts of which $w$ is the only compatible observation:

$$\{w\}_* := \{y \in \mathcal{Y} : \Gamma(y) = \{w\}\}.$$

## 3.3 Refining Facts, Observations, and the Incompleteness Process

In this section, we add some structure to the incompleteness process by representing it as the combination of two different incompleteness processes, which act on different parts of a fact. We model these two parts by writing $Y := (\bar{Y}, \hat{Y})$, where $\bar{Y}$ and $\hat{Y}$ are two new variables, with values in $\bar{\mathcal{Y}}$ and $\hat{\mathcal{Y}}$, respectively, such that $\mathcal{Y} = \bar{\mathcal{Y}} \times \hat{\mathcal{Y}}$. In an analogous way, we regard the observation $W$ of $Y$ as the observation $\bar{W}$ of $\bar{Y}$ jointly with the observation $\hat{W}$ of $\hat{Y}$, and hence write $W := (\bar{W}, \hat{W})$. Here again, $\bar{W}$ and $\hat{W}$ are two new variables, with values in $\bar{\mathcal{W}}$ and $\hat{\mathcal{W}}$, respectively, such that $\mathcal{W} = \bar{\mathcal{W}} \times \hat{\mathcal{W}}$.

The additional variables introduced indeed allow us to think of two new IPs: the first acts on variable $\bar{Y}$, leading to observation $\bar{W}$, and thus is characterised by a certain multi-valued map $\Gamma_{\bar{Y}}$; the second acts on variable $\hat{Y}$, leading to observation $\hat{W}$, and is characterised by another multi-valued map $\Gamma_{\hat{Y}}$. We call them the *unknown IP* and the *CAR IP*, respectively, as by the former we aim at modelling an IP whose behaviour is unknown to us and by the latter a CAR IP.

Assuming that

$$y \in \mathcal{Y} \Rightarrow \Gamma(y) = \Gamma(\bar{y}) \times \Gamma(\hat{y}) \tag{IP3}$$

allows us to regard the two IPs, taken together, as the single incompleteness process introduced in Section 3.2, i.e., the one that maps $(\bar{Y}, \hat{Y}) = Y$ into $(\bar{W}, \hat{W}) = W$. From now on, we call this the *overall IP*. Assumption (IP3) follows as a consequence of our intention to model problems where there are two IPs which observe different parts of a fact and therefore do not interact. This is discussed at some length at the end of Section 7.2.

We now impose some assumptions about the unknown and the CAR IP. Consider the sets $\{\bar{w}\}^*$, $\{\bar{w}\}_*$, $\{\hat{w}\}^*$, and $\{\hat{w}\}_*$, defined in the obvious way on the basis of $\Gamma_{\bar{Y}}$ and $\Gamma_{\hat{Y}}$. The following assumptions resemble Assumptions (IP1)–(IP2) and are motivated by the same arguments:

$$\bar{y} \in \bar{\mathcal{Y}} \Rightarrow \Gamma(\bar{y}) \neq \emptyset \tag{IP4}$$

---

9. To avoid confusion, it may be worth outlining that whether a process is a coarsening or a missingness process depends on what we focus on. In the case of the Asia network, for instance, the process can be a missingness process for each single variable of the net (i.e., it yields either the value of the variable or a question mark) and simultaneously be a coarsening process for the vector of variables (in the sense that it does not only yield the vector of values or the vector made entirely of question marks).

$$\hat{y} \in \hat{\mathcal{Y}} \Rightarrow \Gamma(\hat{y}) \neq \emptyset \tag{IP5}$$

$$\bar{w} \in \bar{\mathcal{W}} \Rightarrow \{\bar{w}\}^* \neq \emptyset \tag{IP6}$$

$$\hat{w} \in \hat{\mathcal{W}} \Rightarrow \{\hat{w}\}^* \neq \emptyset. \tag{IP7}$$

We also impose an additional requirement on the unknown IP:

$$\bar{w} \in \bar{\mathcal{W}} \Rightarrow \{\bar{w}\}_* = \emptyset. \tag{IP8}$$

We do so as a means to start implementing the idea that there is ignorance about the procedure used by the unknown IP to select an observation starting from a fact.

Having $\Gamma(y) = \Gamma(\bar{y}) \times \Gamma(\hat{y})$ for all $y \in \mathcal{Y}$ implies that $\{w\}^* = \{\bar{w}\}^* \times \{\hat{w}\}^*$ and $\{w\}_* = \{\bar{w}\}_* \times \{\hat{w}\}_*$ for all $w \in \mathcal{W}$. This shows that (IP4)–(IP7) imply (IP1)–(IP2).

## 4. The Conservative Inference Rule

The notations introduced so far finally allow us to write the definition of the conservative inference rule. To this extent, we assume that we have beliefs about $(Z, Y)$ in the form of a joint lower prevision $\underline{P}_1$.

**Definition 7 (Conservative inference rule)** *Consider a gamble $g$ on $\mathcal{Z}$ and $w \in \mathcal{W}$, and assume that $\overline{P}_1(\{w\}^*) > 0$. Let $\{\bar{w}\}_1^* := \{\bar{y} \in \{\bar{w}\}^* : \overline{P}_1(\bar{y}, \{\hat{w}\}^*) > 0\}$, and define*

$$\underline{R}(g|\bar{y}, \{\hat{w}\}^*) := \inf_{P \geq \underline{P}_1 : P(\bar{y}, \{\hat{w}\}^*) > 0} P(g|\bar{y}, \{\hat{w}\}^*)$$

*for all $\bar{y} \in \{\bar{w}\}_1^*$. Then we let*

$$\underline{R}(g|w) := \min_{\bar{y} \in \{\bar{w}\}_1^*} \underline{R}(g|\bar{y}, \{\hat{w}\}^*). \tag{CIR}$$

Later, in Section 7.3, we shall see that this definition actually follows as a theorem under a certain number of assumptions.

Let us clarify the intuition behind the definition of $\underline{R}(g|w)$. Our goal is to update beliefs about a certain function $g$ of $Z$ once we make the observation $w$ about $Y$. Remember that $w$ can be regarded as the pair $(\bar{w}, \hat{w})$, where $\bar{w}$ is the part of the observation originated by the unknown IP and $\hat{w}$ that originated by the CAR IP. The conservative inference rule prescribes to update beliefs by adopting the lower prevision $\underline{R}(g|w)$ computed as in Formula (CIR). This means (i) to consider all the completions $\bar{y}$ of $\bar{w}$ with the property that the conditioning event $(\bar{Y} = \bar{y}, \hat{Y} \in \{\hat{w}\}^*)$ has positive upper probability under $\underline{P}_1$; (ii) to compute our updated beliefs under each of these events: we can do this by applying Bayes' rule to each linear prevision that dominates $\underline{P}_1$ and for which the event has positive probability, so as to create a set of posterior linear previsions whose lower envelope is $\underline{R}(g|\bar{y}, \{\hat{w}\}^*)$; finally, to define the lower prevision $\underline{R}(g|w)$ as the minimum of the lower previsions obtained under all the completions considered. This minimum, in particular, has the meaning to be as conservative as possible with respect to the data that has been made incomplete by the unknown IP: we deal with them by considering all the complete data that could have been there before the IP started operating. On the other hand, the data subject to the CAR IP are treated as usual, that is, by conditioning on the set of completions $\{\hat{w}\}^*$.

## 4.1 Significance of the Conservative Inference Rule

We can think of CIR as a generalisation of two kinds of updating rules. It generalises the traditional updating rule, the one that, for instance, prescribes discarding missing observations. CIR coincides with such a rule in the case the IP is made only of the CAR component. On the other hand, if there is no CAR component, and the overall IP is unknown, CIR is similar to (but more powerful than) the so-called *conservative updating rule* (CUR) proposed by De Cooman and Zaffalon (2004). When both components are present, CIR acts as a mix of traditional updating and CUR.

As CUR, CIR is an imprecise-probability rule: it generally leads to lower and upper expectations, and partially determined decisions. This follows, in part, as it allows $\underline{P}_1$ to be imprecise. Yet, even if we take that to be a linear prevision $P_1$, CIR turns out to be an imprecise-probability rule: imprecision arises as a logical consequence of our ignorance about the unknown IP.

### 4.1.1 COMPARISON WITH CUR

It is instructive to analyse more deeply the difference between CIR and CUR. To this extent, let us first give the definition of CUR using the notation introduced so far:

**Definition 8 (Conservative updating rule)** *Assume that $\mathcal{Z}$ is a finite set, that $Y = \bar{Y}$ and $W = \bar{W}$, that is, that the overall IP is entirely unknown. Furthermore, assume that $\Gamma_{\bar{Y}}$ is originated by missing values: this means that if we regard $\bar{Y}$ as vector-valued, each of its components is either observed precisely or it is missing. Assume also that $\overline{P}_1(\bar{y}) > 0$ for all $\bar{y} \in \{\bar{w}\}^*$. For every gamble $g$ on $\mathcal{Z}$, define*

$$\underline{R}(g|\bar{y}) := \inf_{P \geq \underline{P}_1 : P(\bar{y}) > 0} P(g|\bar{y})$$

*for all $\bar{y} \in \{\bar{w}\}^*$. Then we let*

$$\underline{R}(g|\bar{w}) := \min_{\bar{y} \in \{\bar{w}\}^*} \underline{R}(g|\bar{y}). \tag{CUR}$$

The major differences between CIR and CUR are discussed below.

- The first difference is that CUR allows only the unknown IP to be present. In doing so, CUR does not model beliefs stronger than ignorance about the IP, for which CUR-based inferences will be more conservative than necessary. CIR tries to remedy this by allowing for mixed states of knowledge made of ignorance and CAR. This makes CIR a flexible rule that should lead to strong enough conclusions in many applications.

- The by far most important difference between CIR and CUR is the generality of application. The theory used to derive CUR is restricted to the case when $\underline{P}_1$ is directly assessed rather than obtained through a number of conditional and unconditional lower previsions, possibly together with some notion of independence. This is the case, for instance, of statistical inference which involves using a certain number of lower previsions to model prior knowledge and the likelihood function and some notion of independence (or exchangeability) to build $\underline{P}_1$. Although one can apply CUR also in these more general and common conditions, its theory does not guarantee that CUR leads to self-consistent inference in those cases. CIR, on the other hand, is shown in Sections 7.3 to lead to coherent inference under a very wide spectrum of conditions, which should cover nearly all the practical situations. This means that applying CIR always leads to probabilities that are self-consistent as well as consistent with the original assessments. (Observe that since CUR is a special case of CIR, our proofs finally show that CUR also leads to self-consistent inference.) In other words, CIR is more much than CUR a rule for the general task of updating beliefs based on the observation of facts. In this, it is similar in spirit to traditional updating, which we typically use in every situation: e.g., in the case of expert systems as well as to update our beliefs given data in problems of statistical inference. With CIR, different applications follow by simply giving different meanings to facts $(\bar{Y}, \hat{Y})$, observations $(\bar{W}, \hat{W})$, and to the quantity $Z$ in which we are interested. Section 5 will give examples to show how this is done in a variety of cases.

There are three further characteristics that make CIR more flexible than CUR.

- The first is that CIR allows the target space to be infinite, while CUR was defined only for the case of finite spaces.

- The second is that CIR deals with general incompleteness processes, while CUR only with the special case of processes that may originate missing information for the variables under consideration. This is related to the restriction about $\Gamma_{\bar{Y}}$ mentioned in the definition of CUR. This characteristic of CIR is important when aiming to obtain as strong conclusions as possible: by taking advantage of partially observed facts, CIR leads in general to stronger conclusions than CUR.

- The third is somewhat more technical. CUR requires that *every element* in $\{\bar{w}\}^*$ is given positive upper probability. CIR requires this only for the entire set $\{\bar{w}\}^*$. The difference is important in practice. To see this, consider that in applications it is quite common to represent deterministic relations among variables by degenerate probabilities equal to zero or one. These relations naturally give rise to zero upper probabilities: every joint state of the variables that does not satisfy the relation has zero upper probability by definition. CUR cannot be used in these applications. CIR can, and it simply leads to neglect the states with zero upper probabilities: more precisely, given an observation $(\bar{w}, \hat{w})$, it leads to consider only the compatible facts $(\bar{y}, \{\hat{w}\}^*)$ for which it is possible to condition on them as they are given positive upper probability, or, in other words, such that $\bar{y} \in \{\bar{w}\}^*_1$.

Section 5.1 will show how some of these differences between CIR and CUR impact on an example.

### 4.1.2 ORIGINALITY OF CIR

Something interesting to note is that CIR, such as traditional updating and CUR, is based only on the variables $Z$ and $Y$: i.e., for applying CIR one does not need to consider the $W$-variables. This makes CIR particularly simple to use. Another consideration concerns originality: to the best of our knowledge, CIR appears here for the first time. There are contributions in the literature similar to CIR for the case of statistical model learning from samples made incomplete by an unknown missingness process (e.g., Manski, 2003; Ramoni & Sebastiani, 2001; Zaffalon, 2002). This is not surprising, as the intuition to take all the completions of an incomplete sample is actually very natural. But we are not aware of any work proposing, and especially deriving, an abstract and general rule to update beliefs from incomplete information such as CIR.

## 5. Applications

In the following sections we shall show how CIR leads easily to several different rules according to the applications under study, and in doing so we present a number of examples. To make things simpler we do not go in the details of the multi-valued maps used; we shall only take $\mathcal{W}$ to be the set of all the nonempty subsets of $\bar{\mathcal{Y}}$, and assume that the hypotheses done throughout the paper hold. We adopt similar considerations for the CAR IP.

### 5.1 CIR for Expert Systems

Probabilistic expert systems represent the key quantities in a domain by a vector of variables, which in this section are assumed to take values from finite spaces. One or more of these variables are the target, i.e., the variables that are the objective of inference. The remaining ones are introduced to the extent of inferring the value of the target.

The Asia network in Figure 1 represents a well-known artificial example of expert system in the medical domain. It is supposed that an expert has provided the graph as well as the probabilities that define the Asia network, as reported in Table 1. If we want to use the network to make diagnosis, we first choose a target node, such as $R$ or $B$. Then we collect information about the patient, as well as the results of medical tests, such as X-rays, and make it available to the network by instantiating the related nodes in the graph to the observed values. Finally, we query the network to update the probability of the target node given the evidence inserted.

But usually not all the nodes, apart from the target node, are instantiated as some values are *missing*, and the way the network has to compute belief updating depends on our assumptions about them. The traditional way entails marginalising them out and it is equivalent to assuming that they are subject to a MAR process. This is the special case of CIR obtained by dropping the unknown IP. The general version of CIR offers more flexibility in their treatment, and can produce a very different solution, as we illustrate by an example below. In it we also use coarsened observations so as to show how set-valued observations may enter the picture of probabilistic inference both in the CIR and in the CAR case.

**Example 3** *Say that you are a senior doctor at a hospital and find on your desk a report from a junior colleague about a certain patient he visited. The patient came to the hospital under a state of dyspnea $(A = a')$. Your colleague visited the patient and did not find any sign of bronchitis $(H = h'')$. He also made some questions to the patient to collect background information. He remembered that he had to ask something about smoking as well as about some recent trip to Asia; but in the end he was not clear nor determinate enough and he got to know only that the patient was concerned with only one of the two*

| | | | | |
|---|---|---|---|---|
| $V = v'$ | 0.01 | | | |
| $K = k'$ | 0.5 | | | |

| | $v'$ | $v''$ | | |
|---|---|---|---|---|
| $B = b'$ | 0.05 | 0.01 | | |

| | $k'$ | $k''$ | | |
|---|---|---|---|---|
| $R = r'$ | 0.1 | 0.01 | | |

| | $k'$ | $k''$ | | |
|---|---|---|---|---|
| $H = h'$ | 0.6 | 0.3 | | |

| | $b'r'$ | $b'r''$ | $b''r'$ | $b''r''$ |
|---|---|---|---|---|
| $O = o'$ | 1 | 1 | 1 | 0 |

| | $o'$ | $o''$ | | |
|---|---|---|---|---|
| $L = l'$ | 0.98 | 0.05 | | |

| | $o'h'$ | $o'h''$ | $o''h'$ | $o''h''$ |
|---|---|---|---|---|
| $A = a'$ | 0.9 | 0.7 | 0.8 | 0.1 |

Table 1: Asia example: probabilities for each variable (first column) in the graph conditional on the values of the parent variables. The state 'prime' corresponds to 'yes'.

*things. Your unexperienced colleague thought then the situation was not serious and sent the patient back home. You are instead a bit more cautious. In fact you suspect that the patient might have hidden some information for privacy reasons, although you recognise that you do not know these reasons. Overall, you are implicitly assuming the existence of an unknown IP for the variables $V$ and $K$ that leads to $\{\bar{w}\}^* = \{(v', k''), (v'', k')\}$, where we denote by $\bar{w}$ the value of $\bar{W}$ that corresponds to the observation of $(V, K)$. In other words you regard $(v', k'')$ and $(v'', k')$ as the only two possible completions for the information that you lack about $(V, K)$.*

*Let us make two remarks before proceeding: (i) that the unknown IP is a coarsening process, not just a missingness process, and this is what allows us to restrict the completions of $(V, K)$ only to two elements; and (ii) that despite the name the unknown IP is (also very) informative when we use coarsened rather than missing values.*

*You finally judge the remaining variables to be subject to a MAR IP. This is very reasonable as your colleague simply decided not to have the patient do any test. In this case the probability of missingness is one independently of the actual values of the variables under consideration. It is instructive to stress that in this case the MAR IP is just your colleague.*

*At this point you are ready to do some diagnosis. You first run an algorithm for Bayesian nets using the completion $(v', k'')$, obtaining that the posterior probability for cancer is 0.052 and that for tuberculosis is 0.258; then you run the algorithm once more, in this case using the completion $(v'', k')$, which leads to 0.423 for the updated probability of cancer and 0.042 for that of tuberculosis. Now the question is that the first run suggests that you should diagnose tuberculosis, and the second that you should diagnose cancer. Since you have no idea about which one is the right completion of $(V, K)$, you admit that you are not able to discriminate between tuberculosis and cancer at this time. In other words, your model is telling you that the information you have is too weak to draw any useful conclusion, and is implicitly suggesting to collect stronger information. In order to do so, you invite the patient for a next visit, explaining the importance of knowing whether he has been to Asia, something that he actually confirms, thus letting you know eventually that $V = v'$ and $K = k''$. Your updated probabilities are then 0.052 for cancer and 0.258 for tuberculosis. On this basis, you ask the patient to undergo X-rays, which turn out to be abnormal ($L = l'$), leading to 0.151 and 0.754, respectively, as probabilities for cancer and tuberculosis, and you diagnose tuberculosis.*

*Consider what might happen by assuming all the variables to be subject to a CAR/MAR IP, as it is common with Bayesian nets. In that case, the information you have initially is that $A = a'$, $H = h''$, and that $(V, K) \in \{(v', k''), (v'', k')\}$. It is just CAR that allows us to forget about $\bar{W}$ and to write down that $(V, K)$ belongs to $\{(v', k''), (v'', k')\}$. This event is easy to incorporate in a Bayesian network: it is enough to insert a new node in the Asia network that is child of both $V$ and $K$ and that is in state 'yes' with probability one if and only if $(V, K) \in \{(v', k''), (v'', k')\}$. This new node is then instantiated in state 'yes'. At this point, running an algorithm for Bayesian nets yields that the probabilities for cancer and tuberculosis conditional on $A = a'$, $H = h''$, and $(V, K) \in \{(v', k''), (v'', k')\}$, are 0.418 and 0.045, respectively. This would make you suspect that there is cancer; moreover, it might well induce you to take the following erroneous course of reasoning: 'since the probability of cancer is that high even without knowing the exact values of $V$ and $K$, trying to obtain that information is a waste of time; I must rather focus on more concrete evidence such*

*as X-rays to make my diagnosis.' But obtaining the positive X-rays test ($L = l'$) would enforce your beliefs even more by raising the probability of cancer up to 0.859, leading you to a mistaken diagnosis.* ◇

These considerations are not limited to expert systems based on precise probability; completely analogous considerations would be done in the case of expert systems that model knowledge using closed convex sets of mass functions (or, equivalently, by the coherent lower prevision which is their lower envelope) for which CIR is also suited. *Credal networks*, for example, provide such modelling capabilities (Cozman, 2000, 2005).

It is easy to rephrase expert system models in the setting of this paper. Say that the expert system is based on the vector of variables $(Z, \bar{Y}_1, \ldots, \bar{Y}_m, \hat{Y}_1, \ldots, \hat{Y}_n)$, where $Z$ is the target variable, and the others are those subject to the unknown and the CAR IP, respectively. Then it is sufficient to write $\bar{Y} := (\bar{Y}_1, \ldots, \bar{Y}_m)$, $\hat{Y} := (\hat{Y}_1, \ldots, \hat{Y}_n)$, and to consider that a set $\mathcal{M}$ of joint mass functions for $(Z, Y)$, or equivalently the lower prevision $\underline{P}_1$ made by taking its lower envelope, is given. Doing inference with an expert system, in quite a general form, corresponds then to compute $\underline{R}(g|\bar{w}, \hat{w})$, where $\bar{w}$ is now the observation of the $\bar{Y}$ vector and $\hat{w}$ that of the $\hat{Y}$ vector.

We consider some cases to make things clearer. At an extreme, which we already mentioned, there is the case $m = 0$, which means that there is only the CAR IP. The updating rule that follows from CIR is then the traditional updating: $\underline{R}(g|\hat{w}) = \underline{R}(g|\{\hat{w}\}^*)$. Say that, to be even more specific, $g$ is the indicator function $\mathbb{I}_z$ of some $z \in \mathcal{Z}$, and that $\{\hat{w}\}^* = \{(\hat{y}_1, \ldots, \hat{y}_j, \hat{y}'_{j+1}, \ldots, \hat{y}'_n) \in \hat{\mathcal{Y}} : \hat{y}'_{j+1} \in \hat{\mathcal{Y}}_{j+1}, \ldots, \hat{y}'_n \in \hat{\mathcal{Y}}_n)\}$, i.e., that the first $j$ variables of $\hat{Y}$ are observed precisely, while the others are missing. The updating rule becomes:

$$\underline{R}(g|\{\hat{w}\}^*) = \underline{R}(\mathbb{I}_z|\hat{y}_1, \ldots, \hat{y}_j, \hat{\mathcal{Y}}_{j+1}, \ldots, \hat{\mathcal{Y}}_n) = \underline{R}(\mathbb{I}_z|\hat{y}_1, \ldots, \hat{y}_j),$$

which is equal to $\inf_{P \geq \underline{P}_1 : P(\hat{y}_1, \ldots, \hat{y}_j) > 0} P(z|\hat{y}_1, \ldots, \hat{y}_j)$. The latter is an updating rule implemented by credal networks; and if $\underline{P}_1$ is a linear prevision, it is the rule used with Bayesian networks.

Consider now the other extreme: $n = 0$, i.e., when there is only the unknown IP. Similarly to the previous case, say that $g$ is the indicator function for $z$; and that $\{\bar{w}\}^* = \{(\bar{y}_1, \ldots, \bar{y}_i, \bar{y}'_{i+1}, \ldots, \bar{y}'_m) \in \bar{\mathcal{Y}} : \bar{y}'_{i+1} \in \bar{\mathcal{Y}}_{i+1}, \ldots, \bar{y}'_m \in \bar{\mathcal{Y}}_m)\}$. CIR becomes then $\underline{R}(g|\bar{w}) = \min_{\bar{y}_{i+1} \in \bar{\mathcal{Y}}_{i+1}, \ldots, \bar{y}_m \in \bar{\mathcal{Y}}_m} \inf_{P \geq \underline{P}_1 : P(\bar{y}_1, \ldots, \bar{y}_m) > 0} P(z|\bar{y}_1, \ldots, \bar{y}_m)$. This case nearly coincides with the conservative updating rule proposed by De Cooman and Zaffalon (2004), with the differences already discussed in Section 4.1. These differences are important: for instance, that CUR cannot deal with coarsened observations makes it impossible to use it to model the observation in Example 3. Some of them would even prevent CUR from being applied more generally to the example: on the one hand, the presence of the logical 'or' gate in the Asia network creates states of zero upper probability (e.g., $B = b'$, $R = r'$, $O = o''$) that are incompatible with the assumptions underlying CUR; on the other, since domain knowledge (i.e., the Asia net) is built out of a number of conditional mass functions, we cannot guarantee that CUR leads to self-consistent inference as its proof does not deal with such an extended case. We know that CIR instead does lead to self-consistent inference as a consequence of Corollary 3 in Appendix B.

When neither $m$ nor $n$ are equal to zero, CIR becomes a mix of the two extreme cases just illustrated, as in the example: it leads to treat the variables subject to the CAR IP by the traditional updating, while treating those subject to the unknown IP similarly to what CUR does. Mixing the two things has the advantage of greater expressivity: it allows us to represent beliefs about the overall IP that are in between the two extremes.

We conclude this section by briefly discussing the problem of doing computations with CIR and its special case CUR.

In the context of doing classification with Bayesian nets according to CUR, Antonucci and Zaffalon (2007) have proved an NP-hardness result, as well as given an exact algorithm. The algorithm is a variant of the variable elimination algorithm that has a better complexity than the traditional algorithms on Bayesian nets (implicitly using CAR): it works in linear time not only on *polytree* networks (i.e., those for which there is at most one path between any two nodes in the graph after dropping the orientation of the arcs) but also on a number of more general nets; on the remaining ones it takes exponential time. Moreover, in another paper, Antonucci and Zaffalon (2006) have shown that when there are missing observations the problem of CIR-updating in Bayesian nets and that of traditional (i.e., MAR-based) updating in credal nets are equivalent. This is exploited in a recent paper (Antonucci & Zaffalon, 2008, Section 9) to show the NP-hardness of CIR-updating in Bayesian nets, and to give a procedure to solve such a problem via existing algorithms for credal nets. The idea behind such a procedure is relatively straightforward, and it takes inspiration from a method proposed by Pearl (1988, Section 4.3) to represent an instantiated node in a Bayesian net using an equivalent formulation: the formulation is based on removing the instantiation from the node and on adding a dummy child to it that is actually instantiated to a certain value. This approach is basically taken as it is for CIR-updating in the case of a node missing in an unknown way; CIR imposes to deal with it by considering all its possible instantiations. Considering the dummy-child method for each of them is shown to be equivalent to using a single imprecise-probability dummy child that is instantiated to a unique value. In this way the node that was originally missing in an unknown way can be treated as a node missing at random, while the dummy child is simply an instantiated imprecise-probability node. As a consequence, the original Bayesian net becomes a credal network and our task becomes the computation of MAR-based updating on such a net. Since there are already many algorithms designed for this task, the described transformation allows one to solve the CIR-updating problem via known algorithms for the MAR-updating problem on credal nets. Yet, MAR-updating on credal networks is a more difficult problem than MAR-updating on Bayesian nets, as it is NP-hard also on polytrees (De Campos & Cozman, 2005). Therefore CIR-updating appears to be more demanding than MAR- or CUR-updating. This is not necessarily going to be a problem in practice as approximate algorithms for credal nets nowadays allow one to solve updating problems on large-scale networks (e.g., Antonucci et al., 2008).

## 5.2 CIR for Parametric Statistical Inference

In a statistical problem of *parametric inference*, we are given a sample that we use to update our beliefs about a so-called *parameter*, say $\Theta$, with values $\theta \in \mathcal{T}$. The admissible values for $\Theta$ index a family of data generation models that we consider possible for the problem under consideration. Here we do not exclude the possibility of this set being infinite, as it is often the case in statistics.

In this setting, a fact is thus the true, and hence complete, sample. We represent it as the following vector of variables:

$$Y := \begin{bmatrix} D_1 \\ D_2 \\ \vdots \\ D_N \end{bmatrix}. \tag{1}$$

The elements of a sample are also called *units* of data. We assume from now on, that the variables $D_i$ have the same space of possibilities for every $i$.

We can exemplify the problem of parametric statistical inference by focusing once again on the Asia Bayesian network in Figure 1: a frequent step in the construction of Bayesian nets is the inference of the network parameters from a data set. Say, for instance, that we are interested in evaluating the chance $\Theta$ that a person suffers from tuberculosis if we know that the same person has made a recent visit to Asia. To this extent we might exploit a data set for the variables $B$ and $V$, such as the one below:

$$y := \begin{bmatrix} (b', v') \\ (b', v'') \\ (b', v'') \\ (b'', v'') \\ (b'', v') \\ (b'', v') \end{bmatrix}, \tag{2}$$

for which we assume in addition that the data have been generated according to an *identical and independently distributed* (IID) process.

We can do parametric inference relying on the tools of Bayesian statistics. This means to compute the *posterior* linear prevision (i.e., the posterior expectation) $P(g|y)$, for a certain function $g : \mathcal{T} \to \mathbb{R}$. This is obtained integrating the posterior *density function* for $\Theta$, obtained in turn by Bayes' rule applied to a prior density function, and the *likelihood function* (or *likelihood*) $L(\theta) := P(y|\theta)$.

In this case, a common choice for the prior would be a *Beta* density for $\Theta$: $B_{s,t'}(\theta) \propto \theta^{st'-1}(1-\theta)^{s(1-t')-1}$, where $s$ and $t'$ are positive real hyper-parameters that are respectively the overall strength of the prior knowledge and the prior expectation of $\Theta$. In the case of the data set (2), this choice leads to a posterior expectation for $\Theta$ that is equal to $\frac{1+st'}{3+s}$. Setting $s := 1$ and $t' := 1/2$, namely, Perks' prior (see Perks, 1947), we obtain $\frac{1+st'}{3+s} = 0.375$, which can be regarded as an approximation to the true parameter value.

Imprecise probability approaches to parametric inference often extend Bayesian statistical methods by working with a *set* of prior density functions, and by updating each of them by Bayes' rule, whenever possible, using the same likelihood function, thus obtaining a set of posteriors. An example is the *imprecise Dirichlet model* proposed by Walley (1996a). In our notation and terminology, this means that in the imprecise case parametric inference is based on the unconditional lower prevision $\underline{P}_1(\Theta, Y)$, obtained by means of the prior $\underline{P}_1(\Theta)$ and the likelihood $\underline{P}(Y|\Theta)$ through a rule called *marginal extension*, that is updated into the posterior lower prevision $\underline{R}(\Theta|Y)$ by a procedure called *regular extension*. Marginal and regular extension will be introduced more precisely in Section 6.

In the previous example related to the data set (2) where the variables are binary, the imprecise Dirichlet model is called *imprecise Beta model* (Bernard, 1996), and can be easily applied as follows. The idea is to consider the set of all the Beta densities with fixed prior strength, say $s = 1$: i.e., the set of all $B_{s,t'}$ such that $s = 1$ and $t' \in (0, 1)$.[10] This set can be regarded as a single *imprecise* prior that only states that the prior probability for category $b'$ lies in $(0, 1)$, which seems a more reasonable way to model a state of prior ignorance about $\Theta$ than a Bayesian prior, such as Perks' above. This imprecise prior leads to an imprecise posterior expectation for $\Theta$: to compute it, it is enough to reconsider the expression $\frac{1+st'}{3+s}$ and let $t'$ take all the values in $(0, 1)$; this leads to the interval $[0.25, 0.50]$ that is delimited by the lower and the upper posterior probabilities obtained from $\frac{1+st'}{3+s}$ when $t' = 0$ and $t' = 1$, respectively. Again, this 'interval estimate'[11] appears to be much more reasonable and reliable than the precise posterior expectation obtained by the Bayesian approach, taking especially into account that the sample available to infer the parameter under consideration has size three!

But in real problems we often face the further complication of having to deal with incompleteness in statistical data: rather than a complete sample $y$, we may be given an *incomplete* sample $w$, i.e., one that corresponds to a set $\{w\}^*$ of complete samples. The set $\{w\}^*$ arises as a consequence of missingness or partial observability of some values. The data set (2) might for instance be turned by an incompleteness process into the following incomplete data set:

$$w := \begin{bmatrix} (b', v') \\ (b', v'') \\ (b', ?) \\ (b'', v'') \\ (b'', ?) \\ (?, v') \end{bmatrix}. \tag{3}$$

In this example $\{w\}^*$ is the set of the eight complete data sets that are obtained by replacing the question marks in all the possible ways.

The question is then how to do parametric inference with an incomplete sample. We represent each unit in Vector (1) by regarding the generic variable $D_i$ as the pair $(\bar{Y}_i, \hat{Y}_i)$, and let $\bar{Y} := (\bar{Y}_1, \ldots, \bar{Y}_N)$, $\hat{Y} := (\hat{Y}_1, \ldots, \hat{Y}_N)$, so that $Y = (\bar{Y}, \hat{Y})$.

Now it is easy to use CIR to address the question of parametric inference with incomplete data, if we know that $\hat{Y}$ is subject to a CAR IP, and we do not know the IP that acts on $\bar{Y}$. Observing that $\Theta$ plays the role of $Z$, we obtain that CIR prescribes using the following rule: $\underline{R}(g|w) = \min_{\bar{y} \in \{\bar{w}\}_1^*} \underline{R}(g|\bar{y}, \{\hat{w}\}^*)$, which rewrites also as

$$\underline{R}(g|\bar{w}, \hat{w}) = \min_{\bar{y} \in \{\bar{w}\}_1^*} \quad \inf_{P \geq \underline{P}_1 : P(\bar{y}, \{\hat{w}\}^*) > 0} P(g|\bar{y}, \{\hat{w}\}^*), \tag{4}$$

---

10. The reason why we exclude the extreme points of the interval is to avoid creating conditioning events with zero lower probability and have to discuss their implications. In the present setup this is not restrictive, as it can be shown that the closed interval would eventually lead to the same inferences.

11. It is important to be aware that these intervals are conceptually very different tools from Bayesian credible intervals, which arise out of second-order information about a chance. The present intervals arise only out of ignorance about the unknown IP, something that makes a set of IPs be consistent with our assumptions and about which *we have no second-order information*. Stated differently, the counterpart of these intervals in the precise case are the point estimates, *not* the credible intervals, which should instead be compared with imprecise credible intervals (e.g., see Walley, 1996a, Section 3).

because imprecision on $(\Theta, Y)$ originates via prior imprecision only. We prove in Corollary 4 in Appendix B that this rule leads to self-consistent inference.

We can regard the lower expectation in Equation (4) as arising from two kinds of imprecise beliefs. The first are beliefs about $\Theta$ that we model by the lower prevision $\underline{P}_1(\Theta)$. The remaining beliefs are embodied by the likelihood function. This is imprecise knowledge, too, since there are actually *multiple* likelihood functions, because of the unknown IP, and we model them using the different conditional previsions $P(\Theta|\bar{y}, \{\hat{w}\}^*)$ for $\bar{y} \in \{\bar{w}\}_1^*$. In other words, working with incomplete data according to CIR is equivalent to working with the set of posteriors that arise by applying Bayes' rule, whenever possible, to each prior-likelihood pair in the model.

In the case of the data set (2), we might know that in order to create the incomplete sample (3) out of it, the variable $B$ has been subject to a MAR missingness process, while we might not know what kind of missingness process has acted on variable $V$; in this case, the generic unit $i$ of (2) can be written as the pair $(\hat{y}_i, \bar{y}_i)$. Say that we focus again on computing the posterior expectation for $\Theta$, chosen to be that a person suffers from tuberculosis if we know that the same person has made a recent visit to Asia. Say also that we use Perks' prior as we did before in the Bayesian case, in order not to get distracted in the discussion by prior imprecision. In this case, the outer minimum in (4) corresponds to minimising over the four completions for variable $V$; the following infimum corresponds to take into account Perks' prior. The four completions mentioned above give rise to four different data sets that contain a single missing value in the last unit. For each of them, we can compute the lower probability for the category $b'$ conditional on $v'$, using the *Expectation-Maximisation* algorithm (Dempster, Laird, & Rubin, 1977) together with Perk's prior, obtaining four values: 0.63, 0.87, 0.50, 0.83 (these values are rounded to the second digit). The lower probability is then their minimum: 0.5; analogously, their maximum is the upper probability: 0.87. We can again think of these two values as delimiting an interval for the probability under consideration: $[0.5, 0.87]$. The width of this interval reflects our ignorance about the missingness process for $V$.

We consider now the Bayesian case, using for both $B$ and $V$ the MAR assumption, as it is very common when learning parameters from data sets. In this case, the Expectation-Maximisation algorithm together with Perks' prior yields an estimate for the probability equal to 0.86. Apart from the unrealistic precision of this estimate obtained from such a small data set, we can observe that using MAR has arbitrarily led us very close to the upper extreme of the interval $[0.5, 0.87]$. This is even more questionable when we consider that the Bayesian estimate from the complete data set (2) was close to the lower extreme.

In summary, by using CIR we are able to obtain interval estimates that are arguably more realistic than the point estimates provided by more traditional methods when the MAR assumption is not justified. This also follows by exploiting the option given by CIR to use models of prior knowledge that carefully model the available information, or the absence of information, such as the imprecise Dirichlet model.

Finally, as the example illustrates, working with multiple likelihoods is a consequence of taking all the possible completions of the incomplete part of the sample subject to the unknown IP. This is a very intuitive procedure; therefore, it is not surprising that analogous procedures have already been advocated with missing data in the context of robust statistical inference (see, e.g., Manski, 2003; Ramoni & Sebastiani, 2001; Zaffalon, 2002). Actually, the discussion in this (and the following) section can be regarded as a formal justification of the cited approaches from the point of view of the theory of coherent lower previsions. It can also be regarded as a generalisation of some of those approaches in that it considers the joint presence of the CAR and the unknown IP, and because it allows one to work with incomplete rather than missing data.

### 5.2.1 IID+ID CASE

In the previous section, we have not made hypotheses about the generation of the complete sample. However, in practice it is very frequent to deal with independent and identically distributed data (also called *multinomial data*), and indeed we have already assumed IID for the data in (2). With multinomial data, the units are identically distributed conditional on $\theta$, and this helps simplifying the developments. Call $\mathcal{D}$ the space of possibilities common to the units. When we assume that $\mathcal{D}$ is finite (as we shall do in our derivation of the CIR rule in Section 7.3), the parameter $\Theta$ is defined as the vector $[\Theta_d]_{d \in \mathcal{D}}$, where the generic element $\Theta_d$ represents the aleatory probability of $D = d$. It follows that $\theta \in \mathcal{T}$ is the vector $[\theta_d]_{d \in \mathcal{D}}$, whose generic element $\theta_d$ is $P(d|\theta)$; and that $\mathcal{T}$ is a subset of the $|\mathcal{D}|$-dimensional unit simplex. Taking into account that units are also independent conditional on $\theta$, it follows that the likelihood factorises as $\prod_{i=1}^{N} \theta_{d_i}$.

When complete data are IID, it may be reasonable to assume that also the overall IP is *independently distributed* (ID). Such an assumption allows us to represent the observation of the complete sample as a vector, i.e., as an *incomplete* sample:

$$W := \begin{bmatrix} W_1 \\ W_2 \\ \vdots \\ W_N \end{bmatrix}.$$

Here the generic unit $W_i$ is the observation of $D_i$. Remember that we regard $D_i$ as the pair $(\bar{Y}_i, \hat{Y}_i)$; as a consequence, we also regard $W_i$ as the pair of variables $(\bar{W}_i, \hat{W}_i)$, with $\bar{W}_i$ and $\hat{W}_i$ the observations of $\bar{Y}_i$ and $\hat{Y}_i$, respectively. Finally, let $\bar{W} := (\bar{W}_1, \bar{W}_2, \ldots, \bar{W}_N)$ and $\hat{W} := (\hat{W}_1, \hat{W}_2, \ldots, \hat{W}_N)$. For an example it is sufficient to consider the data set in (3), which is just an instance of $W$ in the current language; and its generic unit $i$ is an instance of $W_i = (\bar{W}_i, \hat{W}_i)$, where $\bar{W}_i$ corresponds to variable $V$ and $\hat{W}_i$ to $B$.

Using the newly introduced notation, the form that CIR takes is the following (with obvious meaning of the symbols):

$$\underline{R}(g|\bar{w}, \hat{w}) = \min_{(\bar{y}_1, \ldots, \bar{y}_N) \in \{\bar{w}\}_1^*} \underline{R}(g|\bar{y}_1, \ldots, \bar{y}_N, \{\hat{w}_1\}^*, \ldots, \{\hat{w}_N\}^*), \tag{5}$$

where $\underline{R}(g|\bar{y}_1, \ldots, \bar{y}_N, \{\hat{w}_1\}^*, \ldots, \{\hat{w}_N\}^*)$ is equal to

$$\inf_{P \geq \underline{P}_1, P(\bar{y}_1, \ldots, \bar{y}_N, \{\hat{w}_1\}^*, \ldots, \{\hat{w}_N\}^*) > 0} P(g|\bar{y}_1, \ldots, \bar{y}_N, \{\hat{w}_1\}^*, \ldots, \{\hat{w}_N\}^*).$$

That this rule leads to self-consistent inference is established in Corollary 4 in Appendix B. Moreover, the new formulation shows that the ID assumption for the incompleteness process leads in practice to inhibiting certain types of coarsening: those that create logical

'connections' between different units. On the one hand, this confirms the expressive power of coarsening. On the other, it suggests that in applications it might be easier to forget about the ID assumption and simply focus on the specification of the kind of coarsening behaviour. This approach appears to be more intuitive and hence more accessible especially to people with little experience in statistics that might be involved in the analysis.

### 5.2.2 Full IID Case?

Is it reasonable to assume that an IP is *identically* distributed besides independently distributed? And what are the consequences of such an assumption?

We start by addressing the second question. Under IID for complete data, the generic element $d \in \mathcal{D}$ has a fixed aleatory probability, which we denoted by $\Theta_d$. Call $\mathcal{W}$ the space of possibilities common to the variables $W_i$, $i = 1, \ldots, N$. Under IID for the IP, the generic element $w$ of $\mathcal{W}$ has a fixed aleatory probability to be produced conditional on $d$, let us call it $\Phi_w^d$. It follows that $w$ has a fixed *unconditional* aleatory probability to be produced: $\Psi_w := \sum_{d \in \mathcal{D}} \Theta_d \cdot \Phi_w^d$, which means that the process that produces elements of $\mathcal{W}$ is also multinomial. This seems to considerably simplify the setup considered so far, and can thus be regarded as an advantage of the full IID assumption (this advantage has clear practical consequences in the case of pattern classification, as illustrated in Section 5.3.2.)

On the other hand, and here we turn to address the first question, the possible advantages seem to be dwarfed by the strength of the assumption itself. It is indeed questionable that such an assumption may be largely valid in applications. IPs are often generated by sophisticated behaviour patterns that involve humans, or other complex agents, and that can be regarded as protocols of communications. In this case giving or not giving some information is a fundamental and an intelligent part of the communication. Assuming in this case that the IP is IID, seems to be too much of a strong assumption. In fact, there is a fundamental difference between data-generating processes (giving rise to $Z$ and $Y$) and observational processes (giving rise to $W$). We believe that the essence of this difference is that in a number of cases it does not make much sense to assume that an observational process is identically distributed, while IID does make sense, or is at least a reasonable approximation for many data-generating processes.

### 5.2.3 CAR Units

Having considered the IID+ID setup in Section 5.2.1 gives us the opportunity to slightly extend the analysis done so far. The idea is that the since the *overall* IP does not need to be identically distributed, it may happen to act as a CAR process for *some* units. Say that a certain unit 0 is *entirely* coarsened at random. The question is what form takes CIR now.

This is easy to see by defining $\bar{Y} := (\bar{Y}_1, \ldots, \bar{Y}_N)$, $\hat{Y} := (D_0, \hat{Y}_1, \ldots, \hat{Y}_N)$, and the corresponding observation variables $\bar{W} := (\bar{W}_1, \ldots, \bar{W}_N)$, $\hat{W} := (W_0, \hat{W}_1, \ldots, \hat{W}_N)$, which lead to the following rule:

$$\underline{R}(g|\bar{w}, \hat{w}) = \min_{(\bar{y}_1, \ldots, \bar{y}_N) \in \{\bar{w}\}_1^*} \underline{R}(g|\bar{y}_1, \ldots, \bar{y}_N, \{w_0\}^*, \{\hat{w}_1\}^*, \ldots, \{\hat{w}_N\}^*).$$

In particular, if Unit 0 is missing, i.e., $\{w_0\}^* = \mathcal{D}$, the rule prescribes to discard such a unit from consideration:

$$\underline{R}(g|\bar{w}, \hat{w}) = \min_{(\bar{y}_1,\dots,\bar{y}_N) \in \{\bar{w}\}_1^*} \underline{R}(g|\bar{y}_1,\dots,\bar{y}_N, \{\hat{w}_1\}^*,\dots,\{\hat{w}_N\}^*).$$

This observation is particularly important for applications because it may well be the case that in practice some units are entirely missing in a non-selective way. CIR tells us that all such units, as Unit 0 above, are not going to alter our beliefs about $Z$.

The situation is obviously different if some units are turned into missing units by an unknown process; in this case there is no justification to discard them. The correct way to proceed is to make those missing values part of the observed data, and then to apply CIR again. This would lead us to consider all the completions of such missing units into account.

## 5.3 CIR for Pattern Classification

This section focuses on problems of *pattern classification*, a special case of so-called *predictive inference*. Loosely speaking, this kind of inference is concerned with predicting future elements of a sequence based on the available part of the sequence itself. In a classification problem, the available sequence is represented by the following matrix:

$$\begin{bmatrix} C_1 & F_1 \\ C_2 & F_2 \\ \vdots & \vdots \\ C_N & F_N \end{bmatrix}. \tag{6}$$

The generic line $i$, or unit $i$, of the matrix represents an object described by a pair of variables that we relabel as $D_i := (C_i, F_i)$ to be consistent with the notation used for parametric inference. Variable $C_i$ represents a characteristic of the object in which we are interested, and that we call *class*. Denote the set of classes by $\mathcal{C}$. By definition of pattern classification, $\mathcal{C}$ is a finite set. Variable $F_i$ represents some features of the object that are informative about the class.

We can try to make this more clear by an example focused again on the Asia network. Say that we have a data set that records the state of the variables in the Asia network for a number of persons. Say also that we are interested in classifying people as smokers or nonsmokers. In this case, the value of $C_i$, i.e., the class variable for the i-th person in the data set, is the state of variable $K$ in the Asia network for person $i$. Variable $F_i$ is then the vector of all the remaining variables in the Asia network for such a person; in other words, $F_i$ represents the profile of person $i$ in the data set.

Now we consider the next element in the sequence, represented as the further unit $D_{N+1} := (C_{N+1}, F_{N+1})$, also called the *unit to classify*. In the previous example, $D_{N+1}$ would be the next person we see. The goal of classification is to predict the value that $C_{N+1}$ assumes given values of all the other variables; in the example this amounts to predict whether the next person is smoker or not given the profile of that person, and the relationship between profiles and classes that is suggested by the historical data. In other words, in this framework $C_{N+1}$ is $Z$ and $(D_1,\dots,D_N, F_{N+1})$ is $Y$.

Predicting the class $c'$ for $C_{N+1}$ can be regarded as an action with uncertain reward $g_{c'}$, whose value $g_{c'}(c)$ depends on the value $c$ that $C_{N+1}$ actually assumes. In the case of precise probability, the *optimal prediction* is a class $c_{\text{opt}}$ that maximises the expected reward:

$$P(g_{c_{\text{opt}}}|d_1, \ldots, d_N, f_{N+1}) \geq P(g_{c'}|d_1, \ldots, d_N, f_{N+1}), \quad c' \in \mathcal{C}.$$

The previous expression is equivalent to the next:

$$P(g_{c_{\text{opt}}} - g_{c'}|d_1, \ldots, d_N, f_{N+1}) \geq 0, \quad c' \in \mathcal{C}.$$

This is very similar to the expression used with imprecise probability: in this case an optimal class is one that is *undominated*, i.e., such that for all $c' \in \mathcal{C}$:[12]

$$\overline{P}(g_{c_{\text{opt}}} - g_{c'}|d_1, \ldots, d_N, f_{N+1}) > 0.$$

Despite the similarity in notation, it is important to realise that there is a fundamental difference in the two cases: precise probability always leads to a determinate prediction, imprecise probability does not. With imprecise probability there is a progression according to which stronger beliefs lead to less indeterminate decisions, up to completely determined ones (see the paper by De Cooman & Zaffalon, 2004, Section 2.10, for a wider discussion of decision making with imprecise probability).

The discussion up to this point has highlighted that both with precise and with imprecise probabilities, predictions are based on (upper, and hence lower) previsions of functions $g : \mathcal{C} \to \mathbb{R}$. In order to address the issue of incompleteness in pattern classification, we can therefore, without loss of generality, restrict the attention to the task of updating beliefs about $C_{N+1}$.

This task is usually regarded as made of two distinct steps. The first is concerned with learning from a sample, represented here by Matrix (6), a probabilistic model of the relationship between features and classes. The second is applying the model to the observed value $f_{N+1}$ of $F_{N+1}$ in order to finally compute the (lower) prevision of a gamble $g$.

We illustrate the two steps in the case of IID data; more general cases can be treated analogously.

In the Bayesian framework, the first step is often done in a way similar to parametric inference. One defines a family of density functions that may be responsible for producing the units of data, indexes them by the admissible values $\theta \in \mathcal{T}$ of a (usually continuous) parameter $\Theta$, and then assesses a prior density for $\Theta$. The prior and the likelihood lead via Bayes' rule to the posterior density for $\Theta$ conditional on $(d_1, \ldots, d_N)$. The second step corresponds to use such a posterior together with the density (or the mass function) for $D_{N+1}$ conditional on $\Theta$, to obtain (integrating $\Theta$ out) the so-called *predictive* posterior: i.e., the mass function for $C_{N+1}$ conditional on $(d_1, \ldots, d_N, f_{N+1})$, which embodies the wanted probabilistic model of the relationship between features and classes. Computing the posterior prevision $P(g|d_1, \ldots, d_N, f_{N+1})$ is trivial at that point.

The situation is similar in the imprecise case, with the difference that using a set of prior densities, i.e., a lower prior $\underline{P}_1(\Theta)$, leads to a lower predictive prevision:

$$\underline{P}(g|d_1, \ldots, d_N, f_{N+1}) := \inf_{P \geq \underline{P}_1 : P(d_1, \ldots, d_N, f_{N+1}) > 0} P(g|d_1, \ldots, d_N, f_{N+1}).$$

---

12. Walley (1991) calls *maximality* the related decision criterion. See the work of Troffaes (2007) for a comparison with other criteria.

Now we are ready to address the issue of incomplete data using CIR. We must first define the variables that are subject to the CAR and the unknown IP. With respect to feature variables, we assume that the generic variable $F_i$ $(i = 1, \ldots, N + 1)$ is equal to the pair $(\bar{F}_i, \hat{F}_i)$, with the usual meaning of the symbols. Define the vectors $\bar{Y}$ and $\hat{Y}$ as $\bar{Y} := (C_1, \bar{F}_1, \ldots, C_N, \bar{F}_N, \bar{F}_{N+1})$, $\hat{Y} := (\hat{F}_1, \ldots, \hat{F}_N, \hat{F}_{N+1})$. To make things more readable, we also define $\bar{Y}_i := (C_i, \bar{F}_i)$, $\hat{Y}_i := \hat{F}_i$, for all $i = 1, \ldots, N$; and $\bar{Y}_{N+1} := \bar{F}_{N+1}$, $\hat{Y}_{N+1} := \hat{F}_{N+1}$.

There is obviously arbitrariness in the above definitions of the vectors $\bar{Y}$ and $\hat{Y}$, and it might well be the case that some applications require different choices. The present definitions are only done for illustrative purposes.

With the above definitions, CIR leads to the following rule to update beliefs about $g$:

$$\underline{R}(g|\bar{w}, \hat{w}) = \min_{(\bar{y}_1, \ldots, \bar{y}_{N+1}) \in \{\bar{w}\}_1^*} \inf_{P \geq \underline{P}_1 : P(\bar{y}_1, \ldots, \bar{y}_{N+1}, \{\hat{w}_1\}^*, \ldots, \{\hat{w}_{N+1}\}^*) > 0}$$
$$P(g|\bar{y}_1, \ldots, \bar{y}_{N+1}, \{\hat{w}_1\}^*, \ldots, \{\hat{w}_{N+1}\}^*). \tag{7}$$

Corollary 5 proves that this rule leads to self-consistent inference. Equation (7) states that one should consider (i) the set of possible completions of the part of the observed sample originated by the unknown IP, $\times_{i=1}^{N} \{\bar{w}_i\}_1^*$; (ii) the set of possible completions of the part of the unit to classify originated by the unknown IP, $\{\bar{w}_{N+1}\}_1^*$; and (iii) the set of (possible) precise priors, which are those dominating our prior lower prevision $\underline{P}_1(\Theta)$. For each of the above choices, the problem becomes the computation of the posterior expectation of $g$ as in a single Bayesian (i.e., precise-probability) classifier. When we consider all the possible choices in (i)–(iii) above, working with (7) amounts to work with a set of Bayesian classifiers. The set of posterior expectations originated by the Bayesian classifiers is summarised in (7) as a lower expectation.

We can look at the problem also from another point of view. Imprecision that produces the lower expectation (7) can be regarded as originated by three components. The first is the presence of multiple priors. The second is the presence of multiple likelihoods, each of which is consistent with a certain completion of the part of the sample originated by the unknown IP. The third component is the set-valued observation of the part $\bar{F}_{N+1}$ of the features of the unit to classify, which we can regard as an imprecise description of the object to classify.

### 5.3.1 An Example

Let us focus again on the initial example in Section 5.3 in which we were interested in classifying people as either smokers or nonsmokers. To make things easier, we take $g_{c'}$ to be the 0-1 loss function, i.e., we compute posterior probabilities rather than expectations in order to issue a classification. Moreover, we take the graph of the Asia net to represent the probabilistic dependencies between the variables involved (yet, we do not assume the net parameters to be given, as learning the parameters from data is part of the inferential task of classification that we have to solve). To make the example more handy, we assume that for all the people we want to classify we miss information about the variables $V$, $B$, $O$, $L$ and $A$ in the Asia net, and that this missingness is MAR: that is, these variables constitute $\hat{F}_{N+1}$. This implies, through (7), that we can actually discard variables $V$, $B$, $O$, $L$ and $A$ from consideration: we can equivalently work out our inference task by relying only on a learning set for the variables $K$, $R$ and $H$ alone. This holds because working with (7) amounts to work with a set of Bayesian classifiers, and the above property holds for each of them.[13] As a consequence, it turns out that the only portion of the Asia net which is relevant for the classification is that displayed in Figure 2. Such a structure, with the feature variables represented as disconnected children of the class variable is called a *naive* network. When we use it as a precise-probability classifier, it is best known as *naive Bayes classifier* (Duda & Hart, 1973).

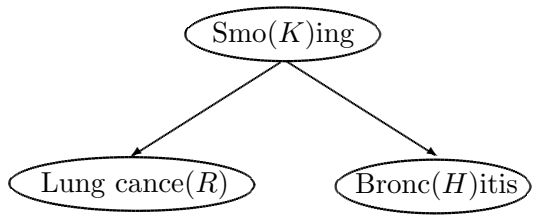

Figure 2: The naive sub-network of the Asia net related to the variables $K$, $R$ and $H$.

Say that our learning set, originated by an IID process, is the following:

$$\bar{d} := \begin{bmatrix} (r', h'', k') \\ (r', h'', k') \\ (r', h', k') \\ (r', h', k') \\ (r'', h', k') \\ (r'', h', k'') \\ (r'', h', k'') \\ (r'', h'', k'') \\ (r'', h'', k'') \\ (r', h'', k'') \end{bmatrix} .$$

This data set is characterised by a strong positive relationship between smoking and cancer and a weak positive relationship between smoking and bronchitis. Learning naive Bayes from $\bar{d}$ and applying the model to the four joint instances of the feature variables, we obtain the following predictions:

$$\begin{bmatrix} (r', h', k') \\ (r', h'', k') \\ (r'', h', k'') \\ (r'', h'', k'') \end{bmatrix} , \tag{8}$$

where we have used Perks' prior for naive Bayes[14] in the above experiments. If we replace such a prior with an imprecise one, we obtain an extension of naive Bayes to imprecise probability. If we moreover define such a prior $\underline{P}_1(\Theta)$ so as to model a state of prior ignorance, similarly to what we have done in the case of parametric inference, we obtain an imprecise-probability classifier that is an instance of the so-called *naive credal classifier 2* (or NCC2, see Corani & Zaffalon, 2008). We can reinterpret NCC2 as a set of naive Bayes classifiers. In the present case, we have a classifier per each precise prior consistent with (i.e., dominating) the imprecise one.

---

13. A proof in relationship with a specific classifier can be found in the work by Corani and Zaffalon (2008, Section 3.1.2).

14. Modified as described by Zaffalon (2001, Section 5.2) (refer to Perks' prior defined 'in analogy with the IDM' in that section) to make the comparison easier.

The way NCC2 issues a classification on a specific instance of $R$ and $H$ can be understood easily in terms of the set of naive Bayes classifiers to which it corresponds: the classification of NCC2 is the union of the classes issued by all the naive Bayes classifiers in the set.[15] This means that if all the naive Bayes classifiers issue the same class, then the classification of NCC2 is *determinate*, i.e., made of a single class. Otherwise, when there is disagreement among the naive Bayes classifiers, then NCC2 issues the entire set $\mathcal{K}$, i.e., an *indeterminate* classification.[16]

If we infer NCC2 from the learning set $\bar{d}$, we obtain again the determinate predictions reported in (8):[17] this means that the information in the data is strong enough to smooth the contribution of each precise prior used by NCC2 in favor of a single class.

We can move to more interesting examples by introducing missing values. Consider the following two instances to classify that are missing the information about cancer:

$$\left[ \begin{array}{c} (?, h') \\ (?, h'') \end{array} \right]. \tag{9}$$

The treatment of these instances will be very different from naive Bayes to NCC2. NCC2 can naturally embed our assumption that $R$ and $H$ are made missing by an unknown process, and deal with it by considering the two completions of the question mark in each instance. In other words, the classification of NCC2 will be not only, as before, equal to the union of the classes delivered by the equivalent set of naive Bayes classifiers, but the union will also be taken with respect to the two completions of the instance to classify. On the other hand, naive Bayes will have to assume MAR to deal with the incomplete instances, and this will lead it to discard $R$ from the conditioning variables.

The classifications issued by naive Bayes and NCC2 are respectively:

$$\left[ \begin{array}{c} (?, h', k') \\ (?, h'', k'') \end{array} \right], \left[ \begin{array}{c} (?, h', \mathcal{K}) \\ (?, h'', \mathcal{K}) \end{array} \right].$$

In other words, for naive Bayes knowing the state of bronchitis is sufficient to determine whether a person is smoker or not, despite the weak relationship in the learning set between those two characteristics; NCC2 is more cautious and believes that it is not possible to determine the smoking state because it is missing, not necessarily in an ignorable way, the value of a very important predictor variable. On the other hand, it can be checked that if the missing variable is $H$ instead of $R$, both naive Bayes and NCC2 agree that the prediction should be determinate: knowing that a person has cancer tells us that the person is a smoker, and vice versa.

---

15. This is the case because the class variable is binary.

16. This total indeterminacy is a consequence of the class variable being binary; in other cases the output set of classes can be any subset of classes, not only the least informative one. In these cases, the classification is only partially indeterminate, and carries therefore useful information.

17. The predictions of NCC2 can be computed by an open-source software freely available at the address http://www.idsia.ch/∼giorgio/jncc2.html.

The situation can become even more critical for naive Bayes if we use an incomplete learning set $\bar{w}_{\bar{d}} := (\bar{w}_1, \ldots, \bar{w}_N)$ rather than $\bar{d}$, such as

$$\bar{w}_{\bar{d}} := \left[ \begin{array}{c} (r', h'', k') \\ (r', h'', k') \\ (r', ?, k') \\ (r', ?, k') \\ (r'', ?, k') \\ (r'', h', k'') \\ (r'', h', k'') \\ (r'', ?, k'') \\ (r'', ?, k'') \\ (r', ?, k'') \end{array} \right].$$

We have originated this data set from $\bar{d}$ by turning some values of $H$ into missing so that the complete portion of the data set suggests that $H$ is a very good predictor for $K$. This is going to be a problem for naive Bayes, because it has still to deal with the incompleteness assuming MAR, which leads it to compute the counts from the learning set discarding the missing values. NCC2, on the other hand, implicitly considers all the complete data sets that are consistent with $\bar{w}_{\bar{d}}$. Therefore in this case NCC2 can be regarded as the set of naive Bayes classifiers that are obtained by considering all the precise priors previously described in relationship with NCC2 *and* all the likelihoods that arise from the complete data sets consistent with $\bar{w}_{\bar{d}}$.

Running both classifiers to predict the class of the units in (9), we obtain that naive Bayes predicts that the first person is not a smoker and that the second is, in both cases *with posterior probability equal to 0.90*! NCC2, more reasonably, outputs $\mathcal{K}$ in both cases, as a way to say that there is not information enough to make any reliable prediction. This example also points out that NCC2 correctly suspends the judgment even on instances on which precise models such as naive Bayes are actually very confident (for more details about this point, see Section 4.4 in the paper by Corani & Zaffalon, 2008).

### 5.3.2 On the Failure of Empirical Evaluations

Consider the question of the IID vs. the ID assumption for an IP, as discussed in Section 5.2.2, but placed here in the context of classification. To make things easier, we focus on finite possibility spaces, and assume in addition that the CAR IP is not present and that the classes are always observed precisely (i.e., as singletons).

The first interesting thing to note is the following. Recall that imposing the full IID assumption (i.e., IID both for the data-generating process and the IP) implies that the (set-valued) observations can be regarded as the outcomes of a multinomial process. This, together with the fact that the classes are always observed precisely, enables us to discard the $Y$-variables from consideration: they are not necessary because one can learn the relationship between the classes and the features directly by using the $W$-variables. To make this more clear, consider the case of incomplete samples originated by missing values. The full IID assumption implies in this case that one is allowed to regard the symbol '?' of missing value as a further possible value, and to proceed with the traditional learning methods for the case of complete samples. This shows that the full IID assumption in classification makes CIR collapse to the traditional updating rule, although applied to the $W$-variables. This makes things easier to deal with, and can thus be regarded as an advantage of the full IID assumption in classification. But we have already argued that the IID assumption for an IP is strong, and so we turn to consider the weaker ID assumption also for pattern classification, that we consider more tenable.

Let us focus on a very simple classification setting where the complete data are generated in an IID way, and where the IP is ID. The problem is to predict the class of the unit to classify given its features and the previous units $(1, \ldots, N)$. As already mentioned in Section 5.3, this is usually done in the precise case by following a maximum expected utility approach. Call a model that acts in such a way a *precise classifier*.

It is common practice with precise classifiers to measure their accuracy empirically. In its simplest form, this is obtained by randomly splitting the available data into a *learning* and a *test set*, by inferring the classifier from the former and testing it on the latter. This simple, yet powerful, idea is responsible for much of the success and popularity of classification, as it enables one to be relatively confident about how well a classifier performs on previously unseen data. Unfortunately, this key characteristic is lost when we cannot assume that the unknown IP is IID.

To see why, consider (for all $i = 1, \ldots, N + 1$) a Boolean class variable $C_i$ and two Boolean feature variables $A_i$ and $B_i$, such that $F_i = \bar{F}_i = (A_i, B_i)$. Assume that the class is the result of the exclusive logical disjunction of the two attributes: i.e., the class equals one if and only if either the first or the second attribute equals one, but not both. Assume that the complete data are eventually turned into incomplete data by an (unknown) IP whose only action is to make $B_i$ missing if and only if $(A_i, B_i) = (1, 0)$, for all $i$. Let $W_{A_i}$ and $W_{B_i}$ be the observation variables for $A_i$ and $B_i$, respectively. The IP is then characterised by $P(W_{B_i} = ?|A_i = 1, B_i = 0) = 1$, so that observing the pattern $(W_{A_i} = 1, W_{B_i} = ?)$ implies $C_i = 1$ with certainty. In these conditions, any precise classifier is clearly expected to learn that $B_{N+1}$ being missing is irrelevant to predict $C_{N+1}$, whose value coincides with the value of $A_{N+1}$. Since this is true for all the available data, partitioning them into learning and test set will do nothing but confirm it: the prediction accuracy on the pattern $(W_{A_{N+1}} = 1, W_{B_{N+1}} = ?)$ will be perfect, i.e., 100%. But the IP is not identically distributed, and it happens that when the classifier is put to work in an operative environment, the IP changes, in particular by making $B_i$ missing if and only if $(A_i, B_i) = (1, 1)$; or, in other words: $P(W_{B_i} = ?|A_i = 1, B_i = 1) = 1$. Once put to work in practice, the classifier will be always wrong on the pattern $(W_{A_{N+1}} = 1, W_{B_{N+1}} = ?)$, the prediction accuracy dropping to 0%.

Of course the example is designed so as to illustrate an extreme situation. However, experiments on real data sets done without using such an extreme unknown IP, as reported by Corani and Zaffalon (2008, Section 4.6), show that this phenomenon can indeed severely bias the empirical measures of performance. This appears to point to a fact: that empirical evaluations are doomed to failure in general when the data are made incomplete by a non-IID unknown process.

These considerations may have profound implications for classification, and more generally for data analysis. These fields of scientific research rest on two fundamental pillars: (i) that the assumptions made to develop a certain model (e.g., a classifier) are tenable; and (ii) that empirical evaluations are reliable. The crucial point is that both pillars may be very fragile with incomplete data, so being unable to sustain credible models and conclusions.

The way left to cope with such critical issues seems necessarily to rely on doing tenable assumptions. This involves recognising that the incompleteness process may not be IID.

## 6. Advanced Notions about Coherent Lower Previsions

We introduce some advanced notions about coherent lower previsions that we need in the rest of the paper, and in particular for the derivation of CIR in Section 7.

### 6.1 Coherence of a Number of Conditional Lower Previsions

Reconsider the setup introduced in Section 2 made of variables $X_1, \ldots, X_n$ about which a subject expresses beliefs. In practice, he can provide assessments for any disjoint subsets $O, I$ of $\{1, \ldots, n\}$; it is thus not uncommon to model a subject's beliefs using a finite number of different conditional lower previsions. Formally, we are going to consider what we shall call *collections* of conditional lower previsions.

**Definition 9** *Let* $\underline{P}_1(X_{O_1}|X_{I_1}), \ldots, \underline{P}_m(X_{O_m}|X_{I_m})$ *be conditional lower previsions with respective domains* $\mathcal{K}^1, \ldots, \mathcal{K}^m \subseteq \mathcal{L}(\mathcal{X}^n)$, *where* $\mathcal{K}^j$ *is the set of* $\mathcal{X}_{O_j \cup I_j}$-*measurable gambles,*[18] *for* $j = 1, \ldots, m$. *This is called a* collection *on* $\mathcal{X}^n$ *when for each* $j_1 \neq j_2$ *in* $\{1, \ldots, m\}$, *either* $O_{j_1} \neq O_{j_2}$ *or* $I_{j_1} \neq I_{j_2}$.

This means that we do not have two different conditional lower previsions giving information about the same set of variables $X_O$, conditional on the same set of variables $X_I$.

Even if all the conditional lower previsions in a collection are separately coherent, they do not need to be coherent with one another, so that a collection could still express inconsistent beliefs. To be able to model this joint coherence, we first introduce some new concepts. Remember that we use $\mathbb{I}_A$ to denote the *indicator function* of the set $A$, i.e., the function whose value is 1 for the elements of $A$ and 0 elsewhere.

**Definition 10** *For all gambles* $f$ *in the domain* $\mathcal{K}_{O \cup I}$ *of the conditional lower prevision* $\underline{P}(X_O|X_I)$, *and all* $x \in \mathcal{X}_I$, *we shall denote by* $G(f|x)$ *the gamble that takes the value* $\mathbb{I}_{\pi_I^{-1}(x)}(y)(f(y) - \underline{P}(f|x))$ *for all* $y \in \mathcal{X}^n$, *and by* $G(f|X_I)$ *the gamble equal to* $\sum_{x \in \mathcal{X}_I} G(f|x)$.

These are (almost-)desirable gambles for our subject: under the behavioural interpretation of $\underline{P}(f|x)$ as the supremum acceptable buying price for $f$ contingent on $x$, the gamble $G(f|x) + \epsilon \mathbb{I}_{\pi_I^{-1}(x)}$ is equivalent to buying $f$ for the price $\underline{P}(f|x) - \epsilon$, provided that $X_I = x$, and is therefore desirable. Since this happens for $\epsilon$ arbitrarily small, we deduce that the transaction $G(f|x)$ is almost-desirable. That $G(f|X_I) = \sum_{x \in \mathcal{X}_I} G(f|x)$ is also almost-desirable follows from the rationality principle that a sum of gambles which are almost-desirable for our subject should also be almost-desirable.[19]

---

18. We use $\mathcal{K}^j$ instead of $\mathcal{K}_{O_j \cup I_j}$ in order to alleviate the notation when no confusion is possible about which variables are involved.

19. In the case of an infinite $\mathcal{X}_I$ we need to add another rationality principle (Walley, 1991, Sect. 6.3.3).

**Definition 11** *The $\mathcal{X}_I$-support $S(f)$ of a gamble $f$ in $\mathcal{K}_{O \cup I}$ is given by*

$$S(f) := \{\pi_I^{-1}(x) : x \in \mathcal{X}_I, f\mathbb{I}_{\pi_I^{-1}(x)} \neq 0\}, \tag{10}$$

*i.e., it is the set of conditioning events for which the restriction of $f$ is not identically zero.*

**Definition 12** *Let $\underline{P}_1(X_{O_1}|X_{I_1}), \ldots, \underline{P}_m(X_{O_m}|X_{I_m})$ be separately coherent. They are co-herent when for all $f_j \in \mathcal{K}^j$, $j = 1, \ldots, m$, $j_0 \in \{1, \ldots, m\}$, $f_0 \in \mathcal{K}^{j_0}, z_0 \in \mathcal{X}_{I_{j_0}}$, there is some $B \in \{\pi_{I_{j_0}}^{-1}(z_0)\} \cup \cup_{j=1}^m S_j(f_j)$ such that*

$$\sup_{x \in B} \left[ \sum_{j=1}^m G_j(f_j|X_{I_j}) - G_{j_0}(f_0|z_0) \right](x) \geq 0,$$

*where $S_j(f_j)$ is the $\mathcal{X}_{I_j}$-support of $f_j$, as defined by Equation (10).*

We restrict the supremum to a subset of $\mathcal{X}^n$ because a gamble $f \leq 0$ such that $f < 0$ on some subset $A$ of $\mathcal{X}^n$ should not be desirable for our subject.

The notion in Definition 12 is sometimes also called *joint* (or *strong*) coherence. It is the strongest notion of self-consistency in Walley's theory, and can be regarded as the unique axiom of such a theory. The intuition behind this notion is that we should not be able to raise the conditional lower prevision for a gamble taking into account the acceptable transactions implicit in the other conditional previsions. Let us make this clearer. Assume Definition 12 fails, and that there is some $\delta > 0$ such that

$$\left[ \sum_{j=1}^m G_j(f_j|X_{I_j}) - G_{j_0}(f_0|z_0) \right](x) \leq -\delta < 0, \tag{11}$$

for every $x$ in $B \in \{\pi_{I_{j_0}}^{-1}(z_0)\} \cup \cup_{j=1}^m S_j(f_j)$. As a consequence,

$$G_{j_0}(f_0|z_0) - \delta \mathbb{I}_{z_0} = \mathbb{I}_{z_0}(f_0 - (\underline{P}_{j_0}(f_0|z_0) + \delta)) \geq \sum_{j=1}^m G_j(f_j|X_{I_j}),$$

taking into account that the inequality holds trivially on the elements outside $\{\pi_{I_{j_0}}^{-1}(z_0)\} \cup \cup_{j=1}^m S_j(f_j)$, and that on this class it is a consequence of Equation (11). But since the right-hand side is a sum of almost-desirable gambles, this means that the left-hand side should also be almost-desirable. This means that any price strictly smaller than $\underline{P}_{j_0}(f_0|z_0) + \delta$ should be an acceptable buying price for the gamble $f_0$, conditional on $z_0$. This contradicts the interpretation of $\underline{P}_{j_0}(f_0|z_0)$ as the supremum acceptable buying price.

**Example 4** *(Walley, 1991, Example 7.3.5) Let $X_1, X_2$ be two random variables taking values in $\mathcal{X}_1 = \mathcal{X}_2 = \{1, 2, 3\}$, and assume that we make the somewhat contradictory assessments $X_1 = 1 \Rightarrow X_2 = 1, X_1 = 2 \Rightarrow X_2 = 2, X_2 = 1 \Rightarrow X_1 = 2, X_2 = 2 \Rightarrow X_1 = 1$ and $X_1 = 3 \Leftrightarrow X_2 = 3$. These assessments can be modelled by the conditional previsions $\underline{P}(X_1|X_2)$ and $\underline{P}(X_2|X_1)$ given by*

$$\underline{P}(f|X_2 = 1) = f(2, 1), \underline{P}(f|X_2 = 2) = f(1, 2), \underline{P}(f|X_2 = 3) = f(3, 3)$$

$$\underline{P}(f|X_1 = 1) = f(1, 1), \underline{P}(f|X_1 = 2) = f(2, 2), \underline{P}(f|X_1 = 3) = f(3, 3)$$

for any gamble $f$ on $\mathcal{X}_1 \times \mathcal{X}_2$. To see that they are not coherent, consider the gambles $f_1 = \mathbb{I}_{\{(1,2),(2,1)\}}, f_2 = \mathbb{I}_{\{(1,1),(2,2)\}}, f_3 = 0$. It follows from Equation (10) that $S_1(f_2) = \{\{1\} \times \mathcal{X}_2, \{2\} \times \mathcal{X}_2\}$ and $S_2(f_1) = \{\mathcal{X}_1 \times \{1\}, \mathcal{X}_1 \times \{2\}\}$. Then

$$G(f_1|X_2) + G(f_2|X_1) - G(f_3|X_1 = 1) = G(f_1|X_2) + G(f_2|X_1) < 0$$

on the set $\{(x_1, x_2) \in B \text{ for some } B \in S_1(f_2) \cup S_2(f_1) \cup \{\pi_1^{-1}(1)\}\} = \mathcal{X}_1 \times \mathcal{X}_2 \setminus \{(3,3)\}$. ◇

## 6.2 Coherence Graphs

A collection of lower previsions can be given a graphical representation that we call a *coherence graph* (Miranda & Zaffalon, 2009). A coherence graph is a directed graph made of two types of nodes: *actual* and *dummy* nodes. Dummy nodes are in one-to-one correspondence with the lower previsions in the collection; actual nodes represent the variables $X_1, \dots, X_n$.

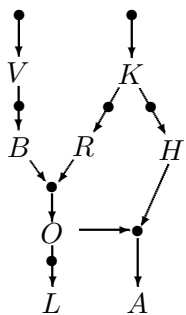

Figure 3: The coherence graph originated by the A1$^+$-representable collection: $\underline{P}_1(V)$, $\underline{P}_2(K)$, $\underline{P}_3(B|V)$, $\underline{P}_4(R|K)$, $\underline{P}_5(H|K)$, $\underline{P}_6(O|B,R)$, $\underline{P}_7(L|O)$, $\underline{P}_8(A|O,H)$.

Figure 3 shows the coherence graph for the conditional distributions owned by the nodes of the Asia network which we represent as lower previsions, i.e., $\underline{P}_1(V)$, $\underline{P}_2(K)$, $\underline{P}_3(B|V)$, $\underline{P}_4(R|K)$, $\underline{P}_5(H|K)$, $\underline{P}_6(O|B,R)$, $\underline{P}_7(L|O)$, $\underline{P}_8(A|O,H)$. In order to avoid confusion, we stress that a coherence graph can be built out of any set of lower previsions, not just those arising from graphical models. We use the Asia network here only for illustrative purposes. Moreover, in building the coherence graph from the Asia net, we completely disregard any independence information coded by the net, and we focus only on the list of conditional distributions. Also, the coherence graph is the same irrespective of whether the conditional assessments are linear or not.

We adopt some conventions to display a coherence graph: we denote actual nodes with the same letter of the corresponding variable; dummy nodes are instead denoted by black solid circles and are not labelled. Finally, for a dummy node with both a single parent and a single child, we do not show the arrow entering the node, so as to make the graph simpler to see.

A collection of lower previsions is turned into a coherence graph by turning each of its lower previsions into a subgraph called a *D-structure*: a directed graph consisting of a dummy node, its directed predecessors (i.e., its *parents*) and successors (i.e., its *children*), and the arcs connecting the dummy node to its parents and children. The parents of the dummy node are the actual nodes corresponding to the variables on the right-hand side of the conditioning bar of the prevision related to the dummy node. The children are the actual nodes corresponding to the variables on the left-hand side of the conditioning bar. For example, the D-structure for $\underline{P}_1(V)$ is the subgraph in Figure 3 consisting of $V$, its dummy parent, and the arc connecting them; the D-structure for $\underline{P}_6(O|B, R)$ is the subgraph consisting of the actual nodes $B$, $R$, $O$, the dummy that is child of the first two and parent of the last one, and the three arcs connecting them.

In order to distinguish a coherence graph from other graphical models such as Bayesian nets, note that in a coherence graph we cannot deduce independence between nodes in the graph: for instance, in Figure 3 we cannot conclude that the variables $L, A$ are independent given $O, H$. The reason is that we can have instances of conditional previsions with this very same coherence graph for which the variables $L$ and $A$ are dependent.

Coherence graphs can have different forms, and each of these forms has some implication on the coherence of the related collection of lower previsions. In this paper we focus on coherence graphs called of *type A1+*: these coherence graphs are acyclic and have the property that each actual node in the graph has *exactly* one parent. This is indeed the case of the graph in Figure 3. A more general class of graphs is A1 graphs, which is defined similarly to the A1$^+$ case but with the difference that each actual node in the graph is required to have *at most* one parent.

When a coherence graph is of type A1$^+$ (resp. A1), we say that the related collection of lower prevision is *A1+-representable* (resp. A1-representable). A1-representable collections are interesting because the separate coherence of the lower previsions in the collection is equivalent to their joint coherence, as shown by Miranda and Zaffalon (2009, Proposition 4). This means that A1-representable collections are known to be coherent irrespective of the numbers that make up their lower previsions, provided that they give rise to separately coherent lower previsions.

## 6.3 Updating Coherent Lower Previsions

One updating rule that can be used with lower previsions is the regular extension:

**Definition 13** *Let $\underline{P}$ be a coherent lower prevision, $f \in \mathcal{K}_{O \cup I}$ and $x \in \mathcal{X}_I$ an element for which $\overline{P}(x) > 0$, where $\overline{P}$ is the conjugate upper prevision derived from $\underline{P}$. The* regular extension $\underline{R}(f|x)$ *is $\underline{R}(f|x) := \inf\left\{ \frac{P(f\mathbb{I}_x)}{P(x)} : P \in \mathcal{M}(\underline{P}), P(x) > 0 \right\}$.*

Recall that $\mathcal{M}(\underline{P})$ is the set of linear previsions $P$ with domain $\mathcal{L}(\mathcal{X}^n)$ that dominate $\underline{P}$. When $\mathcal{X}_I$ is finite and $\overline{P}(x) > 0$ for all $x \in \mathcal{X}_I$, the conditional lower prevision $\underline{R}(X_O|X_I)$ defined by regular extension is coherent with $\underline{P}$. The definition shows that the regular extension also has a nice sensitivity analysis interpretation: i.e., applying Bayes rule to the dominating linear previsions whenever possible. Perhaps for this reason, the regular extension has been proposed and used a number of times in the literature as an updating rule (e.g., De Campos, Lamata, & Moral, 1990; De Cooman & Zaffalon, 2004; Fagin & Halpern, 1991; Jaffray, 1992; Walley, 1981, 1991, 1996b).

**Example 5** *Assume we know that a coin is loaded, in the sense that it either always lands on heads or on tails, but we do not know on which. Let $X_i$ be the outcome of a throw for* $i = 1, 2$. *Our beliefs about* $(X_1, X_2)$ *may be modelled by the vacuous lower prevision* $\underline{P}$ *on* $\{(heads, heads), (tails, tails)\}$. *If we apply regular extension to define* $\underline{R}(X_2|X_1)$, *we obtain*

$$\underline{R}(X_2 = heads|X_1 = heads) = 1, \underline{R}(X_2 = tails|X_1 = tails) = 1,$$

*taking into account that* $P(heads, heads) + P(tails, tails) = 1$ *for any linear prevision* $P \in \mathcal{M}(\underline{P})$. $\diamond$

## 6.4 Products of Conditional Lower Previsions

We shall later use a generalisation of the marginal extension theorem (Walley, 1991, Theorem 6.7.2) established by Miranda and De Cooman (2007).

**Definition 14** *Let* $\underline{P}_1(X_{O_1}), \underline{P}_2(X_{O_2}|X_{I_2}), \ldots, \underline{P}_m(X_{O_m}|X_{I_m})$ *be separately coherent conditional lower previsions with respective domains* $\mathcal{K}^1, \ldots, \mathcal{K}^m$, *where* $I_1 = \emptyset$ *and* $I_j = \cup_{i=1}^{j-1}(I_i \cup O_i) = I_{j-1} \cup O_{j-1}$ *for* $j = 2, \ldots, m$. *Their* marginal extension *to* $\mathcal{L}(\mathcal{X}^n)$ *is given by*

$$\underline{P}(f) = \underline{P}_1(\underline{P}_2(\ldots(\underline{P}_m(f|X_{I_m})|\ldots)|X_{I_2})),$$

*and it is the smallest unconditional coherent lower prevision which is (jointly) coherent with* $\underline{P}_1(X_{O_1}), \underline{P}_2(X_{O_2}|X_{I_2}), \underline{P}_3(X_{O_3}|X_{I_3}), \ldots, \underline{P}_m(X_{O_m}|X_{I_m})$.

To see that the above definition makes sense, note that for any gamble $f$ on $\mathcal{X}^n$, $\underline{P}_m(f|X_{I_m})$ is a gamble on $\mathcal{X}_{I_m}$, and hence belongs to the domain of $\underline{P}_{m-1}(X_{O_{m-1}}|X_{I_{m-1}})$; then $\underline{P}_{m-1}(\underline{P}_m(f|X_{I_m})|X_{I_{m-1}})$ is a gamble on $\mathcal{X}_{I_{m-1}}$ which belongs to the domain of $\underline{P}_{m-2}(X_{O_{m-2}}|X_{I_{m-2}})$; and by repeating the argument, $\underline{P}_2(\ldots(\underline{P}_m(f|X_{I_m})|\ldots)|X_{I_2})$ is a gamble on $\mathcal{X}_{I_2} = \mathcal{X}_{O_1}$ which belongs to the domain of the unconditional prevision $\underline{P}_1$.

The idea behind the construction of the marginal extension is that, when we have hierarchical information, the way to combine it into a joint prevision is to follow the order in which this information is structured. It reduces, for instance, to the law of total probability in the case of linear previsions and finite spaces. But it is applicable in more general situations: when we combine a finite number of previsions, and not only two of them; when we are dealing with infinite spaces; and when we have lower previsions instead of linear ones.

We give next a notion of conditional independence for coherent lower previsions that we shall use in the following.

**Definition 15** *Consider three variables,* $X_i, X_j, X_k$, *and a coherent conditional lower prevision* $\underline{P}(X_i, X_j|X_k)$. $X_i$ *and* $X_j$ *are* strongly independent *conditional on* $X_k$ *if* $X_i$ *and* $X_j$ *are stochastically independent given* $X_k$ *for all the extreme points of* $\mathcal{M}(\underline{P}(X_i, X_j|x_k))$ *and for all* $x_k \in \mathcal{X}_k$.

The previous definition allows joint beliefs to be created out of marginal ones, as it is common with independence in precise probability: consider the coherent conditional lower previsions $\underline{P}_i(X_i|X_k)$ and $\underline{P}_j(X_j|X_k)$. We build their so-called *strong product* $\underline{P}_{\mathrm{sp}}(X_i, X_j|X_k)$, i.e., the least-committal lower prevision that follows from them and the assumption of strong independence, as the lower envelope of the following set:

$$
\begin{aligned}
\mathcal{M}_{\mathrm{sp}} := \quad & \{P(X_i, X_j | X_k) := P_i(P_j(X_j | X_i, X_k) | X_k) \text{ s.t.} \forall x_k \in \mathcal{X}_k, \\
& P_i(X_i | x_k) \in ext(\mathcal{M}(\underline{P}_i(X_i | x_k))), P(X_j | x_k) \in ext(\mathcal{M}(\underline{P}_j(X_j | x_k))), \\
& \text{and } \forall x_i \in \mathcal{X}_i, P_j(X_j | x_i, x_k) := P(X_j | x_k)\},
\end{aligned}
$$

where $P_j(X_j | X_i, X_k)$ is defined using stochastic independence, and the linear prevision $P(X_i, X_j | X_k)$ is obtained by applying the marginal extension theorem.

Strong independence is a relatively straightforward generalisation of stochastic independence to imprecision. It is also a notion that can immediately be given a sensitivity analysis interpretation: since for every $x_k \in \mathcal{X}_k$ each extreme point of $\mathcal{M}(\underline{P}_{\mathrm{sp}}(X_i, X_j | x_k))$ satisfies stochastic independence, we can interpret $\underline{P}_{\mathrm{sp}}(X_i, X_j | X_k)$ as a model arising from partial knowledge of an underlying linear prevision $P(X_i, X_j | X_k)$ which is known to satisfy stochastic independence. In other words, $\underline{P}_{\mathrm{sp}}(X_i, X_j | X_k)$ could be regarded as obtained by listing a number of candidate linear previsions $P(X_i, X_j | X_k)$, each of which satisfies stochastic independence, and by taking their lower envelope.

It should also be remarked that in the imprecise case, where we work with sets of probabilities or previsions, there is not a unique extension of the notion of stochastic independence of probability measures (see Campos & Moral, 1995; Couso et al., 2000; or Miranda, 2008, Section 4). The notion we have just considered is the strongest, and therefore more informative, of the possibilities considered in the references above.[20]

A notion alternative to strong independence is for instance epistemic irrelevance as proposed by Walley (1991, Section 9.1.1).

**Definition 16** *Given the coherent lower previsions $\underline{P}_i(X_i | X_j, X_k)$ and $\underline{P}_i(X_i | X_k)$, we say that $X_j$ is epistemically irrelevant to $X_i$ conditional on $X_k$ if it holds that $\underline{P}_i(X_i | X_j, X_k) = \underline{P}_i(X_i | X_k)$.*

Epistemic irrelevance is naturally suited for a behavioural interpretation: $X_j$ is irrelevant to $X_i$ given $X_k$ because in the context of $X_k$ the beliefs of a subject about $X_i$, expressed by the coherent lower prevision $\underline{P}_i(X_i | X_k)$, do not change after getting to know the value of $X_j$. A sensitivity analysis interpretation of epistemic irrelevance is not possible in general because for a joint created out of marginal information and epistemic irrelevance, the related extreme points do not necessarily satisfy stochastic independence. In fact, strong independence implies epistemic irrelevance while the converse implication does not hold. Moreover, irrelevance is an asymmetric notion: knowing that $X_j$ is irrelevant to $X_i$ does not imply that $X_i$ is irrelevant to $X_j$. One can create a symmetric notion of *epistemic independence* (see, e.g., Walley, 1991; Miranda, 2008) by requiring that both $X_j$ is irrelevant to $X_i$ and $X_i$ is irrelevant to $X_j$; still, strong independence implies epistemic independence but the other way around does not hold.

Strong independence can be expressed by means of epistemic irrelevance together with a further requirement. Consider $\underline{P}_i(X_i | X_k)$ and $\underline{P}_j(X_j | X_k)$; we want to create their strong product using epistemic irrelevance. We first write that $\underline{P}_j(X_j|X_i, X_k) = \underline{P}_j(X_j|X_k)$, as strong independence implies epistemic irrelevance. Then we apply the marginal extension theorem to create the so-called *irrelevant product* $\underline{P}_{\mathrm{ip}}(X_i, X_j|X_k) = \underline{P}_i(\underline{P}_j(X_j|X_i, X_k)|X_k)$, which is the least committal lower prevision that follows from the marginals and the assumption of epistemic irrelevance. In terms of the sets of dominating linear previsions, the marginal extension theorem states that $\underline{P}_{\mathrm{ip}}(X_i, X_j|X_k)$ is the lower envelope of the set

$$\mathcal{M}_{\mathrm{ip}} := \quad \{P(X_i, X_j|X_k) = P_i(P_j(X_j|X_i, X_k)|X_k) : \forall x_i \in \mathcal{X}_i, x_k \in \mathcal{X}_k,$$
$$P_i(X_i|x_k) \in ext(\mathcal{M}(\underline{P}_i(X_i|x_k))), P_j(X_j|x_i, x_k) \in ext(\mathcal{M}(\underline{P}_j(X_j|x_i, x_k)))\}.$$

---

20. This is related to the fact that it *never* holds that *all* the precise previsions in a convex set with more than one element satisfy stochastic independence; therefore requiring it for the extreme points of the credal set, which are the ones keeping all the behavioural information, amounts to go as far as possible on the direction to independence.

Even if for all $x_k \in \mathcal{X}_k$ the sets $ext(\mathcal{M}(\underline{P}_j(X_j|x_i, x_k)))$, $x_i \in \mathcal{X}_i$, are identical to one another because of the irrelevance condition, when building a joint prevision $P(X_i, X_j|X_k)$ we are not forced to choose the same extreme point in each of them. This is just what makes the difference between $\mathcal{M}_{\mathrm{ip}}$ and $\mathcal{M}_{\mathrm{sp}}$, and in fact we can write $\mathcal{M}_{\mathrm{sp}}$ in a way more similar to $\mathcal{M}_{\mathrm{ip}}$ as follows:

$$\mathcal{M}_{\mathrm{sp}} := \quad \{P(X_i, X_j|X_k) = P_i(P_j(X_j|X_i, X_k)|X_k) : \forall x_i \in \mathcal{X}_i, x_k \in \mathcal{X}_k,$$
$$P_i(X_i|x_k) \in ext(\mathcal{M}(\underline{P}_i(X_i|x_k))), P_j(X_j|x_i, x_k) \in ext(\mathcal{M}(\underline{P}_j(X_j|x_i, x_k)))$$
$$\text{s.t. } P_j(X_j|x_i', x_k) = P_j(X_j|x_i'', x_k) \text{ if } x_i' = x_i''\}.$$

In other words, we can think of the strong product as obtained by the same procedure that we use for the irrelevant product, with the additional requirement that for each $x_k \in \mathcal{X}_k$, the extreme points chosen in $ext(\mathcal{M}(\underline{P}_j(X_j|x_i, x_k)))$, $x_i \in \mathcal{X}_i$, must coincide every time. We shall use this observation in Section 7.1.3.

# 7. Derivation of CIR

In the next sections we state a number of assumptions, discuss them, and eventually use them to derive CIR.

## 7.1 Modelling Beliefs

Our aim in this section is to represent our beliefs about the vector $(Z, Y, W)$. We intend to model beliefs using coherent lower previsions. We refer to Sections 2 and 6 and, more generally, to the book by Walley (1991) for the concepts and results we shall need.

A way to represent beliefs about $(Z, Y, W)$ is through a coherent lower prevision on $\mathcal{L}(\mathcal{Z} \times \mathcal{Y} \times \mathcal{W})$, representing our *joint* beliefs about these variables. But, as with precise probability, it is often easier to build joint models out of the composition of simpler conditional and unconditional models. We shall therefore start by focusing on the following lower previsions, which we illustrate for clarity by a coherence graph in Figure 4:

(LP1) A coherent lower prevision, denoted by $\underline{P}_1$, on the set of $\mathcal{X}_{Z,Y}$-measurable gambles;

(LP2) A separately coherent conditional lower prevision $\underline{P}_2(\bar{W}|Z, Y)$ on the set of $\mathcal{X}_{Z,Y,\bar{W}}$-measurable gambles, modelling our beliefs about $\bar{W}$ given the value $(z, y)$ taken by $(Z, Y)$.

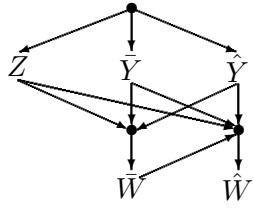

Figure 4: The coherence graph for the initial conditional lower previsions that express beliefs about $(Z, Y, W)$.

(LP3) A conditional lower prevision $\underline{P}_3(\hat{W}|Z, Y, \bar{W})$ on $\mathcal{L}(\mathcal{Z} \times \mathcal{Y} \times \mathcal{W})$, representing our beliefs about $\hat{W}$ given the value that $(Z, Y, \bar{W})$ takes.

The coherent lower prevision $\underline{P}_1$ is intended to express our beliefs about the domain of interest, which is modeled by the variables $Z$ and $Y$. In other words, by $\underline{P}_1$ we express our beliefs about facts. The conditional lower prevision $\underline{P}_2(\bar{W}|Z, Y)$ is concerned with our beliefs about the unknown incompleteness process. Finally, by the conditional lower prevision $\underline{P}_3(\hat{W}|Z, Y, \bar{W})$ we express our beliefs about the CAR incompleteness process. How exactly this is done is detailed in the next sections. For the moment, we are going to build a joint coherent lower prevision for $(Z, Y, W)$ out of the lower previsions listed in (LP1)–(LP3). This is done by the marginal extension theorem introduced in Section 6.4.

In the present case, the generalised marginal extension theorem states that the lower prevision $\underline{P}$ given by

$$\underline{P}(f) = \underline{P}_1(\underline{P}_2(\underline{P}_3(f|Z, Y, \bar{W})|Z, Y))$$

for all gambles $f$ on $\mathcal{Z} \times \mathcal{Y} \times \mathcal{W}$ is the smallest (using the point-wise ordering of lower previsions) coherent lower prevision on $\mathcal{L}(\mathcal{Z} \times \mathcal{Y} \times \mathcal{W})$ that is jointly coherent with the lower previsions listed in (LP1)–(LP3). This implies, for example, that $\underline{P}$ coincides with the coherent lower prevision $\underline{P}_1$ on its domain. Because it is the smallest lower prevision to be coherent with our assessments, $\underline{P}$ captures the behavioural implications present in $\underline{P}_1, \underline{P}_2(\bar{W}|Z, Y)$ and $\underline{P}_3(\hat{W}|Z, Y, \bar{W})$, without making any additional assumptions; we shall therefore use it as the model of our beliefs about the vector $(Z, Y, W)$.

$\underline{P}$ is also the lower envelope of the closed and convex set $\mathcal{M}(\underline{P})$ of dominating linear previsions. It will be useful for the purposes of this paper to construct this set explicitly. Consider the set

$$\mathcal{M}' := \left\{ P : P(g) = P_1(h_g), \ h_g(z, y) := \sum_w g(z, y, w) P_2(\bar{w}|z, y) P_3(\hat{w}|z, y, \bar{w}) \right\}, \quad (12)$$

for all $g \in \mathcal{L}(\mathcal{Z}, \mathcal{Y}, \mathcal{W})$, where $P_1 \in ext(\mathcal{M}(\underline{P}_1)), P_2(\bar{W}|z, y) \in ext(\mathcal{M}(\underline{P}_2(\bar{W}|z, y)))$ and $P_3(\hat{W}|z, y, \bar{w}) \in ext(\mathcal{M}(\underline{P}_3(\hat{W}|z, y, \bar{w})))$ for any $(z, y, w) \in \mathcal{Z} \times \mathcal{Y} \times \mathcal{W}$. Then $\mathcal{M}(\underline{P}) = \overline{CH}(\mathcal{M}')$, where the operator $\overline{CH}(\cdot)$ stands for convex closure. As a consequence, $\underline{P}$ can be calculated as the lower envelope of the class $\mathcal{M}'$.

### 7.1.1 Domain Beliefs

Consider variables $Z$ and $Y$, which represent facts. Beliefs about facts, in the form of a coherent lower prevision $\underline{P}_1$, are specific of the domain under consideration; for this reason, we call them *domain beliefs*. In the case of the Asia network, $\underline{P}_1$ would correspond simply to the joint mass function coded by the network itself.

We shall impose only a minimal assumption about domain beliefs, just because $\underline{P}_1$ must be as flexible a tool as possible in order to express beliefs in a wide range of domains. This task will be postponed after focusing on a very different kind of beliefs, which we call *beliefs about the incompleteness process*.

### 7.1.2 Beliefs about the Incompleteness Process

These are naturally of a conditional type: they formalise what we believe about the 'modus operandi' of the overall IP, i.e., the procedure by which it turns facts into observations. We represent these beliefs through assumptions to be satisfied by the conditional lower prevision $\underline{P}_2(\bar{W}|Z,Y)$, related to the unknown IP, and $\underline{P}_3(\hat{W}|Z,Y,\bar{W})$, related to the CAR IP. We start with the unknown IP.

The unknown IP has been introduced to formalise the idea of an incompleteness process about which we are nearly ignorant. The term 'nearly' is there to emphasise that, despite a deep kind of ignorance, something is assumed to be known about the unknown IP. For one thing, it produces incompleteness only on the basis of $\bar{Y}$. More formally, we assume that there is a separately coherent conditional lower prevision $\underline{P}_2(\bar{W}|\bar{Y})$ such that

$$\underline{P}_2(f|z,y) = \underline{P}_2(f|\bar{y}) \tag{BIP1}$$

for all $\mathcal{X}_{\bar{W},Y,Z}$-measurable gambles $f$ and any $(z,y)$ in $(\mathcal{Z},\mathcal{Y})$.

(BIP1) states that observing different values for $Z$ and $\hat{Y}$ does not change our beliefs about $\bar{W}$, once we know what value $\bar{Y}$ takes in $\bar{\mathcal{Y}}$. This assumption turns the coherence graph in Figure 4 into that of Figure 5. Assumption (BIP1) arises somewhat naturally within our interpretation of IPs as observational processes, in which we regard the unknown IP just as a process that takes $\bar{Y}$ in input and outputs $\bar{W}$. Still, it is an important assumption for the following developments, and is therefore discussed in some detail in Section 7.2.

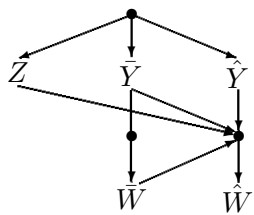

Figure 5: The coherence graph in Figure 4 after the application of Assumption (BIP1).

The second thing that we assume about the unknown IP is related to the perfection of the multi-valued map $\Gamma_{\bar{Y}}$:[21]

$$\bar{y} \in \bar{\mathcal{Y}} \Rightarrow \overline{P}_2(\Gamma(\bar{y})^c | \bar{y}) = 0. \tag{BIP2}$$

In other words, by this assumption we practically exclude the possibility that for some $\bar{y} \in \bar{\mathcal{Y}}$, the unknown IP may lead to observations outside $\Gamma(\bar{y})$. Note that for this to be well defined, it is important to assume that $\Gamma(\bar{y})$ is nonempty as required by (IP4). Remember that, here and elsewhere, we use $\overline{P}$ to refer to the conjugate upper prevision of $\underline{P}$; see Remark 1 for more details.

Nothing more is assumed about the unknown IP. We are then led to introduce some assumptions about the CAR IP, of which the first two are analogous to (BIP1) and (BIP2), respectively. The first one is the existence of a separately coherent conditional lower prevision $\underline{P}_3(\hat{W}|\hat{Y})$ such that

$$\underline{P}_3(f|z, y, \bar{w}) = \underline{P}_3(f|\hat{y}) \tag{BIP3}$$

for all $\mathcal{X}_{W,Y,Z}$-measurable gambles $f$ and $(z, y, \bar{w})$ in $(\mathcal{Z}, \mathcal{Y}, \bar{\mathcal{W}})$. This assumption turns the coherence graph in Figure 5 into that of Figure 6.

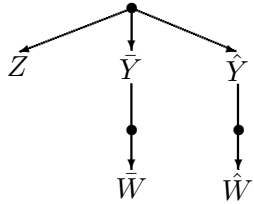

Figure 6: The coherence graph in Figure 5 after the application of Assumption (BIP3).

The second is an assumption of perfection of the multi-valued map $\Gamma_{\hat{Y}}$:

$$\hat{y} \in \hat{\mathcal{Y}} \Rightarrow \overline{P}_3(\Gamma(\hat{y})^c | \hat{y}) = 0. \tag{BIP4}$$

Again, this assumption is made possible by (IP5).

Now we assume something more substantial about the CAR IP. Indeed, the CAR IP has been introduced to model a process that produces incompleteness in a random fashion, that is to say, in a way that is not related to the underlying value that $\hat{Y}$ takes in $\hat{\mathcal{Y}}$. We model this belief as follows:

$$\hat{w} \in \hat{\mathcal{W}}, \hat{y} \in \{\hat{w}\}^* \Rightarrow \underline{P}_3(\hat{w}|\hat{y}) = \overline{P}_3(\hat{w}|\hat{y}) = \alpha_{\hat{w}}, \tag{BIP5}$$

where the $\alpha_{\hat{w}}$'s are positive constants that satisfy $\sum_{\hat{w} \in \Gamma(\hat{y})} \alpha_{\hat{w}} = 1$, where the restriction $\hat{w} \in \Gamma(\hat{y})$ is justified by (BIP4).

This assumption is usually called *coarsening at random* (or CAR, see Gill et al., 1997). In the special case of missingness processes, one usually refers to CAR as *missing at random* (or MAR, see Little & Rubin, 1987). CAR/MAR is probably the most frequently imposed assumption on IPs in the literature; it embodies the idea that an IP is non-selective (or non-malicious) in producing observations. Together with (BIP4), it makes the conditional prevision $\underline{P}_3(\hat{W}|\hat{Y})$ precise: i.e., our beliefs about the CAR IP are determinate (they are then the conditional expectation with respect to a probability distribution). An important point is that we must require explicitly that $\Gamma_{\hat{Y}}$ be compatible with the CAR assumption (see Appendix A for details); this leads to an assumption about $\Gamma_{\hat{Y}}$ other than those reported in Section 3.3: we assume that

$$\Gamma_{\hat{Y}} \text{ leads through (BIP5) to an } \textit{admissible} \text{ system of linear constraints,} \qquad \text{(IP9)}$$

that is, one that has a solution.

---

21. It could be argued that this assumption and the corresponding one for the CAR IP, that is, (BIP4), follow from the definition of the multi-valued maps. This nevertheless, we state the assumptions explicitly for clarity.

Now that we have characterised the unknown and the CAR IPs, we have also to determine the way in which they interact to make up the overall IP; we do this by imposing an assumption of strong independence, as introduced in Section 6.4:

$$\bar{W} \text{ and } \hat{W} \text{ are } \textit{strongly independent} \text{ conditional on } Y. \qquad \text{(BIP6)}$$

This assumption is discussed in Section 7.2.[22]

### 7.1.3 Joint Beliefs

In this section we aim at constructing the joint coherent lower prevision $\underline{P}$ mentioned at the beginning of Section 7.1. To do that, we need first to find explicit representations for the assessments related to the unknown and the CAR IP, on the basis of Assumptions (BIP1)–(BIP6). The joint will be constructed then using the marginal extension theorem mentioned at the beginning of the section, and taking into account the assumption of strong independence (BIP6).

Consider the unknown IP. Assumption (BIP1) allows us to focus our attention on the conditional lower prevision $\underline{P}_2(\bar{W}|\bar{Y})$. Thanks to Assumption (BIP2), we know that for any $\bar{y}$ in $\bar{\mathcal{Y}}$, $\underline{P}_2(\bar{W}|\bar{y})$ gives lower probability one to $\Gamma(\bar{y})$. Since we know nothing else, we take $\underline{P}_2(\bar{W}|\bar{y})$ to be the least-committal such lower prevision, i.e., the smallest separately coherent conditional lower prevision that assigns probability one to $\Gamma(\bar{y})$. Such a lower prevision is also called *vacuous relative to* $\Gamma(\bar{y})$, and is given by $\underline{P}_2(f|\bar{y}) = \min\{f(\bar{w}, \bar{y}) : \bar{w} \in \Gamma(\bar{y})\}$ for any $\mathcal{X}_{\bar{W},\bar{Y}}$-measurable gamble $f$.[23]

---

22. There are similarities between the model developed in this section and the more traditional hidden Markov models (see for instance Rabiner, 1989). First, both focus on the distinction between the two levels of information, the latent and the manifest one, as discussed in the Introduction. For this reason both are concerned with an explicit representation of the observational process, which is what relates the $W$ and the $Y$ variables, and which in our case coincides with the incompleteness process. If we refer to Figure 6, to make things more concrete, we see that the $W$-variables are manifest and they depend directly only on the related hidden $Y$-variables. This is a common representation in hidden Markov models. Second, Assumption (BIP6) could be interpreted as a kind of Markovianity property. Yet, there are also differences: a major one is the two $W$-variables are not related to an order (such as a time order), and more generally that we are not explicitly representing any order in our model. Another is that we quantify the probabilistic information of an observation conditional on the related hidden variable by a set of conditional mass functions rather than a single one. This gives us more expressivity and is the key to model the unknown IP.

23. It follows from this that the implication in (BIP2) is actually an equivalence.

Using indicator functions, we can easily write the extreme points of $\mathcal{M}(\underline{P}_2(\bar{W}|\bar{y}))$. Let $\mathbb{I}_{\bar{\gamma}} : \bar{\mathcal{W}} \to \{0, 1\}$ be the indicator function for $\bar{\gamma} \in \Gamma(\bar{y})$. We write $\mathbb{I}_{\bar{\gamma}}$ rather than $\mathbb{I}_{\{\bar{\gamma}\}}$ to simplify notation. Then $ext(\mathcal{M}(\underline{P}_2(\bar{W}|\bar{y}))) = \{\mathbb{I}_{\bar{\gamma}} : \bar{\gamma} \in \Gamma(\bar{y})\}$. This concludes the definition of $\underline{P}_2(\bar{W}|\bar{y})$ and also of $\underline{P}_2(\bar{W}|z, y)$, given that they coincide. Yet, when dealing with $\underline{P}_2(\bar{W}|z, y)$, we shall use a slightly different formulation: $ext(\mathcal{M}(\underline{P}_2(\bar{W}|z, y))) = \{\mathbb{I}_{\bar{\gamma}(z,y)} : \bar{\gamma}(z, y) \in \Gamma(\bar{y})\}$. The reason is that the extended notation $\bar{\gamma}(z, y)$ allows us to know, in addition to the vertex, the value $(z, y)$ on which we focus, and, consequently, that we focus on the specific lower prevision $\underline{P}_2(\bar{W}|z, y)$. Hence, we shall see $\bar{\gamma}$ as a mapping from $\mathcal{Z} \times \mathcal{Y}$ to $\bar{\mathcal{W}}$ such that for any $(z, y)$, $\bar{\gamma}(z, y)$ belongs to $\Gamma(\bar{y})$.

The situation is somewhat easier with the CAR IP. Similarly to the previous case, we know that we can restrict our attention to the conditional lower prevision $\underline{P}_3(\hat{W}|\hat{Y})$, thanks to (BIP3). But we also know that the conditional lower prevision $\underline{P}_3(\hat{W}|\hat{Y})$ is actually precise (we shall denote it by $P_3(\hat{W}|\hat{Y})$ from now on), and $P_3(\hat{W}|\hat{y})$ consists of the single mass function determined by Assumption (BIP5): i.e., the mass function that assigns probability $P(\hat{w}|\hat{y}) = \alpha_{\hat{w}} \mathbb{I}_{\{\hat{w}\}^*}(\hat{y})$ to the generic element $\hat{w} \in \hat{\mathcal{W}}$.

At this point we can define the coherent lower prevision $\underline{P}$ that models our beliefs about the value that $(Z, Y, W)$ assume jointly. We do this by rewriting Expression (12) according to the previous arguments and using in addition (BIP6):

$$\begin{aligned}
\mathcal{M}' := \{P : P(g) = P_1(h_g), \ h_g(z, y) = \sum_w \alpha_{\hat{w}} g(z, y, w) \mathbb{I}_{\bar{\gamma}(z,y)}(\bar{w}) \mathbb{I}_{\{\bar{w}\}^*}(\bar{y}) \mathbb{I}_{\{\hat{w}\}^*}(\hat{y}), \\
P_1 \in ext(\mathcal{M}(\underline{P}_1)), \bar{\gamma}(z, y) \in \Gamma(\bar{y}) \text{ s.t. } \bar{\gamma}(z, y) = \bar{\gamma}(z', y') \text{ if } \bar{y} = \bar{y}', \\
(z, y, w) \in \mathcal{Z} \times \mathcal{Y} \times \mathcal{W}\},
\end{aligned} \tag{13}$$

where we have introduced the new term $\mathbb{I}_{\{\bar{w}\}^*}(\bar{y})$. Such a term is actually redundant, and is introduced only because it is convenient for the next developments. To see that it is redundant, note that $\mathbb{I}_{\{\bar{w}\}^*}(\bar{y}) = 0$ implies that $\bar{w} \notin \Gamma(\bar{y})$ and hence, since $\bar{\gamma}(z, y) \in \Gamma(\bar{y})$ by definition, we have that $\mathbb{I}_{\bar{\gamma}(z,y)}(\bar{w}) = 0$. The usage of (BIP6), on the other hand, leads to the more substantial introduction of the requirement that $\bar{\gamma}(z, y) = \bar{\gamma}(z', y')$ when $\bar{y} = \bar{y}'$. This is needed to represent strong independence by means of epistemic irrelevance, as discussed at the end of Section 6.4. The lower envelope $\underline{P}$ of the set $\mathcal{M}'$ thus constructed is therefore called the strong product of the assessments $\underline{P}_1(Z, Y), \underline{P}_2(\bar{W}|\bar{Y}), P_3(\hat{W}|\hat{Y})$.

That the strong product is a coherent lower prevision, which is also coherent with $\underline{P}_1(Z, Y), \underline{P}_2(\bar{W}|\bar{Y}), P_3(\hat{W}|\hat{Y})$, follows from Corollary 1 in Appendix B. Importantly, this is also the case if the coherent lower prevision $\underline{P}_1(Z, Y)$ is constructed in a modular way from smaller pieces of information (i.e., as a joint which is coherent with a number of smaller pieces of information), provided that the coherence graph representing them is A1$^+$; such a situation is very often the case. It is the case, for instance, for all the the examples in Sections 5.1–5.3, as shown by Corollaries 3–5 in Appendix B.

### 7.1.4 An Assumption about Domain Beliefs

Recall that we have introduced $W$ as a way to model observations, and hence our beliefs should be consistent with the possibility of doing such an observation. To this extent, it seems necessary that we at least believe that $\{w\}^*$ can be produced:

$$w \in \mathcal{W} \Rightarrow \overline{P}(\{w\}^*) > 0, \tag{DB1}$$

taking into account that coherence implies that $\overline{P}(\{w\}^*) = \overline{P}_1(\{w\}^*)$. Assumption (DB1) is equivalent to the existence of an extreme point $P_1$ of $\mathcal{M}(\underline{P}_1)$ such that $P_1(\{w\}^*) > 0$. The following proposition shows that this assumption is sufficient to make our beliefs consistent with the possibility to do the observation $w$.

**Proposition 1** *Assumption (DB1) implies that* $\overline{P}(w) := \overline{P}(\mathcal{Z}, \mathcal{Y}, w) > 0$.

*Proof.* Let $P_1$ be an extreme point of $\mathcal{M}(\underline{P}_1)$ satisfying $P_1(\{w\}^*) > 0$, which exists by Equation (DB1), and take $\bar{\gamma}(z, y) = \bar{w}$ for any $\bar{y} \in \{\bar{w}\}^*$. The joint $P$ constructed from Equation (13) satisfies $P(w) = \alpha_{\bar{w}} P_1(\{w\}^*) > 0$. Hence, $\overline{P}(w) > 0$. □

## 7.2 Assumptions Discussed

We now discuss (BIP1), (BIP5) and (BIP6) in more detail. The other assumptions are quite weak and relatively easy to accept. Assumptions (BIP2) and (BIP4) may be exceptions as it makes sense to consider IPs that are imperfect (or that can lie), but this is out of the scope of the present paper.

We start with Assumption (BIP5), i.e., CAR. CAR models a process that does not coarsen facts with a specific purpose. CAR excludes in this way many common and important processes. Consider the medical domain, for example, focusing on a diagnostic application. A fact in this case might describe information about a patient, such as gender, age, lifestyle, and also the results of medical tests. Conclusions would be made by the possible diseases. The IP in this case (at least part of it) often results from the interaction between the doctor and the patient; indeed, there is usually a systematic bias in reporting, and asking for, symptoms that are present instead of symptoms that are absent; and a bias to report, and ask for, urgent symptoms more than others (Peot & Shachter, 1998). Furthermore, a doctor typically prescribes only a subset of the possible diagnostic tests, according to personal views and cost/benefit criteria.[24] Overall, the process described is non-CAR by definition, because the incompleteness arises following patterns that do depend on the specific facts under consideration.

Descriptions such as the one above support the idea that CAR is strong by means of informal arguments. But also on the formal level, recent research has suggested that CAR should be assumed to hold less frequently that it appears to be in practice (Grünwald & Halpern, 2003); and one should also remember that there is no way to test CAR statistically (Manski, 2003), so that there is always some degree of arbitrariness in assuming it. All of this should be taken into account in order to put CAR in a more balanced perspective. This is not to say that CAR should be rejected a priori, as there are situations when CAR is completely justified. Consider one notable example: the case when we know that $\hat{Y}$ is missing with probability equal to one. In this case the related IP clearly satisfies MAR: the probability of missingness is one irrespective of the value of $\hat{Y}$. More broadly speaking, MAR holds for processes that produce missingness in an unintentional way. In these cases, not assuming MAR would lead to results that are far too weak. The CAR IP in the modelling framework presented in Section 7.1.2 is designed just to account for these situations: i.e., to provide one with some flexibility in stating beliefs about the incompleteness process without having to adopt necessarily a worst-case approach (as opposed to the best case embodied by CAR/MAR), of the kind of the unknown IP.

---

24. This seems to support the idea that we shall hardly get ever rid of incompleteness: often, incompleteness does not happen by mistake, rather, it is generated *deliberately*. In these cases it actually represents *patterns of knowledge* (indeed, one can often tell what disease a patient was, or was not, suspected to have by looking only at the medical tests that a doctor did, and did not, prescribe). In this sense, it seems that incompleteness is doomed to be deeply rooted to many, if not most, real problems; and as such, it appears to be a fundamental, and indissoluble, component of uncertain reasoning.

The unknown IP is designed to model one's ignorance about an incompleteness process. It makes sense to adopt a conservative approach to model IPs in practice, for two specific reasons: first, IPs may be very difficult processes to model. They are a special case of observational processes, which often result from by human-to-human interaction, or by other complex factors; in a number of cases they can actually be regarded as the result of a communication protocol. The medical example above is intended to illustrate just this. It is not saying anything new: the difficulty in modelling IPs has been pointed out already long ago (Shafer, 1985; Grünwald & Halpern, 2003). But IPs are difficult objects to handle also for a second reason: IP models can be tightly dependent on specific situations. Consider the medical example: again different doctors typically ask different questions to diagnose a disease, even in the same hospital. By changing hospital, one can find entirely different procedures to diagnose the same disease, as the procedures depend on the local culture, on the money available to make the tests, and which ones, or the local time constraints. In other words, even if one is able to model an IP for a specific situation, that model may no longer be appropriate when another doctor is in charge of doing the diagnosis, or when one tries to apply the IP model to another hospital, perhaps in another country. In summary, modelling the IP (especially in a precise way) may present serious practical difficulties, as, in contrast with domain beliefs (e.g., medical knowledge), the way information can be accessed may well depend on the particular environment where a system will be used; and this means that models of the IP may not be easily reusable, and may therefore be costly.

These arguments support considering a conservative approach to model the IP that can be effectively implemented, and this is the reason for introducing the unknown IP in Section 7.1.2. Recall that the unknown IP is actually only *nearly* unknown, because we require that (BIP1) holds. On the other hand, observe that by dropping (BIP1) we could draw only vacuous conclusions about $Z$. To see this, suppose that you want to predict the probability of $Z = z$ given a certain observation $w$. Assume from now to the end of the section that there is no CAR IP, in order to make things easier. Then without assuming (BIP1), we could not exclude the possibility that the IP produces $w$ if and only if $Z \neq z$, so that the probability of $Z = z$ is zero. In the same way, we could not exclude the possibility that the IP produces $w$ if and only if $Z = z$, so that the probability of $Z = z$ is one. Of course, all the intermediate randomised cases would also be possible, so that the probability would be vacuous. This is to emphasise, perhaps not surprisingly, that complete ignorance about an IP is not consistent with the possibility of drawing useful conclusions.

Having said this, it is still useful to wonder whether (BIP1) is reasonable in the present setup. This is easier to do if we rewrite (BIP1) in a somewhat more natural form. We focus on the special case where the spaces of possibilities are finite and we have precise beliefs about the unknown IP. We can then multiply both sides of the equation $P(w|z, y) = P(w|y)$ by the conditional probability of $z$ given $y$, obtaining $P(z, w|y) = P(z|y)P(w|y)$. The new equation states that variables $Z$ and $W$ are independent conditional on $Y$. In other words, the original assumption is then equivalent to saying that if we already know fact $y$, making the observation $w$ is completely superfluous for predicting the target variable. This appears to be nothing else but the precise characterisation of the problems of incomplete or missing information: these problems are characterised by the fact that when something that can be missing is actually measured, the problem of missing information disappears. If this were not the case, the observation $w$ would not only carry information about $z$ via its implications on the fact $y$: it would say something about $z$ also on its own. But this means that some information useful to predict the target variable has not been included in the definition of the possible facts (see the work by De Cooman & Zaffalon, 2004 for further discussion about an assumption called MDI that is related to the assumption under consideration).

We conclude this section discussing Assumption (BIP6), namely the strong independence of $\bar{W}$ and $\hat{W}$ conditional on $Y$. We discuss this assumption because one could in principle consider weaker notions of independence to replace strong independence.

The reason why we have used strong independence is that for the kind of general framework that we have developed it is technically quite difficult to embed other judgments of independence of $\bar{W}$ and $\hat{W}$ conditional on $Y$. On the other hand, the problem with strong independence is that it does not easily lend itself to a full behavioural interpretation unlike other notions, such as epistemic irrelevance or independence. And we acknowledge that the behavioural interpretation is important both as a way to guide the mathematical developments and when one comes to decision making.

Yet, one could argue that the inference rule, CIR, that follows from our assumptions is the weakest possible one (if we exclude the rule that leads to vacuous posterior expectations and that does not lead to any informative conclusion) for the part related to the unknown IP, because it leads to consider all the replacements for the missing information as the possible latent data. Therefore one could conjecture that by replacing strong independence of $\bar{W}$ and $\hat{W}$ conditional on $Y$ with a weaker notion of independence, the inference rule could either stay the same or lead to vacuous inferences. Whatever the outcome, the choice of strong independence would be even more reasonable.

### 7.3 Derivation of CIR

In order to formulate the basic problem of this paper in a sufficiently general way, we focus on the problem of updating beliefs about a generic function $g : \mathcal{Z} \to \mathbb{R}$, to posterior beliefs conditional on $W = w$. In the precise case, this would be done by computing

$$P(g|w) = \frac{P(g\mathbb{I}_w)}{P(w)},$$

provided that $P(w) > 0$.

Something similar can be done in the imprecise case, but in such a case it not uncommon for $\underline{P}(w)$ to be zero. In that case, we should obtain an infinite number of conditional lower previsions which are coherent with $\underline{P}(W)$. This has been already discussed by De Cooman and Zaffalon (2004), who proposed the regular extension as a more effective updating rule. This is a very natural choice also under the sensitivity analysis interpretation of coherent lower previsions, as pointed out in Section 6.3, and is therefore also the choice that we pursue. Using the regular extension entails an additional rationality assumption that is reported by Walley (1991, Appendix J3); its coherence with the unconditional lower prevision it is derived from is trivial in the present context, where $\mathcal{W}$ is finite (see Appendix B for more details). We also recall that if $\underline{P}(w) > 0$, the regular extension provides the only coherent updated lower prevision.

The form of the CIR rule given in Definition 7 in Section 4 is derived in the next theorem. It is important to remark that for this theorem to hold, we need to require that the space $\bar{\mathcal{Y}}$ of the unknown IP be finite, while this is not necessary for all the other results we establish in Appendix B. The reason for this is that when $\bar{\mathcal{Y}}$ is infinite the class $\{\bar{w}\}_1^*$ that we define in the theorem may be empty for some $\bar{w}$ in $\bar{\mathcal{W}}$, making the rule inapplicable.

**Theorem 1 (Conservative inference rule theorem)** *Consider a gamble $g$ on $\mathcal{Z}$ and $w \in \mathcal{W}$. Let $\{\bar{w}\}_1^* := \{\bar{y} \in \{\bar{w}\}^* : \overline{P}_1(\bar{y}, \{\hat{w}\}^*) > 0\}$, and let us define the regular extension*

$$\underline{R}(g|\bar{y}, \{\hat{w}\}^*) = \inf_{P \geq \underline{P}_1 : P(\bar{y}, \{\hat{w}\}^*) > 0} P(g|\bar{y}, \{\hat{w}\}^*)$$

*for all $\bar{y} \in \{\bar{w}\}_1^*$. Then*

$$\underline{R}(g|w) = \underline{E}_1(g|w) := \min_{\bar{y} \in \{\bar{w}\}_1^*} \underline{R}(g|\bar{y}, \{\hat{w}\}^*).$$

*Proof.* First of all, for any prevision $P$ constructed using Equation (13),

$$P(w) = \alpha_{\hat{w}} \sum_{\bar{y} \in \bar{\mathcal{Y}}, \bar{\gamma}(z,y) = \bar{w}} P(\bar{y}, \{\hat{w}\}^*) = \alpha_{\hat{w}} \sum_{\bar{y} \in \{\bar{w}\}^*, \bar{\gamma}(z,y) = \bar{w}} P(\bar{y}, \{\hat{w}\}^*), \qquad (14)$$

for any $w \in \mathcal{W}$, where the second equality follows from Assumption (BIP2), and also taking into account that the mapping $\bar{\gamma}$ depends only on the value of $\bar{Y}$. Similarly, we also have

$$P(g\mathbb{I}_w) = \alpha_{\hat{w}} \sum_{\bar{y} \in \bar{\mathcal{Y}}, \bar{\gamma}(z,y) = \bar{w}} P(g\mathbb{I}_{\bar{y}, \{\hat{w}\}^*}) = \alpha_{\hat{w}} \sum_{\bar{y} \in \{\bar{w}\}^*, \bar{\gamma}(z,y) = \bar{w}} P(g\mathbb{I}_{\bar{y}, \{\hat{w}\}^*})$$

for any $w \in \mathcal{W}$ and any gamble $g$ on $\mathcal{Z}$.

Fix $w \in \mathcal{W}$, and define $\mathcal{M} := \{P \geq \underline{P}_1 : P(w) > 0\}$, $\mathcal{M}_1 := \{P \geq \underline{P}_1 : P(\bar{y}, \{\hat{w}\}^*) > 0$ for some $\bar{y} \in \{\bar{w}\}^*\}$. For all gambles $g$ on $\mathcal{Z}$, $\underline{R}(g|w) = \inf\{P(g|w) : P \in \mathcal{M}\}$ and $\underline{E}_1(g|w) = \inf\{P(g|\bar{y}, \{\hat{w}\}^*) : P \in \mathcal{M}_1\}$.

We start proving that $\underline{E}_1(g|w) \geq \underline{R}(g|w)$. For this, we are going to prove that for any $\bar{y} \in \{\bar{w}\}^*$ and any $P \in \mathcal{M}_1$ such that $P(\bar{y}, \{\hat{w}\}^*) > 0$ there is some $P' \in \mathcal{M}$ such that $P'(g|w) \leq P(g|\bar{y}, \{\hat{w}\}^*)$. Consider then such $P$, and let $P_1$ be its restriction to $\mathcal{L}(\mathcal{Z} \times \mathcal{Y})$. Then $P_1 \geq \underline{P}_1$. Let $\bar{\gamma}$ be a mapping such that $\bar{\gamma}(z,y) = \bar{w}$ and $\bar{\gamma}(z',y') \neq \bar{w}$ when $\bar{y}' \neq \bar{y}$. We can construct such a mapping because by Assumption (IP8) the set $\{\bar{w}\}_*$ is empty, and therefore for every $\bar{y}' \in \{\bar{w}\}^*$ there is some $\bar{w}' \neq \bar{w}$ such that $\bar{y}' \in \{\bar{w}'\}^*$. Let $P' \geq \underline{P}$ be the joint prevision constructed from $P_1$ and $\bar{\gamma}$ using Equation (13). Taking into account Equation (14), we see that this prevision satisfies $P'(w) = \alpha_{\hat{w}} P'(\bar{y}, \{\hat{w}\}^*) = \alpha_{\hat{w}} P(\bar{y}, \{\hat{w}\}^*) > 0$. As a consequence,

$$P'(g|w) = \frac{P'(g\mathbb{I}_w)}{P'(w)} = \frac{\alpha_w P'(g\mathbb{I}_{\bar{y}, \{\hat{w}\}^*})}{\alpha_w P'(\bar{y}, \{\hat{w}\}^*)} = \frac{P(g\mathbb{I}_{\bar{y}, \{\hat{w}\}^*})}{P(\bar{y}, \{\hat{w}\}^*)} = P(g|\bar{y}, \{\hat{w}\}^*),$$

where the last equality holds because $P(\bar{y}, \{\hat{w}\}^*) > 0$. Hence, we deduce that $\underline{E}_1(g|w) \geq \underline{R}(g|w)$.

We show now the converse inequality. We are going to prove that for any $P \in \mathcal{M}$ there is some $P' \in \mathcal{M}_1$ such that $P(g|w) \geq P'(g|y, \{\bar{w}\}^*)$. Consider $P \in \mathcal{M}$.

$$P(g|w) = \frac{P(g\mathbb{I}_w)}{P(w)} = \frac{\alpha_{\hat{w}} \sum_{\bar{y} \in \{\bar{w}\}^*, \bar{\gamma}(z,y) = \bar{w}} P(g\mathbb{I}_{\bar{y}, \{\hat{w}\}^*})}{\alpha_{\hat{w}} \sum_{\bar{y} \in \{\bar{w}\}^*, \bar{\gamma}(z,y) = \bar{w}} P(\bar{y}, \{\hat{w}\}^*)} = \frac{\sum_{\bar{y} \in \{\bar{w}\}^*, \bar{\gamma}(z,y) = \bar{w}} P(g\mathbb{I}_{\bar{y}, \{\hat{w}\}^*})}{\sum_{\bar{y} \in \{\bar{w}\}^*, \bar{\gamma}(z,y) = \bar{w}} P(\bar{y}, \{\hat{w}\}^*)}.$$

Define $I := \{\bar{y} \in \{\bar{w}\}^*$ s.t. $P(\bar{y}, \{\hat{w}\}^*) > 0, \bar{\gamma}(z,y) = \bar{w}\}$. Since $P(w) > 0$, it follows from Equation (14) that this set is nonempty, and the above equality can be expressed as

$$P(g|w) = \frac{\sum_{\bar{y} \in I} P(g\mathbb{I}_{\bar{y}, \{\hat{w}\}^*})}{\sum_{\bar{y} \in I} P(\bar{y}, \{\hat{w}\}^*)}.$$

Applying Lemma 3 in the Appendix we deduce the existence of $\bar{y}_1 \in I$ such that

$$\frac{P(g\mathbb{I}_{\bar{y}_1, \{\hat{w}\}^*})}{P(\bar{y}_1, \{\hat{w}\}^*)} \leq \frac{\sum_{\bar{y} \in \{\bar{w}\}^*, \bar{\gamma}(z,y) = \bar{w}} P(g\mathbb{I}_{\bar{y}, \{\hat{w}\}^*})}{\sum_{\bar{y} \in \{\bar{w}\}^*, \bar{\gamma}(z,y) = \bar{w}} P(\bar{y}, \{\hat{w}\}^*)}.$$

Let $P_1$ be the restriction of $P$ to $\mathcal{L}(\mathcal{Z} \times \mathcal{Y})$. Consider a mapping $\bar{\gamma}$ such that $\bar{\gamma}(z,y) = \bar{w}$ when $\bar{y} = \bar{y}_1$ and $\bar{\gamma}(z', y') \neq \bar{w}$ for any other $y'$. We can construct such a mapping because by Assumption (IP8) the set $\{\bar{w}\}_*$ is empty, and therefore for every $\bar{y}' \in \{\bar{w}\}^*$ there is some $\bar{w}' \neq \bar{w}$ such that $\bar{y}' \in \{\bar{w}'\}^*$. Let $P' \geq \underline{P}$ be the joint prevision constructed from $P_1$ and $\bar{\gamma}$ using Equation (13). This prevision satisfies $P'(\bar{y}_1, \{\hat{w}\}^*) = P(\bar{y}_1, \{\hat{w}\}^*) > 0$. Moreover,

$$P'(g|\bar{y}_1, \{\hat{w}\}^*) = \frac{P'(g\mathbb{I}_{\bar{y}_1, \{\hat{w}\}^*})}{P'(\bar{y}_1, \{\hat{w}\}^*)} = \frac{P(g\mathbb{I}_{\bar{y}_1, \{\hat{w}\}^*})}{P(\bar{y}_1, \{\hat{w}\}^*)} \leq \frac{\sum_{\bar{y} \in \{\bar{w}\}^*, \bar{\gamma}(z,y) = \bar{w}} P(g\mathbb{I}_{\bar{y}, \{\hat{w}\}^*})}{\sum_{\bar{y} \in \{\bar{w}\}^*, \bar{\gamma}(z,y) = \bar{w}} P(\bar{y}, \{\hat{w}\}^*)} = P(g|w),$$

where the second equality follows because $P' = P = P_1$ on $\mathcal{L}(\mathcal{Z} \times \mathcal{Y})$. We deduce from this $\underline{E}_1(g|w) \leq \underline{R}(g|w)$ and as a consequence they are equal. $\square$

An important point is whether the lower prevision defined by the CIR rule in this theorem is going to be coherent with the initial assessments: $\underline{P}_1(Z,Y), \underline{P}_2(\bar{W}|\bar{Y}), P_3(\hat{W}|\hat{Y})$. We want this to be the case as we want CIR to lead to self-consistent inference. This question is given a positive answer by Theorem 3 in Appendix B.

Moreover, from this theorem we also deduce (in its subsequent Corollary 2) that coherence will also be maintained under much more general conditions: on the one hand, for some of the examples of application of the CIR rule discussed in Sections 5.1–5.3, $\underline{P}_1(Z,Y)$ is obtained by making the strong product of a number of conditional lower previsions $\underline{P}_1(X_{O_1}|X_{I_1}), \ldots, \underline{P}_m(X_{O_m}|X_{I_m})$. Here we consider a set of variables of interest $\{X_1, \ldots, X_n\}$ that contains $\{Z, Y\}$ and does not include $W$. We assume moreover that the coherence graph associated to $\underline{P}_1(X_{O_1}|X_{I_1}), \ldots, \underline{P}_m(X_{O_m}|X_{I_m})$ is A1$^+$, from which it follows (see Appendix B) that $\cup_{i=1}^m (I_i \cup O_i) = \cup_{i=1}^m O_i = \{1, \ldots, n\}$. An instance of such a situation is given in Figure 7.

Corollary 2 in Appendix B shows that coherence also holds in this more general situation. On the other hand, the corollary also proves that coherence is maintained if we apply the regular extension to a finite number of conditioning events, thus actually creating an additional finite number of coherent lower previsions out of the original assessments: all of these assessments are going to be jointly coherent. This is another important result as in practice one indeed often conditions on a number of events. These theoretical results give us guarantees that the CIR rule is 'well-behaved.'

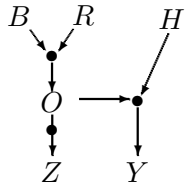

Figure 7: A more general model.

## 8. Conclusions

In this paper we have introduced a new rule to update beliefs under incompleteness that we have called *conservative inference rule* (CIR). CIR has been designed by explicitly taking into consideration the problem of modelling the process that makes our observations incomplete, and that we have called the *incompleteness process* (or IP). We have represented such a process as made of two sub-processes, one that is non-selective and another that is unknown to us. As a consequence, CIR deals differently with data that are known to be coarsened at random from the other ones. In the first case CIR treats them using the CAR/MAR assumption, in the other by taking all the completions of the incomplete data.

CIR can be regarded as a generalisation of both the traditional, Bayesian rule to update beliefs and of the recently proposed *conservative updating rule* (De Cooman & Zaffalon, 2004). CIR is a generalisation of these two rules as it enables one to consider mixed situations. This should make of CIR quite a flexible rule in applications. Moreover, this enables CIR to avoid the risk of being overconfident about the IP, which is a problem for the traditional updating, as well as to avoid being over-pessimistic as it may happen with the conservative updating rule.

CIR is also a very general rule. It can be used with coarsened or missing data, in statistics as well as in expert systems, and it can be applied to predict the state of a target variable irrespective of the cardinality of its space of possibilities. We have indeed tried to illustrate these characteristics using a number of application domains. Moreover, using some examples we have shown that CIR makes quite a difference with the traditional CAR/MAR-based way of dealing with incomplete data: in these examples the traditional updating is shown to lead to conclusions that are hardly justifiable from the evidence at hand, and, even worse, can do it in such a way that a user may not realise this. In these cases CIR is instead naturally more cautious and clearly communicates to a user that there are limits to the strength of the possible conclusions, as a logical consequence of the strength (or weakness) of the available information.

Finally, CIR has some nice theoretical properties, most notably that it is a *coherent* rule. Loosely speaking, this means that by using CIR it is not possible to give rise to probabilistic inconsistencies. As it can be seen from Theorem 3 in Appendix B, the assumptions we require for coherence of CIR are fairly general: if our initial assessments are of a special kind, which we have characterised using a graphical structure called a *coherence graph* of type A1$^+$, then we require only that some of the possibility spaces involved in the analysis are finite and that the conditioning events in these spaces have positive upper probability.

These hypotheses guarantee that we can use Walley's notion of *regular extension* to update our beliefs and that the assessments thus obtained are coherent with our initial assessments.

We should like to conclude this section commenting also a bit on some of the assumptions we have considered in the construction of our model and in some open problems that we can derive from this work.

One point is the requirement that some of the spaces of possibilities be finite. This could be relaxed by taking into account that a conditional prevision defined by regular extension can be coherent with the unconditional it is derived from even in the infinite case. However, we cannot guarantee that the third point of Lemma 2 in Appendix B holds in these more general situations, and this point is necessary for the proof of Theorem 3. This is related to the notion of *conglomerability* discussed in much detail by Walley (1991). An open problem would be to extend our results to more general situations by addressing the question of conglomerability. On the other hand, as we said, our results are applicable even if the target space $\mathcal{Z}$ of our variable of interest, $Z$, is infinite. This has allowed us, for instance, to model parametric inference in the case where the parameter space is infinite, as discussed in Section 5.2. Moreover, we are not assuming that the upper probability of the values of $Z$ is positive, and this allows us to include the case where our prior beliefs about the parameter are precise and the parameter space is infinite, which coincides with the traditional setup.

Another important point is our assumption on the domains of our lower previsions: we have required for instance that $\underline{P}_1(Z, Y)$ is defined on the set of *all* $\mathcal{X}_{Z \cup Y}$-measurable gambles. Similar considerations have been made for $\underline{P}_2(\bar{W}|\bar{Y})$ and $\underline{P}_3(\hat{W}|\hat{Y})$. When these requirements are not met, we can still apply our results by extending our assessments using the notion of natural extension given by Walley (1991). It is easy to see that these extensions will satisfy the hypotheses of our theorems. These considerations allow us to cover in particular the case where we have lower probabilities instead of lower previsions.

An interesting feature of the CIR rule derived in Theorem 1 is that it allows us to make the passage from the updated information about $Z$ knowing the observation $W$ to the information about $Z$ knowing the value that $Y$ takes. We think that the key for this is that we are using only vacuous and linear previsions to express the information that $Y$ provides about $W$, i.e., that the components of the incompleteness process are either unknown or random. It is an open problem to determine whether there are other possibilities allowing us to have an analogous property. Similarly, it would be interesting to study if the other assumptions in the incompleteness process can be weakened.

To conclude, CIR is a new rule to update beliefs under incompleteness, it is very general, and is based on solid theoretical foundations. It gives us for the first time the opportunity to avoid being both too optimistic and too pessimistic about the incompleteness process, thus creating a basis to draw credible and strong enough conclusions in a number of applications.

This is not to say that we regard CIR as the last word in the subject. On the contrary, it seems to us that research on IPs has so far only scratched the surface of uncertain reasoning under incompleteness. In particular, there will be probably a number of applications where rules stronger than CIR (yet based on more tenable assumption than traditional updating) are needed. Developing these rules appears to be an important research avenue. Such an investigation will probably have to be directed primarily at creating new assumptions about IPs that make the resulting rule stronger while still general enough. A way to do this could be using as a starting point the the framework developed here. In particular, one could take Assumptions (IP1)–(IP9), (BIP3)–(BIP6) and (DB1) as they are, while strengthening those related to the unknown IP, that is, (BIP1) and (BIP2). Then the machinery used to derive CIR could be used again to derive the new rule that follows from the stronger assumptions. Which new assumptions to impose and whether or not they will lead to a rule that is useful while not too domain-specific is matter for future investigation.

## Acknowledgments

Preliminary work on the topic of this paper appeared in the proceedings of ISIPTA '05: the fourth *International Symposium on Imprecise Probabilities and Their Applications* (Zaffalon, 2005). We should like to thank the reviewers of this paper for comments that helped improve its clarity and readability. This work was partially supported by the Swiss NSF grants n. 200020-116674/1 and 200020-121785/1, and by the projects TIN2008-06796-C04-01, MTM2007-61193.

## Appendix A. On the CAR Assumption (IP9)

It is important to realise that not all the multi-valued maps $\Gamma_{\hat{Y}}$ are consistent with the CAR assumption. Here is an example: take $\hat{\mathcal{Y}} := \{1, 2, 3, 4\}$, $\hat{\mathcal{W}} := \{a, b, c\}$, and define the multi-valued map by $\Gamma_{\hat{Y}}(1) := \{a, b\}$, $\Gamma_{\hat{Y}}(2) := \{b, c\}$, $\Gamma_{\hat{Y}}(3) := \{a, c\}$, $\Gamma_{\hat{Y}}(4) := \{a, b, c\}$. By (BIP5), we should have then $P_3(a|1) = P_3(a|3) = P_3(a|4) = \alpha_a$, $P_3(b|1) = P_3(b|2) = P_3(b|4) = \alpha_b$, and $P_3(c|2) = P_3(c|3) = P_3(c|4) = \alpha_c$. Requiring in addition the non-negativity of the probabilities and that the conditional mass functions be normalised, leads to the following system of linear constraints:

$$\alpha_a + \alpha_b = 1$$
$$\alpha_b + \alpha_c = 1$$
$$\alpha_a + \alpha_c = 1$$
$$\alpha_a + \alpha_b + \alpha_c = 1$$
$$\alpha_a, \alpha_b, \alpha_c \geq 0.$$

This system has no solution, as adding the first three equations we deduce that $\alpha_a + \alpha_b + \alpha_c = 1.5$, which is incompatible with the fourth equation. Therefore there is no CAR process consistent with $\Gamma_{\hat{Y}}$ as defined above.

The example shows that in order to define $\Gamma_{\hat{Y}}$ properly, we need to add (IP9) as a further assumption to (IP5) and (IP7): that the system of linear constraints originated by $\Gamma_{\hat{Y}}$ through (BIP5) has a solution, namely, that it is *admissible*. Checking admissibility is easy because $\hat{\mathcal{Y}}$ is finite and hence so is the number of constraints in the system; this allows one to use standard techniques from *linear programming* to solve the problem.

The example also points to another question, which concerns its relationship with Gill et al.'s well-known 'CAR is everything' theorem (Gill et al., 1997, Section 2). Loosely speaking, such a theorem states, in a precise probability context, that it is always possible to define a CAR process; and the example seems to contradict the theorem.

In order to show that there is actually no contradiction, we need to represent the example in the setup of Gill et al.'s paper, which is focused on random sets. This means that they regard $\hat{W}$ as a variable that takes values from the powerset of $\hat{\mathcal{Y}}$, and hence that the CAR IP is intended as a function that maps elements of $\hat{\mathcal{Y}}$ into subsets of $\hat{\mathcal{Y}}$ itself.

To make the transition, we identify each element of $\hat{w} \in \hat{\mathcal{W}}$ with the corresponding set $\{\hat{w}\}^*$; in our example, the possible values of $\hat{W}$, interpreted as a random set, are then $\{1, 3, 4\}$, $\{1, 2, 4\}$, and $\{2, 3, 4\}$. The intuition here is that the element 1 of $\hat{\mathcal{Y}}$ can be coarsened by the CAR IP into the set $\{1, 3, 4\}$ or $\{1, 2, 4\}$; the element 2 into $\{1, 2, 4\}$ or $\{2, 3, 4\}$; the element 3 into $\{1, 3, 4\}$ or $\{2, 3, 4\}$; and the element 4 into $\{1, 3, 4\}$, $\{1, 2, 4\}$ or $\{2, 3, 4\}$. For the correspondence to be really consistent, we also need to make sure that zero probability is assigned to all the elements of the power set of $\hat{\mathcal{Y}}$ that do not correspond to any of the sets $\{\hat{w}\}^*$, $\hat{w} \in \hat{\mathcal{W}}$; this is necessary because of (BIP4) and because Gill et al.'s setup does not explicitly use the notion of multi-valued mapping. With this background we can take a closer look at the relationship between the example and Gill et al.'s theorem.

What the theorem states, in our language, is that no matter how $\hat{W}$ is distributed, one can find an unconditional mass function for $\hat{Y}$ and a CAR process that lead $\hat{W}$ to be distributed in the same way. But if we select an unconditional mass function for the random set $\hat{W}$ that respects the constraint of assigning zero probabilities as above, we are actually implementing the multi-valued map of the example, and we know that there is no CAR process consistent with it.

The key to solve the paradox is that Gill et al. allow the CAR-IP probabilities $P_3(\hat{\mathcal{Y}}'|\hat{y})$ to be non-zero for *any* set $\hat{\mathcal{Y}}' \subseteq \hat{\mathcal{Y}}$ that includes $\hat{y}$, not only for the sets determined by the multi-valued map, as we do. This freedom makes it possible to always find a CAR process.

Consider again the example above, and assume that $\hat{W}$ is distributed uniformly over the three sets $\{1, 3, 4\}$, $\{1, 2, 4\}$, $\{2, 3, 4\}$: i.e., each of them is assigned probability equal to $\frac{1}{3}$. We can make this choice consistent with a CAR process by choosing $\alpha_{\{1,3,4\}} := \alpha_{\{1,2,4\}} := \alpha_{\{2,3,4\}} := \frac{1}{3}$ (i.e., $\alpha_a := \alpha_b := \alpha_c := \frac{1}{3}$) if, in addition, we set $\alpha_{\{1\}} := \alpha_{\{2\}} := \alpha_{\{3\}} := \frac{1}{3}$, i.e., if we allow the CAR process to be able to *potentially* output other three sets. But since we know that the three other sets cannot be really returned, we are obliged to also set $P(1) := P(2) := P(3) := 0$, thus solving the problem, although arguably in a somewhat artificial way.

Indeed, since we are forced to set to zero the probability of some elements in $\hat{\mathcal{Y}}$, one might wonder why those elements have to be included in the set $\hat{\mathcal{Y}}$. Precisely this observation has been already used as a source of criticism of Gill et al.'s 'CAR is everything' theorem (Grünwald & Halpern, 2003, Section 4.4), leading the authors of the latter paper to say that in these cases it is like if there is actually no CAR process. Our formulation based on the multi-valued map simply confirms this statement in an alternative way.

## Appendix B. Coherence of CIR

In this appendix, we are going to prove that the assessments in the construction of the conservative inference rule satisfy the notion of coherence and moreover that they are also coherent with any probabilistic assessment that we deduce from them using CIR. In doing so, we want to cover also the case where the unconditional prevision $\underline{P}_1(Z, Y)$ is constructed from a number of other conditional and unconditional previsions, which is a very typical case in practice. We shall use the notations established in Section 2 and Section 6. We consider then variables $X_1, \ldots, X_n$ taking values in respective sets $\mathcal{X}_1, \ldots, \mathcal{X}_n$, and take a collection of conditional previsions $\underline{P}_1(X_{O_1}|X_{I_1}), \ldots, \underline{P}_m(X_{O_m}|X_{I_m})$.

We start with a lemma that shows that A1 graphs naturally entail a notion of order of the corresponding lower previsions: in particular, that it is possible to permute the indexes of the lower previsions in such a way that the only admissible paths between two dummy nodes are those in which the index of the origin precedes that of the destination.[25]

**Lemma 1** *If the coherence graph of $\{\underline{P}_1(X_{O_1}|X_{I_1}), \ldots, \underline{P}_m(X_{O_m}|X_{I_m})\}$ is A1, then we may assume without loss of generality that for any $k = 1, \ldots, m$, $O_k \cap (\cup_{i=1}^{k-1} I_i) = \emptyset$.*

*Proof.* We start proving that $O_k \cap (\cup_{i=1}^m I_i) = \emptyset$ for some $k$. Assume ex-absurdo that this does not hold. Then for any $k \in \{1, \ldots, m\}$ there is some $f(k) \neq k$ such that $O_k \cap I_{f(k)} \neq \emptyset$. Define $z_0 := 1, z_k := f(z_{k-1}) \ \forall k \geq 1$. It must be $\{z_k\} \cap \{z_0, \ldots, z_{k-1}\} = \emptyset$ for all $k \geq 1$, or we should establish a cycle in the coherence graph, contradicting thus that it is A1. Hence, $|\{z_0, z_1, \ldots, z_{k-1}\}| = k$ for all $k \geq 1$, and this means that $z_m$ does not exist, or, equivalently, that $O_{z_{m-1}} \cap I_i = \emptyset \ \forall i = 1, \ldots, m$. Hence, there is some $k$ such that $O_k \cap (\cup_{i=1}^m I_i) = \emptyset$. We can assume without loss of generality that $k = m$.

We prove now that there is some $k \neq m$ such that $O_k \cap (\cup_{i=1}^{m-1} I_i) = \emptyset$. Assume ex-absurdo that this does not hold. Then for all $k \in \{1, \ldots, m-1\}$ there is some $g(k) \neq k, m$ such that $O_k \cap I_{g(k)} \neq \emptyset$. Define $z_0 := 1, z_k := g(z_{k-1}) \ \forall k \geq 1$. It must be $\{z_k\} \cap \{z_0, \ldots, z_{k-1}\} = \emptyset$ for all $k \geq 1$, or we should establish a cycle in the coherence graph, contradicting thus that it is A1. Hence, $|\{z_0, z_1, \ldots, z_{k-1}\}| = k$ for all $k \geq 1$, and this means that $z_{m-1}$ does not exist, or, equivalently, that $O_{z_{m-2}} \cap I_i = \emptyset \ \forall i = 1, \ldots, m-1$. Hence, there is some $k \neq m$ such that $O_k \cap (\cup_{i=1}^{m-1} I_i) = \emptyset$. We can assume without loss of generality that $k = m-1$.

A similar reasoning allows us to deduce the existence of an order $\prec$ in $\{1, \ldots, m\}$ such that $O_k \cap (\cup_{j \prec k} I_j) = \emptyset$ for all $k = 1, \ldots, m$. Finally we assume without loss of generality that this order coincides with the natural order. $\square$

Now we restrict the attention to the particular case of $A1^+$ graphs. From Lemma 1, if a collection $\underline{P}_1(X_{O_1}|X_{I_1}), \ldots, \underline{P}_m(X_{O_m}|X_{I_m})$ is $A1^+$-representable, we can assume without loss of generality that $I_1 = \emptyset$. Let $A_1 := \emptyset, A_j := \cup_{i=1}^{j-1}(I_i \cup O_i)$ for $j = 2, \ldots, m+1$, and let $\underline{P}'_j(X_{O_j}|X_{A_j \cup I_j})$ be given on the set $\mathcal{H}^j$ of $\mathcal{X}_{A_{j+1}}$-measurable gambles for $j = 1, \ldots, m$ by

$$\underline{P}'_j(f|z) := \underline{P}_j(f(z, \cdot)|\pi_{I_j}(z))$$

for all $z \in \mathcal{X}_{A_j \cup I_j}$ and all $f \in \mathcal{H}^j$. Since $\underline{P}_j(X_{O_j}|X_{I_j})$ is separately coherent for $j = 1, \ldots, m$, so is $\underline{P}'_j(X_{O_j}|X_{A_j \cup I_j})$. Moreover, thanks to Lemma 1 and that $\{O_1, \ldots, O_m\}$ forms a partition of $\{1, \ldots, n\}$ when we focus on $A1^+$ graphs, the sets of indexes of the conditional variables in $\underline{P}'_1(X_{O_1}), \ldots, \underline{P}'_m(X_{O_m}|X_{A_m \cup I_m})$ form an increasing sequence and hence they satisfy the hypotheses of the generalised marginal extension theorem. As a consequence, $\underline{P}'_1(X_{O_1}), \ldots, \underline{P}'_m(X_{O_m}|X_{A_m \cup I_m})$ are also coherent.

A similar reasoning shows that if we take for $j = 1, \ldots, m$ a conditional linear prevision $P'_j(X_{O_j}|X_{A_j \cup I_j})$ that dominates $\underline{P}'_j(X_{O_j}|X_{A_j \cup I_j})$, then $P'_1(X_{O_1}), \ldots, P'_m(X_{O_m}|X_{A_m \cup I_m})$ are jointly coherent. Moreover, since $\{O_1, \ldots, O_m\}$ is a partition of $\{1, \ldots, n\}$, Theorem 3 by

---

25. This order notion is similar to the graph-theoretic notion of *topological ordering*, but here it is applied only to the dummy nodes.

Miranda and De Cooman (2007) implies that the *only* prevision $P$ on $\mathcal{X}^n$ which is coherent with the assessments $P'_1(X_{O_1}), \ldots, P'_m(X_{O_m}|X_{A_m \cup I_m})$ is

$$P(f) = P'_1(P'_2(\ldots(P'_m(f|X_{A_m \cup I_m})|\ldots)|X_{A_2 \cup I_2})). \tag{15}$$

In other words, $P'_1(X_{O_1}), \ldots, P'_m(X_{O_m}|X_{A_m \cup I_m})$ give rise to a unique joint lower prevision.

**Definition 17** *Assume that $P_\lambda(X_{O_j}|X_{A_j \cup I_j})$ dominates $\underline{P}'_j(X_{O_j}|X_{A_j \cup I_j})$ for all $\lambda \in \Lambda$ and all $j = 1, \ldots, m$ and*

$$\inf_{\lambda \in \Lambda} P_\lambda(X_{O_j}|X_{A_j \cup I_j}) = \underline{P}'_j(X_{O_j}|X_{A_j \cup I_j}).$$

*The coherent lower prevision $\underline{P}$ defined as $\underline{P} := \inf_{\lambda \in \Lambda} P_\lambda$, where $P_\lambda$ is the coherent prevision determined by $P_\lambda(X_{O_1}), \ldots, P_\lambda(X_{O_m}|X_{A_m \cup I_m})$ and Equation (15), is called a* lower envelope model.

Intuitively, a lower envelope model is a joint lower prevision that is built out of a number of conditional and unconditional assessments. The interest in lower envelope models arises because it is a very common practice to build joint models out of smaller conditional and unconditional ones, and then to use the joint model to draw some conclusions. Lower envelope models abstract this procedure in the general case of coherent lower previsions. As particular cases of lower envelope models, we can consider the following:

1. If for each $j = 1, \ldots, m$ we consider all the $P_\lambda(X_{O_j}|X_{A_j \cup I_j})$ in $\mathcal{M}(\underline{P}'_j(X_{O_j}|X_{A_j \cup I_j}))$, then $\underline{P}$ is the *marginal extension* of $\underline{P}'_1(X_{O_1}), \ldots, \underline{P}'_m(X_{O_m}|X_{A_m \cup I_m})$.

2. If for $j = 1, \ldots, m$ we take all the $P_\lambda(X_{O_j}|X_{A_j \cup I_j})$ in the set of extreme points of $\mathcal{M}(\underline{P}'_j(X_{O_j}|X_{A_j \cup I_j}))$, with the additional requirement that $P_\lambda(X_{O_j}|z) = P_\lambda(X_{O_j}|z')$ if $\pi_{I_j}(z) = \pi_{I_j}(z')$, then the lower envelope model $\underline{P}$ is called the *strong product* of $\underline{P}_1(X_{O_1}), \ldots, \underline{P}_m(X_{O_m}|X_{I_m})$.

In this paper, we make our inferences using the strong product. From the results by Miranda and De Cooman (2007), if we let $P_j(X_{O_j}|X_{I_j})$ be an extreme point of $\mathcal{M}(\underline{P}_j(X_{O_j}|X_{I_j}))$ for $j = 1, \ldots, m$ and we build a linear prevision $P$ in the manner described above, then $P, P_1(X_{O_1}|X_{A_1 \cup I_1}), \ldots, P_m(X_{O_m}|X_{A_m \cup I_m})$ are coherent. Moreover, we deduce the following:

**Theorem 2** *Let $\underline{P}_1(X_{O_1}), \ldots, \underline{P}_m(X_{O_m}|X_{I_m})$ be an A1$^+$-representable collection, and let $\underline{P}$ be a lower envelope model associated to it. Then $\underline{P}, \underline{P}_1(X_{O_1}), \ldots, \underline{P}_m(X_{O_m}|X_{I_m})$ are coherent.*

*Proof.* It is a consequence of the marginal extension theorem by Miranda and De Cooman (2007) that for any $\lambda \in \Lambda$ the previsions $P_\lambda, P_\lambda(X_{O_1}|X_{A_1 \cup I_1}), \ldots, P_\lambda(X_{O_m}|X_{A_m \cup I_m})$ are coherent. Applying Theorem 8.1.6 in Walley (1991), we deduce that the lower envelopes of these families, which are the lower previsions $\underline{P}, \underline{P}'_1(X_{O_1}|X_{A_1 \cup I_1}), \ldots, \underline{P}'_m(X_{O_m}|X_{A_m \cup I_m})$, are also coherent. The result now follows applying that any gamble $f$ in $\mathcal{K}^j$ belongs to $\mathcal{H}^j$, and that $\underline{P}'_j(f|X_{A_j \cup I_j})(x) = \underline{P}'_j(f(\pi_{A_j \cup I_j}(x), \cdot)|\pi_{A_j \cup I_j}(x)) = \underline{P}_j(f(\pi_{I_j}(x), \cdot)|\pi_{I_j}(x))$ for all $x \in \mathcal{X}^n$. $\square$

We have the following useful corollary of this theorem:

**Corollary 1** *Let $\underline{P}$ be the strong product of $\underline{P}_1(Z,Y), \underline{P}_2(\bar{W}|\bar{Y}), P_3(\hat{W}|\hat{Y})$, constructed in Section 7.1.3. $\underline{P}$ is coherent and it is also jointly coherent with the original assessments $\underline{P}_1(Z,Y), \underline{P}_2(\bar{W}|\bar{Y}), P_3(\hat{W}|\hat{Y})$. This holds also if $\underline{P}_1(Z,Y)$ is constructed as a joint which is coherent with a number of smaller pieces of information, provided that the coherence graph representing them is A1$^+$.*

*Proof.* That $\underline{P}$ is a coherent lower prevision follows because it is the lower envelope of a set of linear previsions. Its coherence with $\underline{P}_1(Z,Y), \underline{P}_2(\bar{W}|\bar{Y}), P_3(\hat{W}|\hat{Y})$ follows from Theorem 2, because the coherence graph of $\underline{P}_1(Z,Y), \underline{P}_2(\bar{W}|\bar{Y}), P_3(\hat{W}|\hat{Y})$ is A1$^+$, as it is evident from Figure 6. In the more general situation when $\underline{P}_1(Z,Y)$ is constructed in a modular way the result follows similarly as the coherence graph representing these assessments together with $\underline{P}_2(\bar{W}|\bar{Y}), P_3(\hat{W}|\hat{Y})$ will also be A1$^+$. □

We prove next that if we use regular extension to derive new conditional previsions $\underline{P}_{m+1}(X_{O_{m+1}}|X_{I_{m+1}}), \ldots, \underline{P}_{m+k}(X_{O_{m+k}}|X_{I_{m+k}})$ from the strong product $\underline{P}$, these conditional previsions are coherent with all the initial assessments. For this, we need to establish first the following:

**Lemma 2** *Let $\underline{P}, \underline{P}(X_O|X_I)$ be coherent unconditional and conditional previsions, with $\mathcal{X}_I$ finite. Let $\underline{R}(X_O|X_I)$ be defined from $\underline{P}$ using regular extension for $z \in \mathcal{X}_I$ when $\overline{P}(z) > 0$, and be equal to $\underline{P}(X_O|z)$ when $\overline{P}(z) = 0$. Then:*

1. *$\underline{P}, \underline{R}(X_O|X_I)$ are coherent.*

2. *$\underline{R}(X_O|X_I) \geq \underline{P}(X_O|X_I)$.*

3. *For any $P \geq \underline{P}$, there is some $P(X_O|X_I)$ which is coherent with $P$ and dominates $\underline{P}(X_O|X_I)$.*

*Proof.* Since $\mathcal{X}_I$ is a finite set, we can apply Theorem 6.5.4 in the book of Walley (1991) to deduce that the coherence of $\underline{P}, \underline{R}(X_O|X_I)$ is equivalent to $\underline{P}(\mathbb{I}_z(f - \underline{R}(f|z))) = 0$ for all $z \in \mathcal{X}_I$.[26] If $\overline{P}(z) = 0$, this is trivial. If $\overline{P}(z) > 0$, it follows from Walley (1991, Appendix J3).

For the second statement, consider some $z$ in $\mathcal{X}_I$ with $\overline{P}(z) > 0$, and $f \in \mathcal{K}_{O \cup I}$. Assume ex-absurdo that $\underline{R}(f|z) < \underline{P}(f|z)$. It follows from the definition of the regular extension that there is some $P \geq \underline{P}$ such that $P(z) > 0$ and $P(f|z) < \underline{P}(f|z)$. Since $P(z) > 0$, it follows from the generalised Bayes rule that $P(f|z)$ is the *unique* value satisfying $0 = P(\mathbb{I}_z(f - P(f|z)))$. As a consequence, given $\underline{P}(f|z) > P(f|z)$, we have that $\mathbb{I}_z(f - P(f|z)) \geq \mathbb{I}_z(f - \underline{P}(f|z))$, whence

$$0 = P(\mathbb{I}_z(f - P(f|z))) \geq P(\mathbb{I}_z(f - \underline{P}(f|z))) \geq \underline{P}(\mathbb{I}_z(f - \underline{P}(f|z)) = 0,$$

using that since $\underline{P}, \underline{P}(X_O|X_I)$ are coherent they satisfy the generalised Bayes rule. But this implies that $P(\mathbb{I}_z(f - P(f|z))) = P(\mathbb{I}_z(f - \underline{P}(f|z))) = 0$, and then there are two different values of $\mu$ for which $P(\mathbb{I}_z(f - \mu)) = 0$. This is a contradiction.

---

26. This is called the *generalised Bayes rule*. When $\underline{P}(x) > 0$, there is a *unique* value for which $\underline{P}(G(f|x)) = \underline{P}(\mathbb{I}_x(f - \underline{P}(f|x))) = 0$ holds.

We finally establish the third statement. Consider $P \geq \underline{P}$, and $z \in \mathcal{X}_I$. If $P(z) > 0$, then for any $f \in \mathcal{K}_{O \cup I}$ $P(f|z)$ is uniquely determined by the generalised Bayes rule and dominates the regular extension $\underline{R}(f|z)$. Hence, $P(f|z) \geq \underline{R}(f|z) \geq \underline{P}(f|z)$, where the last inequality follows from the second statement. Finally, if $P(z) = 0$, taking any element $P(X_O|z)$ of $\mathcal{M}(\underline{P}(X_O|z))$ we have that $P(\mathbb{I}_z(f - P(f|z))) = 0$ for all $f \in \mathcal{K}_{O \cup I}$. This completes the proof. $\square$

Before we establish our next result, we need to introduce a consistency notion for conditional lower previsions that is less restrictive than coherence:

**Definition 18** Let $\underline{P}_1(X_{O_1}|X_{I_1}), \ldots, \underline{P}_m(X_{O_m}|X_{I_m})$ be separately coherent conditional previsions. They are weakly coherent when for all $f_j \in \mathcal{K}^j$, $j = 1, \ldots, m$, $j_0 \in \{1, \ldots, m\}$, $f_0 \in \mathcal{K}^{j_0}, z_0 \in \mathcal{X}_{I_{j_0}}$,

$$\sup_{x \in \mathcal{X}^n} \left[ \sum_{j=1}^m G_j(f_j|X_{I_j}) - G_{j_0}(f_0|z_0) \right](x) \geq 0. \tag{16}$$

The intuition behind this notion is that our subject should not raise the conditional lower prevision $\underline{P}_{j_0}(f_0|z_0)$, which represents his supremum acceptable buying price for $f_0$ contingent on $z_0$, in any positive $\epsilon$, using the desirable gambles $G_1(f_1|X_{I_1}), \ldots, G_m(f_m|X_{I_m})$. The difference with the notion of (strong) coherence is that the supremum of sum in Equation (16) is required to be non-negative over the whole space $\mathcal{X}^n$, and not necessarily when at least one of the summands is non-zero. This implies that the condition holds trivially for gambles $f_0, f_1, \ldots, f_m$ for which there are elements $w \in \mathcal{X}^n$ which do not belong to any set in $\{\pi_{I_{j_0}}^{-1}(z_0)\} \cup \cup_{j=1}^m S_j(f_j)$.

From Miranda and Zaffalon (2009, Theorem 1), $\underline{P}_1(X_{O_1}|X_{I_1}), \ldots, \underline{P}_m(X_{O_m}|X_{I_m})$ are weakly coherent if and only if there is a joint lower prevision $\underline{P}(X_1, \ldots, X_n)$ that is pairwise coherent with each conditional lower prevision $\underline{P}_j(X_{O_j}|X_{I_j})$ in the collection. Similar results hold in the case where we focus on collections made of linear previsions, with the difference that the joint whose existence is equivalent to weak coherence is linear, too. However, under the behavioural interpretation, a number of weakly coherent conditional lower previsions can still present some forms of inconsistency; see Walley (1991, Example 7.3.5) for an example and Walley (1991, Chapter 7), Walley, Pelessoni, and Vicig (2004), Miranda (2009) for some discussion.

**Theorem 3** Let $\underline{P}_1(X_{O_1}|X_{I_1}), \ldots, \underline{P}_m(X_{O_m}|X_{I_m})$ be separately coherent conditional lower previsions whose associated coherence graph is $A1^+$. Let $\underline{P}$ be their strong product. Consider disjoint $O_{m+j}, I_{m+j}$ for $j = 1, \ldots, k$. Assume that $\mathcal{X}_{I_{m+j}}$ is finite for $j = 1, \ldots, k$ and that $\overline{P}(z) > 0$ for all $z \in \mathcal{X}_{I_{m+j}}$, and define $\underline{P}_{m+j}(X_{O_{m+j}}|X_{I_{m+j}})$ using regular extension for $j = 1, \ldots, k$. Then

$$\underline{P}, \underline{P}_1(X_{O_1}|X_{I_1}), \ldots, \underline{P}_{m+k}(X_{O_{m+k}}|X_{I_{m+k}}) \text{ are coherent.}$$

*Proof.* Let $\mathcal{M}$ denote the set of linear previsions constructed by combining the extreme points of $\mathcal{M}(\underline{P}_j(X_{O_j}|X_{I_j})), j = 1, \ldots, m$, in the manner described by Equation (15). The strong product $\underline{P}$ is the lower envelope of $\mathcal{M}$.

Since $\mathcal{X}_{I_{m+j}}$ is finite for $j = 1, \dots, k$, it follows from [Appendix J3] in the book of Walley (1991) that $\underline{P}$ and $\underline{P}_{m+j}(X_{O_{m+j}}|X_{I_{m+j}})$ are coherent. Applying Theorem 1 by Miranda and Zaffalon (2009), we deduce that the previsions $\underline{P}, \underline{P}_1(X_{O_1}|X_{I_1}), \dots, \underline{P}_{m+k}(X_{O_{m+k}}|X_{I_{m+k}})$ are weakly coherent. Hence, Theorem 7.1.5 by Walley (1991) implies that it suffices to show that $\underline{P}_1(X_{O_1}|X_{I_1}), \dots, \underline{P}_{m+k}(X_{O_{m+k}}|X_{I_{m+k}})$ are coherent. Consider $f_i \in \mathcal{K}^i$ for $i = 1, \dots, m+k$, $j_0 \in \{1, \dots, m+k\}$, $z_0 \in \mathcal{X}_{I_{j_0}}$, $f_0 \in \mathcal{K}^{j_0}$, and let us prove that

$$\sup_{\omega \in B} \left[ \sum_{i=1}^{m+k} [f_i - \underline{P}_i(X_{O_i}|X_{I_i})] - \mathbb{I}_{z_0}(f_0 - \underline{P}_{j_0}(f_0|z_0)) \right](\omega) \geq 0 \tag{17}$$

for some $B \in \{\pi_{I_{j_0}}^{-1}(z_0)\} \cup \cup_{i=1}^{m+k} S_i(f_i)$.

Assume first that $j_0 \in \{m+1, \dots, m+k\}$. If Equation (17) does not hold, there is some $\delta > 0$ such that

$$\sup_{\omega \in \pi_{I_{j_0}}^{-1}(z_0)} \left[ \sum_{i=1}^{m+k} [f_i - \underline{P}_i(X_{O_i}|X_{I_i})] - \mathbb{I}_{z_0}(f_0 - \underline{P}_{j_0}(f_0|z_0)) \right](\omega) = -\delta < 0.$$

Since $\overline{P}(z_0) > 0$ by assumption, it follows from the definition of the regular extension that there is some $P$ in $\mathcal{M}$ such that $P(z_0) > 0$ and $P_{j_0}(f_0|z_0) - \underline{P}_{j_0}(f_0|z_0) < \frac{\delta}{2}$. From the definition of the elements of $\mathcal{M}$, there are $P_i(X_{O_i}|X_{I_i})$ in $\mathcal{M}(\underline{P}_i(X_{O_i}|X_{I_i}))$, $i = 1, \dots, m$, such that $P$ is coherent with $P_i(X_{O_i}|X_{I_i})$ for $i = 1, \dots, m$. On the other hand, applying Lemma 2, for any $j = 1, \dots, k$, there is a conditional prevision $P_{m+j}(X_{O_{m+j}}|X_{I_{m+j}})$ that dominates $\underline{P}_{m+j}(X_{O_{m+j}}|X_{I_{m+j}})$ and is coherent with $P$. Note that $P_{j_0}(X_{O_{j_0}}|z_0)$ is uniquely determined from $P$ by Bayes's rule because $P(z_0) > 0$.

Using these conditional previsions, we deduce that for any $\omega \in \mathcal{X}^n$,

$$\left[ \sum_{i=1}^{m+k} [f_i - P_i(X_{O_i}|X_{I_i})] - \mathbb{I}_{z_0}(f_0 - P_{j_0}(f_0|z_0)) \right](\omega)$$
$$\leq \left[ \sum_{i=1}^{m+k} [f_i - \underline{P}_i(X_{O_i}|X_{I_i})] - \mathbb{I}_{z_0}(f_0 - \underline{P}_{j_0}(f_0|z_0)) \right](\omega) + \frac{\delta}{2},$$

whence

$$\left[ \sum_{i=1}^{m+k} [f_i - P_i(X_{O_i}|X_{I_i})] - \mathbb{I}_{z_0}(f_0 - P_{j_0}(f_0|z_0)) \right](\omega) \leq -\frac{\delta}{2}$$

for any $\omega \in \pi_{I_{j_0}}^{-1}(z_0)$. Let us denote $g := \sum_{i=1}^{m+k} [f_i - P_i(X_{O_i}|X_{I_i})] - \mathbb{I}_{z_0}(f_0 - P_{j_0}(f_0|z_0))$. The coherence of $P, P_j(X_{O_j}|X_{I_j})$ for $j = 1, \dots, m+k$ implies that $P(g) = 0$. But then we also have $P(g) \leq P(g\mathbb{I}_{z_0}) \leq -\frac{\delta}{2}P(z_0) < 0$, where the first inequality holds because $g \leq 0$ since we are assuming that Equation (17) does not hold. This is a contradiction.

Assume next that $j_0 \in \{1, \dots, m\}$, and let us prove the existence of a linear prevision $Q \in \mathcal{M}$ and conditional previsions $P_j(X_{O_j}|X_{I_j})$ in $\mathcal{M}(\underline{P}_j(X_{O_j}|X_{I_j}))$ coherent with $Q$ for $j = 1, \dots, m+k$ such that $P_{j_0}(f_0|z_0) = \underline{P}_{j_0}(f_0|z_0)$ and $Q(B) > 0$ for some $B \in \{\pi_{I_{j_0}}^{-1}(z_0)\} \cup \cup_{i=1}^{m+k} S_i(f_i)$.

- Assume first that $\overline{P}(z_0) > 0$. Then there is some $P$ in $\mathcal{M}$ such that $P(z_0) > 0$. This prevision is determined by (and therefore coherent with) conditional previsions $P_1(X_{O_{I_1}}|X_{I_1}), \ldots, P_m(X_{O_m}|X_{I_m})$, where $P_i(X_{O_i}|X_{I_i})$ belongs to $\mathcal{M}(\underline{P}_i(X_{O_i}|X_{I_i}))$ for $i = 1, \ldots, m$.

  Let $P'_{j_0}(X_{O_{j_0}}|X_{I_{j_0}})$ be an extreme point of $\mathcal{M}(\underline{P}_{j_0}(X_{O_{j_0}}|X_{I_{j_0}}))$ such that $P'_{j_0}(f_0|z_0) = \underline{P}(f_0|z_0)$, and let $Q$ be the element of $\mathcal{M}$ determined by the conditional previsions $P_1(X_{O_1}|X_{I_1}), \ldots, P'_{j_0}(X_{O_{j_0}}|X_{I_{j_0}}), \ldots, P_m(X_{O_m}|X_{I_m})$. Since the coherence graph of $\underline{P}_1(X_{O_1}|X_{I_1}), \ldots, \underline{P}_m(X_{O_m}|X_{I_m})$ is A1, Lemma 1 implies that we can assume without loss of generality that for all $k = 1, \ldots, m$, $O_k \cap (\cup_{i=1}^{k-1} I_i) = \emptyset$. Since moreover $\{O_1, \ldots, O_m\}$ forms a partition of $\{1, \ldots, n\}$, we deduce that $I_i \subseteq \cup_{j=1}^{i-1} O_j$ for $i = 1, \ldots, n$. In particular, $I_{j_0} \subseteq \cup_{j=1}^{j_0-1} O_j$, whence the value of $Q$ on $\mathcal{X}_{I_{j_0}}$-measurable gambles is uniquely determined by $P_1(X_{O_1}|X_{I_1}), \ldots, P_{j_0-1}(X_{O_{j_0-1}}|X_{I_{j_0-1}})$. As a consequence, $Q(z_0) = P(z_0) > 0$.

  Consider now conditional previsions $P_{m+j}(X_{O_{m+j}}|X_{I_{m+j}})$ which are pairwise coherent with $Q$ and dominate $\underline{P}_{m+j}(X_{O_{m+j}}|X_{I_{m+j}})$ for $j = 1, \ldots, k$. There are such previsions because of Lemma 2. The previsions

$$Q, \quad P_1(X_{O_1}|X_{I_1}), \ldots, P'_{j_0}(X_{O_{j_0}}|X_{I_{j_0}}), \ldots, P_m(X_{O_m}|X_{I_m}),$$
$$P_{m+1}(X_{O_{m+1}}|X_{I_{m+1}}), \ldots, P_{m+k}(X_{O_{m+k}}|X_{I_{m+k}})$$

  satisfy the conditions stated above, with $B = \pi_{I_{j_0}}^{-1}(z_0)$.

- If $\overline{P}(z_0) = 0$, there are two possibilities: either $f_{m+1} = \cdots = f_{m+k} = 0$, and then Equation (17) holds because $\underline{P}_1(X_{O_1}|X_{I_1}), \ldots, \underline{P}_m(X_{O_m}|X_{I_m})$ are coherent; or there is some $j_1 \in \{1, \ldots, k\}$ such that $f_{m+j_1} \neq 0$. In that case, we consider a set $B \in S_{m+j}(f_{m+j_1})$. This set is of the form $B = \pi_{I_{m+j_1}}^{-1}(z_1)$ for some $z_1 \in \mathcal{X}_{I_{m+j_1}}$. Because $\overline{P}(z_1) > 0$ by assumption, there is some $Q \in \mathcal{M}$ such that $Q(z_1) > 0$. From the definition of $\mathcal{M}$, there are $P_i(X_{O_i}|X_{I_i})$ in $\mathcal{M}(\underline{P}_i(X_{O_i}|X_{I_i}))$, $i = 1, \ldots, m$, such that $Q$ is coherent with $P_i(X_{O_i}|X_{I_i})$ for $i = 1, \ldots, m$. Consider a prevision $P'_{j_0}(X_{O_{j_0}}|z_0)$ in $\mathcal{M}(\underline{P}_{j_0}(X_{O_{j_0}}|z_0))$ such that $P'_{j_0}(f_0|z_0) = \underline{P}_{j_0}(f_0|z_0)$, and define the conditional prevision $P'_{j_0}(X_{O_{j_0}}|X_{I_{j_0}})$ by

$$P'_{j_0}(f|z) = \begin{cases} P'_{j_0}(f|z_0) \text{ if } z = z_0 \\ P_{j_0}(f|z_0) \text{ otherwise.} \end{cases}$$

  Since $Q(z_0) = \overline{P}(z_0) = 0$, given any $f \in \mathcal{K}^{j_0}$ $Q(P'_{j_0}(f|X_{I_{j_0}})) = Q(P_{j_0}(f|X_{I_{j_0}})) = Q(f)$, where the last equality follows from the coherence of $Q$ and $P_{j_0}(X_{O_{j_0}}|X_{I_{j_0}})$. Hence, $Q$ and $P'_{j_0}(X_{O_{j_0}}|X_{I_{j_0}})$ are coherent. On the other hand, applying Lemma 2, for all $j = 1, \ldots, k$, there is $P_{m+j}(X_{O_{m+j}}|X_{I_{m+j}}) \geq \underline{P}_{m+j}(X_{O_{m+j}}|X_{I_{m+j}})$ that is coherent with $Q$. From all this we deduce that

$$Q, P_1(X_{O_1}|X_{I_1}), \ldots, P'_{j_0}(X_{O_{j_0}}|X_{I_{j_0}}), \ldots, P_{m+k}(X_{O_{m+k}}|X_{I_{m+k}})$$

  satisfy the above stated conditions, with $B = \pi_{I_{m+j_1}}^{-1}(z_1)$.

815

If we now take these previsions, we deduce that for all $\omega \in \mathcal{X}^n$,

$$\left[ \sum_{i=1}^{m+k} [f_i - P_i(X_{O_i}|X_{I_i})] - \mathbb{I}_{z_0}(f_0 - P_{j_0}(f_0|z_0)) \right] (\omega)$$

$$\leq \left[ \sum_{i=1}^{m+k} [f_i - \underline{P}_i(X_{O_i}|X_{I_i})] - \mathbb{I}_{z_0}(f_0 - \underline{P}_{j_0}(f_0|z_0)) \right] (\omega).$$

Let us denote $g := \sum_{i=1}^{m+k} [f_i - P_i(X_{O_i}|X_{I_i})] - \mathbb{I}_{z_0}(f_0 - P_{j_0}(f_0|z_0))$. The coherence of $Q$ and $P_j(X_{O_j}|X_{I_j})$ for $j = 1, \ldots, m+k$ implies that $Q(g) = 0$. Consider the set $B \in \{\pi_{I_{j_0}}^{-1}(z_0)\} \cup \cup_{i=1}^{m+k} S_i(f_i)$ for which $Q(B) > 0$. If Equation (17) does not hold, then there must be some $\delta > 0$ s.t. $\sup_{\omega \in B} g(\omega) = -\delta < 0$. Since it also follows that $g(\omega) \leq 0$, we deduce that $Q(g) \leq Q(g\mathbb{I}_B) \leq -\delta Q(B) < 0$. This is a contradiction.

We conclude from this that Equation (17) holds and as a consequence the previsions $\underline{P}, \underline{P}_1(X_{O_1}|X_{I_1}), \ldots, \underline{P}_{m+k}(X_{O_{m+k}}|X_{I_{m+k}})$ are coherent. $\square$

Next we use the results above to apply the CIR rule in more general frameworks. This is necessary for some of the examples of application of the CIR rule discussed in Sections 5.1–5.3.

**Corollary 2** *Let $\{X_1, \ldots, X_n\}$ be a set of variables that contains $\{Z, Y\}$ and does not include $W$. Assume that the coherence graph associated to $\underline{P}_1(X_{O_1}|X_{I_1}), \ldots, \underline{P}_m(X_{O_m}|X_{I_m})$ is $A1^+$. Let $\underline{P}$ be the strong product of*

$$\underline{P}_1(X_{O_1}|X_{I_1}), \ldots, \underline{P}_m(X_{O_m}|X_{I_m}), \underline{P}(\hat{W}|\hat{Y}), \underline{P}(\bar{W}|\bar{Y}).$$

*Assume that $\mathcal{X}_{I_{m+j}}$ is finite for $j = 1, \ldots, k$ and that $\overline{P}(z) > 0$ for all $z \in \mathcal{X}_{I_{m+j}}$, and define $\underline{P}_{m+j}(X_{O_{m+j}}|X_{I_{m+j}})$ using regular extension for $j = 1, \ldots, k$. Then:*

(i) *$\underline{P}, \underline{P}_1(X_{O_1}|X_{I_1}), \ldots, \underline{P}_m(X_{O_{m+k}}|X_{I_{m+k}}), \underline{P}(\hat{W}|\hat{Y}), \underline{P}(\bar{W}|\bar{Y})$ are coherent.*

(ii) *If $\underline{P}(Z|W)$ is one of the previsions we derive from the strong product $\underline{P}$ using regular extension, then $\underline{P}(Z|W)$ satisfies Equation (CIR).*

*Proof.* First of all, $\underline{P}_1(X_{O_1}|X_{I_1}), \ldots, \underline{P}_m(X_{O_m}|X_{I_m}), \underline{P}(\hat{W}|\hat{Y}), \underline{P}(\bar{W}|\bar{Y})$ satisfy the hypotheses of Theorem 2: their associated coherence graph is A1$^+$ and $\{W, X_{O_1}, \ldots, X_{O_m}\}$ is our set of variables of interest. Hence, their strong product $\underline{P}$ is coherent with all these assessments. If moreover we use regular extension to build the updated models $\underline{P}_{m+j}(X_{O_{m+j}}|X_{I_{m+j}})$ using regular extension for $j = 1, \ldots, k$, it follows from Theorem 3 that the conditional lower previsions $\underline{P}_{m+1}(X_{O_{m+1}}|X_{I_{m+1}}), \ldots, \underline{P}_{m+k}(X_{O_{m+k}}|X_{I_{m+k}})$ are coherent with $\underline{P}, \underline{P}_1(X_{O_1}|X_{I_1}), \ldots, \underline{P}_m(X_{O_m}|X_{I_m}), \underline{P}(\hat{W}|\hat{Y}), \underline{P}(\bar{W}|\bar{Y})$.

Secondly, the strong product $\underline{P}$ coincides with the one we obtain using the unconditional lower prevision $\underline{P}(X_{O_1}, \ldots, X_{O_m})$ which we obtain by making the strong product of $\underline{P}_1(X_{O_1}|X_{I_1}), \ldots, \underline{P}_m(X_{O_m}|X_{I_m})$ on the one hand and the conditional assessments $\underline{P}(\hat{W}|\hat{Y}), \underline{P}(\bar{W}|\bar{Y})$ on the other hand: from the results by Miranda and De Cooman (2007), the extreme points of the set $\mathcal{M}(\underline{P}(X_{O_1}, \ldots, X_{O_m}))$ are obtained by applying marginal extension on the extreme points of $\mathcal{M}(\underline{P}_1(X_{O_1}|X_{I_1})), \ldots, \mathcal{M}(\underline{P}_m(X_{O_m}|X_{I_m}))$. Hence, the updated prevision $\underline{P}(Z|W)$ we obtain by regular extension also satisfies Equation (CIR) in this case. $\square$

**Corollary 3** *Updating a credal network a finite number of times by CIR leads to coherent inference.*

*Proof.* It is enough to observe that the probabilistic assessments used to build credal networks lead to an A1$^+$ coherence graph, as detailed in Theorem 8 by Miranda and Zaffalon (2009), and that such a graph, supplemented with the parts for $\underline{P}(\bar{W}|\bar{Y})$ and $\underline{P}(\hat{W}|\hat{Y})$, remains an A1$^+$ graph. The result follows then by Corollary 2. $\square$

**Corollary 4** *Doing parametric inference a finite number of times by CIR according to Rules (4) and (5) leads to coherent inference.*

*Proof.* The coherence of Rule (4) is ensured by Corollary 2 since the coherence graph of the overall assessments used is A1$^+$, as shown in Figure 8. The same holds for Rule (5) once we observe that also in this case the related coherence graph is A1$^+$, as illustrated in Figure 9. $\square$

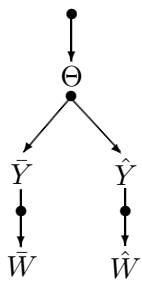

Figure 8: The coherence graph for parametric statistical inference.

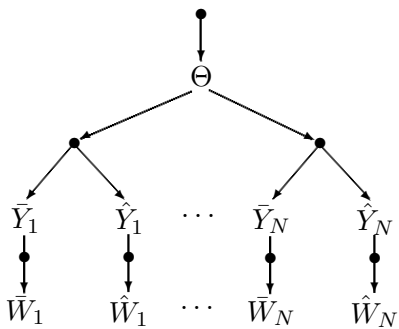

Figure 9: The coherence graph for parametric statistical inference in the IID+ID case.

**Corollary 5** *Computing a finite number of posterior predictive lower previsions by CIR for classification according to Rule (7) leads to coherent inference.*

*Proof.* The coherence of Rule (7) is ensured by Corollary 2 since the coherence graph of the overall assessments used is A1$^+$, as shown in Figure 10. $\square$

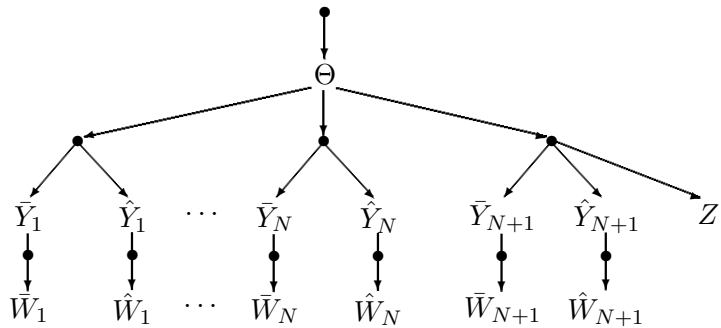

Figure 10: The coherence graph for pattern classification in the IID+ID case.

**Lemma 3** *Consider $b_j \geq 0, c_j > 0$ for $j = 1, \ldots, n$. Then there is some $j_1$ such that*

$$\left(\frac{\sum_{j=1}^n b_j}{\sum_{j=1}^n c_j}\right) \geq \frac{b_{j_1}}{c_{j_1}}. \tag{18}$$

*Proof.* Assume ex-absurdo that Equation (18) does not hold. Then for any $j = 1, \ldots, m$, $\frac{c_j \sum_{k=1}^n b_k}{\sum_{k=1}^n c_k} < b_j$, and by making the sum over all $j$ on both sides of the inequality we obtain that

$$\sum_{j=1}^n \frac{c_j \sum_{k=1}^n b_k}{\sum_{k=1}^n c_k} = \sum_{k=1}^n b_k < \sum_{j=1}^n b_j,$$

a contradiction. □